\documentclass[10pt,twocolumn,letterpaper]{article}

\usepackage{cvpr}              
\usepackage[utf8]{inputenc} 
\usepackage[T1]{fontenc}    
\usepackage{url}            
\usepackage{booktabs}       
\usepackage{amsfonts}       
\usepackage{nicefrac}       
\usepackage{xcolor}         
\usepackage{tocloft}    
\usepackage{xstring}

\usepackage[dvipsnames, svgnames, x11names]{xcolor}  
\usepackage{colortbl}
\usepackage{xcolor}         
\usepackage{times}
\usepackage{epsfig}
\usepackage{graphicx}
\usepackage{amsmath}
\usepackage{makecell} 
\usepackage{booktabs,multirow,array} 
\usepackage{xcolor}
\usepackage{colortbl}
\usepackage{rotating}
\usepackage{pifont}
\usepackage{utfsym}
\usepackage{wrapfig}
\usepackage{tabularx}
\usepackage{xspace}
\usepackage[utf8]{inputenc} 
\usepackage[T1]{fontenc}    
\usepackage{url}            
\usepackage{amsfonts}       
\usepackage{mathtools} 
\usepackage{nicefrac}       
\usepackage{microtype}      
\usepackage[font={small}]{caption}
\usepackage[hang,flushmargin]{footmisc} 
\usepackage{wrapfig}
\usepackage{adjustbox}
\usepackage{textcomp}

\newcommand{\HG}{\mathcal{H}} %
\newcommand{\cmark}{\ding{51}}%
\newcommand{\xmark}{\ding{53}}%

\definecolor{colorbest}{RGB}{255,179,179}
\definecolor{colorsecond}{RGB}{255,217,179}
\definecolor{colorthird}{RGB}{255,255,179}
\definecolor{antiquefuchsia}{rgb}{0.57, 0.36, 0.51}
\definecolor{cite}{RGB}{65,105,225}
\definecolor{cvprblue}{rgb}{0.21,0.49,0.74}
\usepackage[pagebackref,breaklinks,colorlinks,allcolors=cvprblue]{hyperref}

\def\paperID{} 
\def\confName{CVPR}
\def\confYear{2026}

\title{TimeWalker: Personalized Neural Space for Lifelong Head Avatars}

\author{
    Dongwei Pan$^{1}$
    ,
    Yang Li$^{1}$
    ,
    Hongsheng Li$^{2}$
    ,
    Kwan-Yee Lin$^{1}$ \\\\
    $^{1}$ Shanghai AI Laboratory
    \quad
    $^{2}$ CUHK
    \quad
    \\
    {\tt\small linjunyi9335@gmail.com}
}

\begin{document}

\twocolumn[{
\renewcommand\twocolumn[1][]{#1}%
\maketitle
  \begin{center}
  \includegraphics[width=1.0\textwidth]{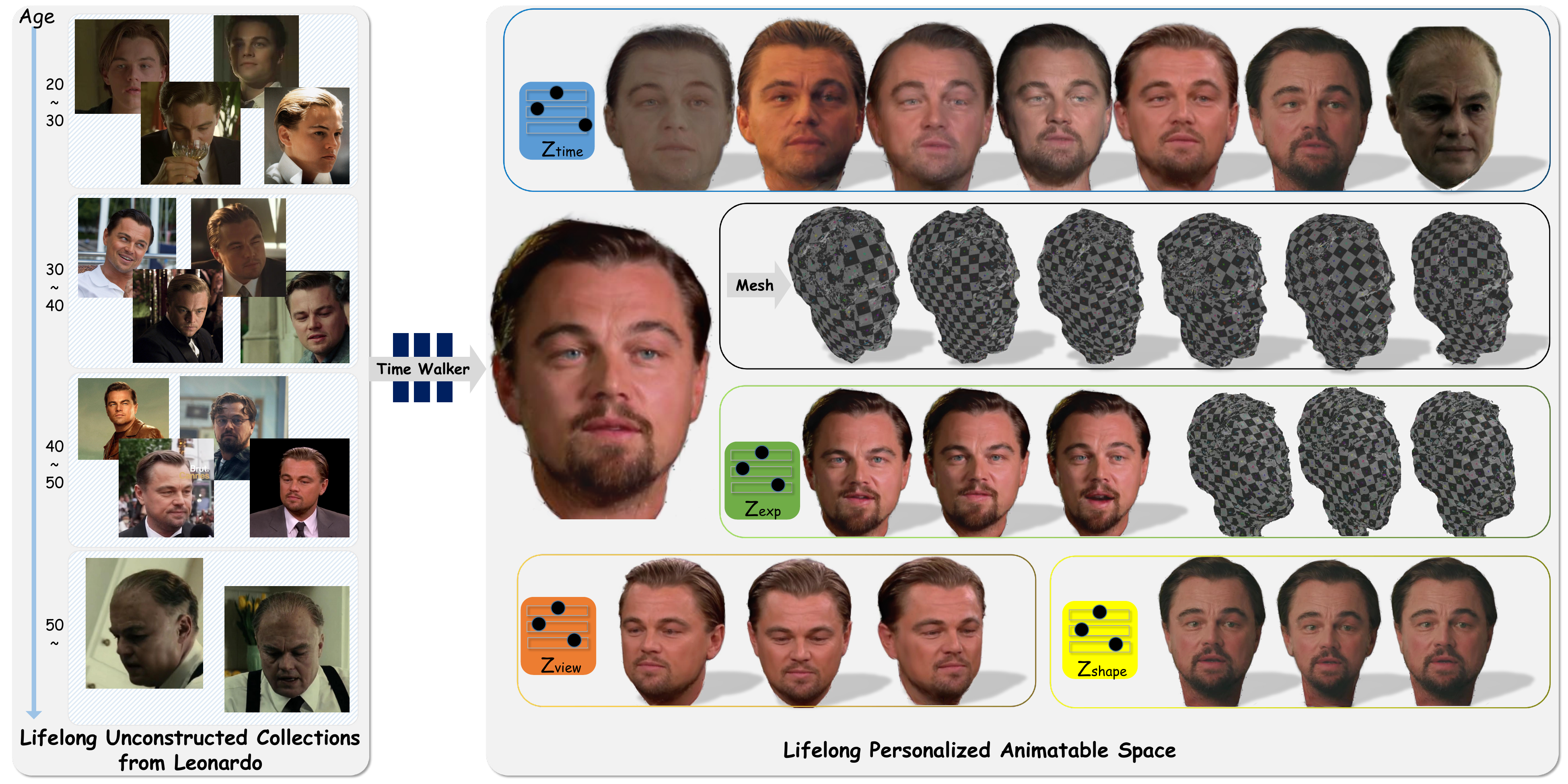}
    \vspace{-0.2cm}
  \captionsetup[figure]{hypcap=false}
  \captionof{figure}{\small{\textbf{TimeWalker.} }Given a set of unstructured data from the Internet or photo collection across years, we build a personalized neural parametric morphable model, {\textit{TimeWalker}}, towards replicating a life-long 3D head avatar of a person. With the TimeWalker, we can control and animate one's avatar in terms of shape, expression, viewpoint, and appearance across his/her different age periods. In this Figure, We show Leonardo Dicaprio's life-long avatar reconstructed and animated by our proposed model.}
  \label{fig:intro-large}
  \end{center}
}]

\begin{abstract}

We present {\textbf{TimeWalker}}\footnote{The workshop version ~\url{https://openreview.net/forum?id=3cUdmVfRb4} is a prior abstract presentation.}, a novel framework that models realistic, full-scale 3D head avatars of a person on {\textit{lifelong scale}}. Unlike current human head avatar pipelines that capture a person's identity only at the momentary level (\textit{i.e.,} instant photography, or short videos), TimeWalker constructs a person's comprehensive identity from unstructured data collection over his/her various life stages, offering a paradigm to achieve full reconstruction and animation of that person at different moments of life.  At the heart of TimeWalker's success is a novel neural parametric model that learns personalized representation with the disentanglement of shape, expression, and appearance across ages. Central to our methodology are the concepts of two aspects: $1)$ We track back to the principle of modeling a person's identity in an additive combination of his/her average head representation in the canonical space, and moment-specific head attribute representations driven from a set of neural head basis. To learn the set of head basis that could represent the comprehensive head variations of the target person in a compact manner, we propose a {\textbf{Dy}namic {\textbf{N}eural} B{\textbf{a}}sis-Blending {\textbf{Mo}}dule (\textbf{Dynamo}). It dynamically adjusts the number and blend weights of neural head bases, according to both shared and specific traits of the target person over ages. $2)$ We introduce {\textbf{D}}y{\textbf{na}}mic {\textbf2D G}}aussian {\textbf{S}}platting (\textbf{DNA-2DGS}), an extension of Gaussian splatting representation, to model head motion deformations like facial expressions without losing the realism of rendering and reconstruction of full head. DNA-2DGS includes a set of controllable 2D oriented planar Gaussian disks that utilize the priors from a parametric morphable face model, and move/rotate with the change of expression.  Through extensive experimental evaluations, we show TimeWalker's ability to reconstruct and animate avatars across decoupled dimensions with realistic rendering effects, demonstrating a way to achieve personalized ~``time traveling'' in a breeze. Project page:\textcolor{pink}{~\url{https://timewalker2025.github.io/timewalker.github.io/}}.

\end{abstract}

\section{Introduction}
~``{\textit{As we grow older, our life stories become our identity. This ongoing narrative of the self is constructed from the past and anticipated future.}'' -- James E. Birren, Psychologist, 1996. 

What forms a person's identity?  In the realm of Computer Graphics and Vision, researchers have traditionally considered an individual's shape as invariant, assuming it to represent the person's identity.  This assumption has driven significant advancements in human faces and head modeling over decades. From classic 3DMMs (3D morphable models)~\cite{3dmm} and its follow-up line of work~\cite{FLAME:SiggraphAsia2017,paysan20093d}, to current advanced neural representations for head modeling~\cite{9577855,giebenhain2023nphm,qian2023gaussianavatars,hong2022headnerf}, 3D head avatars become increasingly lifelike, serving as {\textit{momentary}} replicas of humans.

However, sociology and psychology provide a different perspective to answer the question of identity formation -- the fields emphasis the idea that a person’s identity is shaped by a continuous process influenced by various lifestages and experiences of self over time, rather than a single moment.
Since the $1960s$, sociologists and psychologists (\textit{e.g.,}\cite{Identity_development};~\cite{kro}) have recognized the intrinsic value of {\textit{lifelong}} construction of self-identity, and have developed several cornerstone theories upon this basis. 

Motivated by the critical identity formation gap, in this work, we aim to explore a paradigm of {\textbf{modeling 3D  head avatars on a person's lifelong scale}}. This new problem definition introduces three major challenges to head avatar modeling:  {\textbf{$1)$Long-horizon Identity Consistency Preserving.}} The life-long modeling breaks the long-lasting assumption of the shape invariant of a person. As the musculoskeletal structure changes with age growth, a person's facial/head shape could change significantly over his/her different lifestages. Besides, aesthetic transformation and unique life experiences also stimulate significant physical changes like facial texture features, hairstyle, accessories, and motion behavior of a person, compounding the difficulties of learning effective representation to preserve identity. {\textbf{$2)$ Limited Data Quantity and Quality for Each Lifestage.}} In prior research, specific momentary data is required with either ~``standard inputs'' ({\textit{i.e.,}} high-quality front-view images for 3D-aware head generation/editing~\cite{DBLP:conf/wacv/RaiGPVTAKPT24}), or sufficient geometric cues ({\textit{i.e.,}}multi-view capture systems~\cite{pan2024renderme,kirschstein2023nersemble,yang2020facescape,yu2020humbi}, depth sensors~\cite{livingstone2018ryerson,cosker2011facs}, or short video sequences~\cite{zielonka2023instant,Gafni_2021_CVPR} for 3D/4D head avatar reconstruction). In contrast, it is infeasible to capture lifelong data of a person with sufficient geometric cues, or a unified frontal camera view for each lifestage. Oftentimes, one's lifetime is recorded through unstructured image collection, with extremely uneven data volume and viewpoints over different moments. Such data poses a significant challenge to high-fidelity 3D head avatar modeling in both appearance-realistic and geometry-plausible aspects. {\textbf{ $3)$ Explicit-controlled Animation in Full Scales.}} Aside from lifelike reconstruction, a lifelong avatar is also expected to be controllable in terms of expression, shape, viewpoint, headpose, and appearance across a person's different age periods. Fueled by the disentangled space of parametric morphable head models like FLAME~\cite{FLAME:SiggraphAsia2017} and the expressiveness of neural field representation~\cite{mildenhall2020nerf,muller2022instant}, previous arts of 3D head avatar animation~\cite{zheng2022avatar,Gafni_2021_CVPR, zheng2022pointavatar,zielonka2023instant,qian2023gaussianavatars,kabadayi2023ganavatar} could support head animation at different scales with vivid details. However, none of these methods could manipulate {\textit{moments of life}}.  How to learn a personalized disentangled space from highly unstructured data, that enables full-scale control without losing realism is yet unknown.  

We present {\textbf{TimeWalker}} -- a baseline solution that tackles the above challenges for lifelong head avatar modeling from two aspects: $\textbf{1)}$ To get invariant identity representation, and overcome the limited data issues, we track back to the principle of the classic 3DMMs, where a specific 3D mesh can be approximated by a mean template with shape/expression's main modes of variation in an additive combination manner. Analogously, TimeWalker models a person's each lifestage by the additive combination of a shared representation in canonical space, and a set of neural head basis. The former serves as the invariant, {\textit{i.e.,}average head representation}, of the person's identity across different ages. The latter encodes the main modes of variation of moment-specific head attributes via a {\textit{Dynamic Neural Basis-Blending Module (Dynamo)}. The personalized space parameterized by these two components could provide both geometric and appearance priors to life moments with rare data. $\textbf{2)}$ To achieve full-scale control while ensuring realism, we introduce {\textit{Dynamic 2D Gaussian Splatting (DNA-2DGS)} upon the additive combination framework. It extends static Gaussian splatting representation to a set of animatable Gaussian Surfels for human heads via two deformation guidelines - the deformation fields driven from the neural head basis, and inverse warping operation rooted from FLAME expression coefficients. These ensure the surfels could be animated properly without losing geometric realism caused by FLAME's fixed topology, or explicit control problems caused by implicit neural head basis representation. With above designs, TimeWalker can learn a disentangled personalized neural space (Fig.~\ref{fig:intro-large}) only from one's lifelong {\textit{unstructured}} image collection. 

To enable modeling heads on lifelong scale, a substantial dataset of the same individual at different time points is necessary. However, this requirement is hard to fulfill with current open-source datasets since they neglect the lifelong human concept. Thus, we construct TimeWalker-1.0, a large-scale head dataset comprising $40$ celebrities' lifelong image collection sourced from various Internet data. It contains over $2.7$ million individual frames, with each identity consisting of $15K-260K$ frames and diverse age and head pose distributions. In experiments, we show TimeWalker's ability to reconstruct and animate avatars across decoupled dimensions with realistic rendering effects. Then, we demonstrate the superior rendering and geometry reconstruction quality by comparing our models to SOTA momentary 3D head models. We also demonstrate our designs' effectiveness with ablation studies. Finally, we show our model's potential benefits to downstream applications like 3D Editing.

\section{Related Works}
\label{2.related} 
\noindent\textbf{Personalized Head Avatars.} 3DMM~\cite{3dmm}, as the foundational work in 3D human head modeling, constructs a generic {\textit{head}} space via linear combination of a mean template mesh and low-dimensional linear subspaces of shape and expression from PCA. Subsequent research extends it to personalized head mesh modeling. For instance, ~\cite{PersonalizedFaceModeling} predicts personalized corrections on a 3DMM prior to obtain user-specific expression blendshapes and dynamic albedo maps. ~\cite{DBLP:journals/pami/ZhuYHLWL23} learns personalized face details via multi-view image fusion from virtually rendered multi-view input images. 
Recent advancements have extended focus to creating animatable personal {\textit{head}} avatars with realistic rendering. Methods like NerFace~\cite{Gafni_2021_CVPR} and IM Avatar~\cite{zheng2022avatar} leverage FLAME~\cite{FLAME:SiggraphAsia2017} expression coefficients to drive neural scene representation networks, implicitly representing head avatars from monocular video inputs. INSTA~\cite{zielonka2023instant} enhances training speed and enables avatar control through a mesh-based warping field. Efforts like GaussianAvatars~\cite{qian2023gaussianavatars}, FlashAvatar~\cite{xiang2024flashavatar} and other recent studies  ~\cite{xu2024gaussian,shao2024splattingavatar} leverage 3DGS's representational capabilities. They associate Gaussian points with a 3D parametric model or head mesh, generating personalized head avatars by using expression parameters as conditions for Gaussian offsets. PAV~\cite{caliskan2025pav} shares a conceptual similarity with our approach in modeling variations of a person.  Building upon INSTA, it extends to support multiple appearances of an individual from unstructured video data through the use of appearance embeddings. In contrast, TimeWalker diverges by emphasizing an interpretable, scalable, and steerable framework for constructing lifelong personalized spaces, offering explicit control over multiple-scale animations and preserving identity consistency.

\newcommand{\GS}[0]{\textbf{\color{SpringGreen}GS}}
\newcommand{\N}[0]{\textbf{\color{Brown}N}}
\newcommand{\Tri}[0]{\textbf{\color{RedViolet}Tri}}

\begin{table}[]
\begin{center}
\caption{\small{\textbf{TimeWalker}  enables preserving identity consistency in the long-horizon time spectrum (\textit{Lifelong Replica}), with explicit-controlled animation at full scale ({\textit{Animation-Expression/Shape}}). It also supports surface reconstruction and produces dynamic mesh efficiently under sparse view observations for each life stage (\textit{Mesh Reconstruction}), without losing rendering realism (\textit{High-Fidelity Rendering}). \GS: Gaussian Splatting, \N: Nerf-based, \Tri: Tri-plane. {\textbf{All the mentioned methods are compared in our experiments.}} }}
\resizebox{0.5\textwidth}{!}{
\begin{tabular}{cccccccc}
\toprule
\multirow{2}*{} & \multirow{2}*{\makecell{Lifelong\\Replicas}} & \multicolumn{2}{c}{\makecell{Animation}} & \multirow{2}*{\makecell{Mesh\\Reconstruction}} &  \multirow{2}*{\makecell{High-Fidelity\\Rendering}} & \multirow{2}*{\makecell{Open \\ Source}} & \multirow{2}*{\makecell{Base Repre-\\sentation}} \\
 & & Expression & Shape & & & & \\
\midrule
Gaussian Surfels~\cite{Dai2024GaussianSurfels}    & \textcolor[HTML]{D092A7}\xmark & \textcolor[HTML]{D092A7}\xmark & \textcolor[HTML]{D092A7}\xmark & \textcolor[HTML]{A5B592}\cmark & \textcolor[HTML]{A5B592}\cmark & \textcolor[HTML]{A5B592}\cmark & \GS \\
INSTA~\cite{zielonka2023instant}            & \textcolor[HTML]{D092A7}\xmark & \textcolor[HTML]{A5B592}\cmark & \textcolor[HTML]{A5B592}\cmark & \textcolor[HTML]{A5B592}\cmark & \textcolor[HTML]{A5B592}\cmark & \textcolor[HTML]{A5B592}\cmark & \N \\
FlashAvatar~\cite{xiang2024flashavatar}         & \textcolor[HTML]{D092A7}\xmark & \textcolor[HTML]{A5B592}\cmark & \textcolor[HTML]{D092A7}\xmark & \textcolor[HTML]{D092A7}\xmark & \textcolor[HTML]{A5B592}\cmark & \textcolor[HTML]{A5B592}\cmark & \GS \\
GANAvatar~\cite{kabadayi2023ganavatar}          & \textcolor[HTML]{D092A7}\xmark &  \textcolor[HTML]{A5B592}\cmark & \textcolor[HTML]{D092A7}\xmark & \textcolor[HTML]{D092A7}\xmark & \textcolor[HTML]{A5B592}\cmark & \textcolor[HTML]{A5B592}\cmark & \Tri \\
GAGAvatar~\cite{chu2024generalizable}           & \textcolor[HTML]{D092A7}\xmark &  \textcolor[HTML]{A5B592}\cmark & \textcolor[HTML]{D092A7}\xmark & \textcolor[HTML]{D092A7}\xmark & \textcolor[HTML]{A5B592}\cmark & \textcolor[HTML]{A5B592}\cmark & \GS \\
GPHM~\cite{xu2024gphmv2}           & \textcolor[HTML]{D092A7}\xmark &  \textcolor[HTML]{A5B592}\cmark & \textcolor[HTML]{D092A7}\xmark & \textcolor[HTML]{D092A7}\xmark & \textcolor[HTML]{A5B592}\cmark & \textcolor[HTML]{D092A7}\xmark & \GS \\
PAV~\cite{caliskan2025pav}   & \textcolor[HTML]{A5B592}\cmark &  \textcolor[HTML]{A5B592}\cmark & \textcolor[HTML]{A5B592}\cmark & \textcolor[HTML]{A5B592}\cmark & \textcolor[HTML]{A5B592}\cmark & \textcolor[HTML]{D092A7}\xmark & \Tri \\
\midrule
\textbf{TimeWalker (ours)}   &  \textcolor[HTML]{A5B592}\cmark &  \textcolor[HTML]{A5B592}\cmark &  \textcolor[HTML]{A5B592}\cmark & \textcolor[HTML]{A5B592} \cmark & \textcolor[HTML]{A5B592} \cmark & \textcolor[HTML]{A5B592}\cmark & \GS \\
\bottomrule
\end{tabular}%
}

\label{tab:cmp_table}
\end{center}
\vspace{-0.65cm}
\end{table}

By harnessing generative prior, DreamBooth~\cite{ruiz2023dreambooth} and Lora~\cite{hu2021lora} customize diffusion models from multiple images of a particular subject to produce personalized outcomes. However, these approaches are limited to static 2D results, lacking 3D consistency and animation capability. Other works like DiffusionAvatars~\cite{kirschstein2023diffusionavatars} and GANAvatar~\cite{kabadayi2023ganavatar} utilize generative models to create personalized head avatars. The former fine-tunes ControlNet~\cite{zhang2023adding} with NPHM~\cite{giebenhain2023nphm} features. The later distill EG3D~\cite{chan2022efficient} to single appearance. 
Focusing on one-shot talking head generation, methods like~\cite{ye2024real3d,li2024generalizable,chu2024generalizable,deng2025portrait4d} could achieve effective head reconstruction and animation, benefiting from the generative priors/pretrained vision models to extract fundamental human head features and 3D representation for geometric reasonability.
While these pipelines excel at face reenactment, their momentary representations limit their ability to address the challenge of modeling personalized spaces over a lifelong scale. Our pipeline takes a step in this direction with a baseline solution, enabling the construction of a lifelong replica. A comparison between TimeWalker and representative methods is shown in Tab.~\ref{tab:cmp_table}.

\noindent\textbf{Neural Representation for Static Reconstruction.} In contrast to traditional explicit reconstruction methods like meshes, point cloud and voxel grids, neural representations like NeRF~\cite{mildenhall2021nerf}, show promise with high-fidelity rendering~\cite{barron2022mip}, efficient training~\cite{muller2022instant,yu2021plenoctrees}, and mobile deployment~\cite{chen2023mobilenerf}. These models leverage differentiable rendering to refine parameters and minimize overfitting. 
Recent enhancements introduce explicit structures to boost rendering performance and training efficiency: InstantNGP~\cite{muller2022instant} adopts a multi-resolution hashgrid to streamline scene feature storage and expedite training. 3DGS~\cite{kerbl3Dgaussians} uses explicit Gaussian Splatting for rendering, achieving fast inference rates without network reliance. Gaussian Surfels~\cite{Dai2024GaussianSurfels}, refines Gaussian kernels for depth inconsistency problems rooted in 3DGS, results in high-quality static mesh reconstruction and realistic rendering under sparse view conditions. Our work extends this representation to dynamic human head modeling, enabling effective dynamic avatar animation. Refer to Sec.~$1$ in Appendix for more discussion.

\section{TimeWalker}\label{3.timewalker}
{
  \begin{figure*}[t]
  \centering
    \vspace{-0.5cm}
  \includegraphics[width=0.99\textwidth]{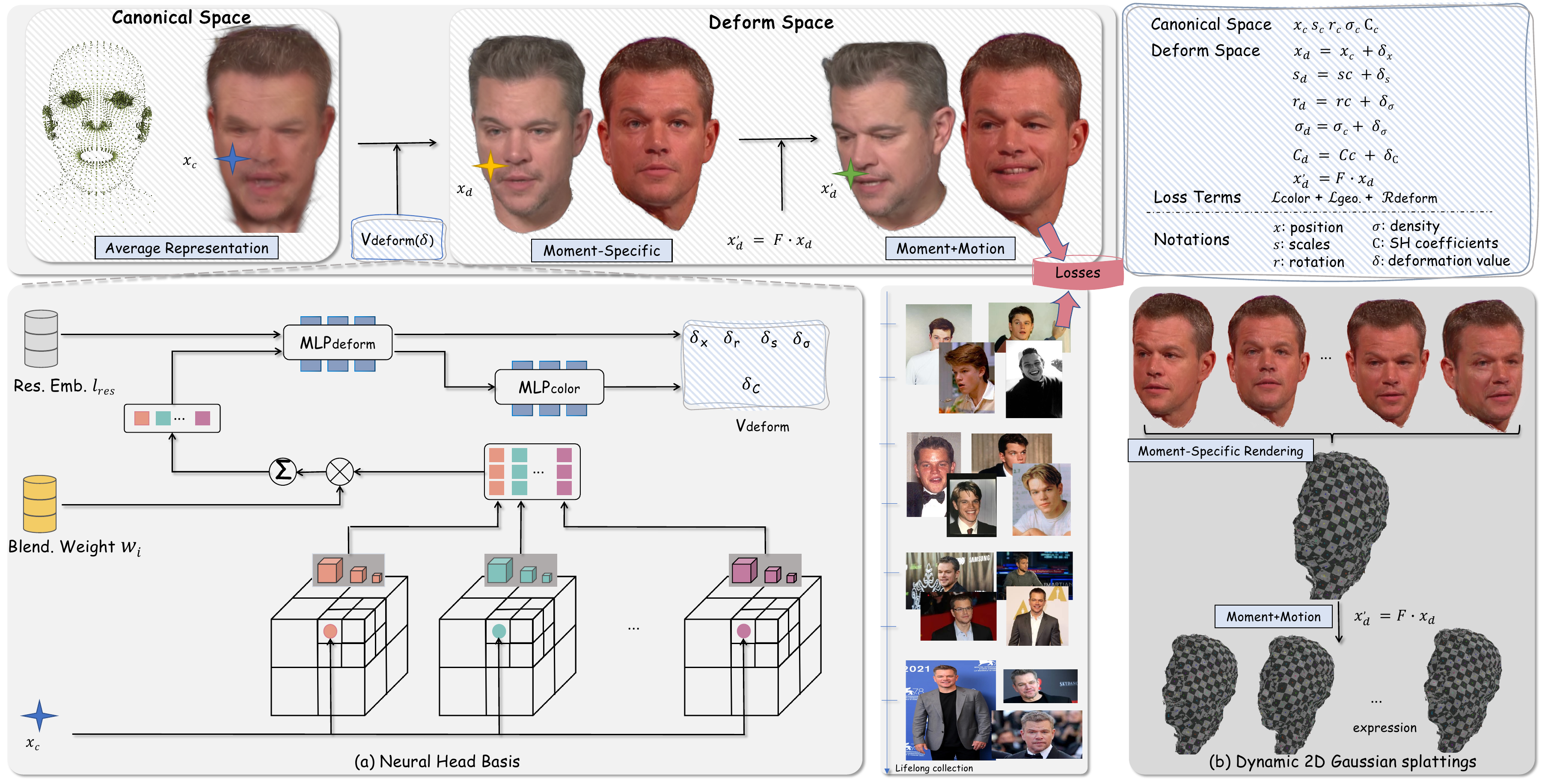}
  \vspace{-0.3cm}
  \caption{\small{\textbf{Method Overview.} TimeWalker constructs a lifelong scale 3D avatar from unstructured photo collections spanning years, maintaining realism and animation fidelity. The model is rooted in the principle of linear combination space, and innovated into an interpretable, scalable, and steerable neural personalized feature space with two key components: Dynamic Neural Basis-Blending Module (Dynamo) and Dynamic 2D Gaussian Splatting (DNA-2DGS). Concretely, Gaussian Surfels initialized in canonical space using the FLAME template represent an individual’s average head. Neural Head Basis deformations model moment-specific representations, while DNA-2DGS applies inverse warping operation driven by FLAME parameters to capture expressions and movements, generating multi-dimensional head avatars (\textit{i.e.,}{\textit{moment-motion} in deform space).
  }}}
  \label{fig:overview}
  \vspace{-0.4cm}
  \end{figure*}
}

Our work targets to construct a 3D head avatar of a person towards a lifelong scale, as opposed to current trends that reconstruct and animate a person at the momentary level. 



To address the challenges this new task brings, {\textit{\textbf{we introduce a novel neural parametric model (Fig.~\ref{fig:overview}) that captures an average representation of a person’s identity in canonical space, and extends to moment-specific head attributes through a set of Neural Head Bases.}} In Sec.~\ref{3.1}, we briefly overview 2D Gaussian representation~\cite{Dai2024GaussianSurfels}, which forms the foundational representation of our pipeline, ensuring high-fidelity rendering and dense meshing for the base frame. In Sec.~\ref{3.2}, we detail how our Personalized Neural Parametric Model constructs a personalized neural feature space via a Linear Combination Space formulation. The model leverages a Linear Combination Space formulation to handle the complexities introduced by different lifestages through Neural Head Basis, while maintaining a shared canonical representation. Then, we explore the geometric representation behind Personalized Neural Parametric Model in Sec.~\ref{3.3}-- how our DNA-2DGS, a dynamic extension of 2DGS, fosters rendering and drives the dense mesh with different motion signals. In Sec.~\ref{3.4}, we introduce training process, and discuss how we build a lifelong personalized space with disentangled control along different dimensions in Sec.~\ref{3.5}. 

\subsection{Preliminaries}\label{3.1}
Gaussian Surfels (2DGS) extends Gaussian Splatting~\cite{kerbl3Dgaussians} via reducing one dimension and transforms Gaussian ellipsoids into Gaussian ellipses. Specifically, a scene is depicted by a set of unconstructed Gaussian kernels with attribute $\left\{ \mathbf{x}_i,\mathbf{r}_i,\mathbf{s}_i,\mathbf{\sigma}_i,\mathbf{\mathcal{C}}_i \right\}_{i\in \mathcal{P}}$, where $i$ is the index of each Gaussian kernel, and $\mathbf{x}_i \in \mathbb{R}^3$, $\mathbf{r}_i \in \mathbb{R}^4$, $\mathbf{s}_i \in \mathbb{R}^3$, $\mathbf{\sigma}_i \in \mathbb{R}$ and $\mathbf{\mathcal{C}}_i \in \mathbb{R}^k$ respectively denotes the center position/rotation/opacity/spherical harmonic coefficients of each Gaussian's kernel. 2D Gaussian kernel could be obtained by flatten the 3D Gaussian's rotation on {$z$}-axis. By blending all Gaussian kernels via depth-ordered rasterization, a pixel's color $\tilde{C}$/normal value $\tilde{N}$/depth value $\tilde{D}$ can be obtained. Unlike 3DGS uses volumetric Gaussians that can blur depth boundaries, especially in areas with complex or discontinuous surfaces, 2DGS are surface-conforming primitives that project directly onto  image plane, ensuring better local depth and normal blending processes. We represent human head upon 2DGS. Notably, 2DGS, built for static reconstruction, would encounter a significant challenge in generating dynamic mesh sequences, due to time-consuming nature of post Poisson Meshing~\cite{kazhdan2013screened}. Thus, we surmount this limitation within framework designs. 
\subsection{Personalized Neural Parametric Model} \label{3.2}

\subsubsection{Neural Linear Combination Space} \label{3.2.1}

In the quest to pioneer lifelong personalized animation spaces, our primary objective is to enable precise and controlled animation of individuals on a comprehensive scale. Leveraging linear combination suits for this context, as it offers flexibility and universality in constructing compact representations through the combination of base vectors that encapsulate core variations. It provides theoretical foundation for constructing a wide range of shapes, textures, and expressions even from limited data. Besides, it simplifies optimization and computation, ensuring consistent and predictable results. This concept traces its origins to traditional 3D Head Morphable Models, where a head is modeled by linearly combining expression and shape parameters with their corresponding basis functions: 
\begin{align}
\mathbf{M}(\alpha, \beta, \gamma) = \left( \bar{S} + \sum_{i=1}^{N_s} \alpha_i \mathbf{s}_i, \ \bar{T} + \sum_{i=1}^{N_t} \beta_i \mathbf{t}_i, \ \bar{E} + \sum_{i=1}^{N_e} \gamma_i \mathbf{e}_i \right)
\end{align}{\footnote{Where \( M(\alpha, \beta, \gamma) \) is the full model (shape, texture, expression), \( \bar{S}, \bar{T}, \bar{E} \) are the mean shape, texture, and expression, \( \mathbf{s}_i, \mathbf{t}_i, \mathbf{e}_i \) are the basis vectors, \( \alpha_i, \beta_i, \gamma_i \) are the corresponding coefficients, and \( N_s, N_t, N_e \) are the number of components for shape, texture, and expression, respectively. Please refer to the original paper for more details~\cite{3dmm,FLAME:SiggraphAsia2017}.}. Building upon this foundation, we extend the concept to a neural linear combination space, integrating multiple animation dimensions in an additive, learnable, and scalable manner, as formulated in Eq~\ref{eq:last}. Particularly, as illustrated in Fig.~\ref{fig:overview}, to characterize the average head representation of individuals, we initialize a set of 2DGS in canonical space based on FLAME template. With linear addition of the deformation value produced from Neural Head Basis (detailed in Sec.~\ref{3.2.2}), the canonical 2DGS could be deformed from the average representation to a specific lifestage (\textit{i.e.,} moment-specific representation in the Figure). This process facilitates the modeling of a neutral head at any life moment, and moreover, within our framework, this is achieved without the need for direct supervision. Sequentially, to enable motion-based modeling (\textit{e.g.,} expression changes) and dynamic avatar animation, we further utilize inverse warping operation rooted from FLAME parameters (Sec.~\ref{3.3.1}), to drive the moment-motion modeling. Along with analysis-by-synthesis training (Sec.~\ref{3.4}), these designs allow creating multi-dimensional realistic head avatars in a breeze.

\subsubsection{Neural Head Basis} \label{3.2.2}

With the assumption that any lifestage of one person can be linearly blended from his/her several key characteristic variations across lifestages, we now introduce how to capture and learn these variations within the linear neural feature space from the design of the Neural Head Basis, and how to obtain a {\textit{compact}} set of the bases from our \textit{Dynamo} module. Suppose the number of head basis is $N$, for a point $\mathbf{x}_{c}$ picked from the canonical space, its feature at a specific lifestage could be formed as a linear combination of $N$ neural head basis $\HG$: $\mathbf{f}(\mathbf{x}_{c}) = \sum_{i=1}^N \omega_{i}\HG_i(\mathbf{x}_{c})$, where~$\omega_{i}$ is the learnable blending weight for each neural head basis.  To store the features compactly, we utilize multi-resolution Hashgrid~\cite{muller2022instant,kirschstein2023nersemble}, a hashmap-based cubic structure that 
 enables the learnable features stored in a condensed form. When querying features for~$\mathbf{x}_{c}$, the hashgrid looks up nearby features at various scales $J$, and cubic linear interpolation is applied to determine the final feature corresponding to the location. Thus, for each $H_{i}$, it is constructed by $\HG_{i}=\text{C-LinearInterp } (\{\ h^i_{j}\}_{j=1}^J)$.

\noindent\textbf{Dynamic Neural Basis-Blending Module (Dynamo).}  {\textit{How to set the number of head basis?}} One intuitive way would be -- assigning a fixed amount that is equivalent to the number of appearances in the character's data, with each basis independently learning the character's features for a specific lifestage. 
However, this idea is inefficient and redundant in the feature space, as a person’s appearance evolves over their lifetime, their core characteristics that define their identity remain fundamentally consistent. Even with temporary alterations like heavy makeup or accessories, we can still recognize individuals by their underlying features. Besides, we expect the feature space could be interpretable, scalable, and controllable. Thus, {\textit{our goal for the neural basis is to capture these deeper characteristics rather than solely memorizing superficial appearances.}}
To this end, we introduce {\textit{Dynamo}} to dynamically adjust the number of bases during the learning process of blending weight and hashgrids. We begin by initializing a set of learnable blending weights $\{\omega\}_{i=1}^N$ and grids $\HG$, which align with the number of lifestages of the target person.
During learning, if $\{{Indicator}(\omega^k_{i}<\kappa)=1\}$ consistently reveals that a hashgrid's weight falls below lower a preset threshold $\kappa$ across modeling multiple lifestages $k$ at an iteration $Q$ with sampled iteration interval $q$, this signals that this grid is not effectively learning meaningful features. In response, we deactivate that hashgrid. By the end of training process, this pruning strategy ensures that each remaining Neural Head Basis is efficiently capturing core, identity-defining features across lifestages. Additionally, it reduces memory requirements for storing these features, leading to a more efficient and scalable representation.

\noindent\textbf{Residual Embedding.}
Using {\textit{Dynamo}} outlined above, we obtain a collection of feature embeddings that compactly capture the subject's moment-specific attribute. As global compensation for each lifestage, we introduce a set of residual embedding $\mathbf{l}_\mathbf{res}$, which are concatenated with the blended features $\mathbf{f}(\mathbf{x}_{c})$. The concatenated features are then forwarded into a $\text{MLP}_{\text{deform}}$ to derive position $\mathbf{x}$, rotation $\mathbf{r}$, scale $\mathbf{s}$, and opacity $\mathbf{\sigma}$  deformations of the Gaussian kernel, as well as another feature vector which is subsequently passed through a $\text{MLP}_{\text{color}}$ to generate the SH coefficients $\mathcal{C}$ deformation:
\begin{align}
    \mathbf{\delta}_\mathbf{x}, \mathbf{\delta}_\mathbf{r}, \mathbf{\delta}_\mathbf{s}, \mathbf{\delta}_\mathbf{\sigma}, \mathbf{f}_{\text{deform}}(\mathbf{x}_{c}) &= \text{MLP}_{\text{deform}}(\mathbf{f}(\mathbf{x}_{c}), \mathbf{l}_\mathbf{res}) \\
    \mathbf{\delta}_\mathbf{C} &= \text{MLP}_{\text{color}}(\mathbf{f}_{\text{deform}}(\mathbf{x}_{c})),
    \label{eq:MLP}
\end{align}
The network learns Gaussian attributes' deformations $\mathbf{\delta}$, which are then additively combined with the Gaussian average in canonical space. It yields the character's moment-specific features:
\begin{align}
    [\mathbf{x}_{d}, \mathbf{r}_{d}, \mathbf{s}_{d}, \mathbf{\sigma}_{d}, \mathbf{C}_{d}] = [\mathbf{x}_{c}, \mathbf{r}_{c}, \mathbf{s}_{c}, \mathbf{\sigma}_{c}, \mathbf{C}_{c}] + [\mathbf{\delta}_\mathbf{x}, \mathbf{\delta}_\mathbf{r}, \mathbf{\delta}_\mathbf{s}, \mathbf{\delta}_\mathbf{\sigma}, \mathbf{\delta}_\mathbf{C}].
    \label{eq:feature_deform}
\end{align}
The whole combination process can be summarized below:
\vspace{-0.15cm}
\begin{equation}
\begin{aligned}
&(\mathbf{x}, \mathbf{r}, \mathbf{s}, \mathbf{\sigma})_d = 
  (\mathbf{x}, \mathbf{r}, \mathbf{s}, \mathbf{\sigma})_{c} + 
  \text{MLP}_{\text{d}}(\sum_{i=1}^N \omega_{i}\HG_i(\mathbf{x}_{c}), \mathbf{l}_\mathbf{r}) \\[-6pt]
& \mathbf{C}_{d} = \mathbf{C}_{c} + 
  \text{MLP}_{\text{c}}(\text{MLP}_{\text{d}}(\sum_{i=1}^N \omega_{i}\HG_i(\mathbf{x}_{c}), \mathbf{l}_\mathbf{r}))
\end{aligned}
\label{eq:last}
\end{equation}

\subsection{Dynamic 2D Gaussian Splattings (DNA-2DGS)} \label{3.3}

The question now is -- given a motion target/reference, how can we effectively drive the moment-specific representation to capture motion dynamics with precision, controllability and realism? This requires further deformation of the Gaussian Surfels to reflect various motion-related changes like expressions or shapes. Thus, we introduce {\textit{DNA-2DGS}}, which tackles the problem from both rendering (upper area of Fig.~\ref{fig:dynamic_rendering_meshing}) and meshing (lower area of Fig.~\ref{fig:dynamic_rendering_meshing}) aspects.

\subsubsection{Dynamic Gaussian Rendering} \label{3.3.1}

\begin{figure}[t]
    \begin{center}
    \vspace{-0.25cm}
        \includegraphics[width=0.415\textwidth]{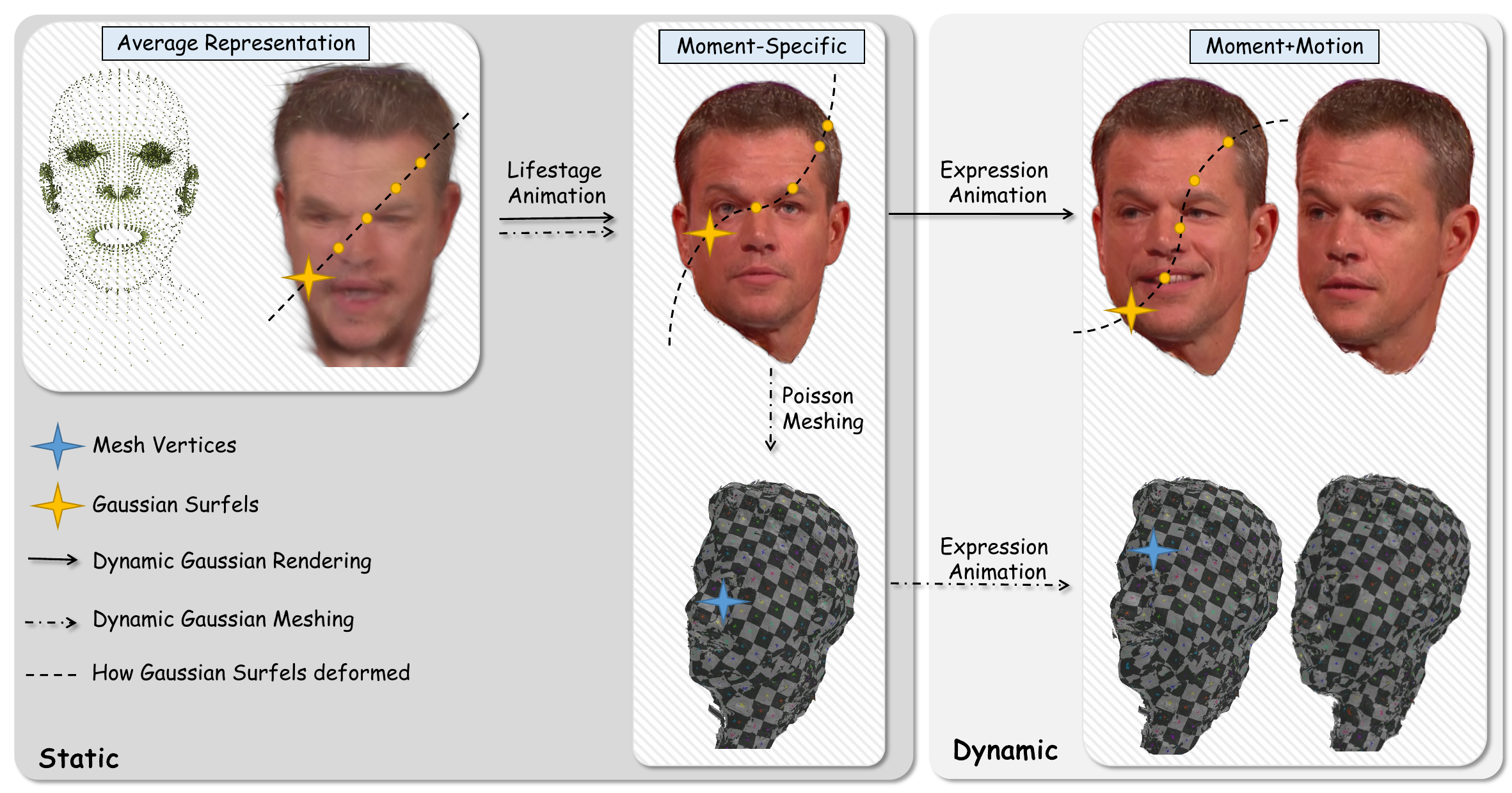}
            \vspace{-0.35cm}
        \caption{\small{\textbf{Dynamic Gaussian Rendering \& Meshing.} For  former, GS are firstly animated with lifestage and expression respectively, followed by rasterization to produce high-fidelity results. For latter, after deformed with lifestage, Poisson Meshing is used to obtain a dense mesh. Then, expression animation is performed on vertices.} }
        \label{fig:dynamic_rendering_meshing}
        \vspace{-0.55cm}
    \end{center}
\end{figure}

To realize animation, one intuitive strategy is to incorporate an additional MLP-based warping field. \footnote{In our experimental setup, we implemented this component as a naive dynamic version of Gaussian Surfels. Please refer to the results of Gaussian-Surfel++ in Appendix Sec.~$6.3$.} While this method shows proficiency in handling deformations induced by various expressions, it falls short in capturing the entire range of human head motions, particularly around eyes. Moreover, disentangling the animation of appearance and expression remains challenging when both components of the warping field are designed similarly. 

In contrast, we integrate inverse warping operation inspired by INSTA~\cite{zielonka2023instant}. 
During preprocessing, we acquire a tracked mesh $M^{def}$ through FLAME fitting and a mean template $M^{canon}$ defined within a canonical space, both sharing identical topology.
Unlike INSTA that utilizes the warping field to inversely project points from deformed space back to canonical space, we define a deformation gradient $\boldsymbol{F} \in \mathbb{R}^{4\times4}$ in the form of a transformation matrix. It projects Gaussian Surfels $\mathbf{x}_{d}$ from moment-specific static space to dynamic motion space. Concretely, for each Gaussian Surfel $\mathbf{x}_{d}$, we employ a nearest triangle search algorithm to compute $\boldsymbol{F} = \boldsymbol{L}_{def} \cdot \Lambda ^ {-1}\cdot\boldsymbol{L}_{canon} ^ {-1}$, where $\boldsymbol{L}_{canon}$ and $\boldsymbol{L}_{def}$ is Frenet coordinate system frames, and $\Lambda$ is diagonal matrix that takes scaling factor into account. Gaussian Surfels are further deformed $\mathbf{x}_{d}^{'} = \boldsymbol{F} \cdot \mathbf{x}_{d}$, guided towards specific movements. 
This process enables translation of expression/shape signals into animated outputs. The double-deformed 2DGS are then rasterized to produce final renderings.

\subsubsection{Dynamic Gaussian Meshing} 

Although vanilla 2DGS~\cite{Dai2024GaussianSurfels} enables high-quality surface reconstructions, it is primarily suited for {\textit{static}} settings, and the time-intensive nature of Poisson reconstruction process makes it impractical for mesh sequence reconstruction. These limit its applicability to dynamic head avatars.  To address these challenges, we introduce {\textit{Defer-Warping}}, an adapted Gaussian surface reconstruction strategy tailored specifically for dynamic head reconstruction. Specifically, unlike standard rendering \& meshing processes that apply deformations before meshing, we delay the motion animation after the Poisson meshing process of moment-specific representation. This allows us to render results under moment-specific conditions customized to every single lifestage, effectively eliminating motion-induced artifacts (such as artifacts on mouth regions caused by talking) and generating appearance-specific static meshes. After Poisson meshing of the static result, the delayed motion animation generates dynamic mesh sequences by directly manipulating the vertices of reconstructed mesh. This process faithfully captures motion-driven deformations (as demonstrated in Fig.~\ref{fig:ablation_mesh}).

\subsection{Training} \label{3.4}

We apply end-to-end training manner that enables the simultaneous optimization of the explicit Gaussian surfels, multiple hashgrid, feature latent, and implicit deform MLP \& color MLP. 
Before formal training, we introduce a warm-up phase where we suspend the optimization of the Neural Head Basis. This step aims to guide the Gaussian surfels in canonical space towards a mean representation.
We use densify and pruning strategies to adaptively adjust the number of Gaussians. Our total loss for whole system are:
$\mathcal{L}_{\mathrm{total}}$ consists of three parts: $(1)$ image-level supervision; $(2)$geometry level supervision; and $(3)$ regulation: 
\begin{align}
    \mathcal{L}_{\mathrm{total}} &= \mathcal{L}_{\mathrm{image}} + \mathcal{L}_{\mathrm{geometry}} + \mathcal{L}_{\mathrm{regulation}},
\end{align}
Refer to Sec.~$4$ in the Appendix for more details.
\subsection{Building a Life Long Personalized Space} \label{3.5}

\textbf{How does TimeWalker construct a lifelong personalized space? } The Gaussian Surfels in canonical space that characterize individual average representation, are additively combined with moment-specific head attribute representations driven from a set of Neural Head Basis, spanning a moment-specific head avatar. For each moment-specific avatar, we further warp Gaussian Surfels via expression or shape signals to get motion-specific head performing. By separating moment-specific deformation from motion-specific warping, we can decouple the driving of the head in multiple dimensions - lifestage, expression, shape, novel view, and {\textit{etc}}, - constructing a comprehensive, steerable personalized space. Refer to Sec.~$4$ in Appendix for details.

\noindent\textbf{What are the core benefits of learned space?} As life-long modeling inherently suffers from rare/long-tail data issues that certain stages have missing observations (like views, expression), the personalized space can effectively mitigate the task’s intrinsic data sparsity by learning a basic, common prior to represent a person's identity across stages. Thus, it can solve several challenges posed by sparse data: ${\textbf1)}$ multi-scale animation (Fig.~\ref{fig:intro-large}), ${\textbf2)}$ steerable reenactment, ${\textbf3)}$ detailed mesh reconstruction (Fig~\ref{fig:ablation_mesh}), and ${\textbf4)}$ ID-preserving rendering across ages and camera views (Sec.~$6.2$ and~$6.6$ in Appendix). With these, our method could improve downstream tasks' quality.  See Sec.~\ref{4.experiments} and Appendix for details.


\section{Experiments}\label{4.experiments}

Due to the page limits, please refer to Appendix Sec.~$6$ for more internal experiments like additional ablations, personalized space, large-viewpoint rendering, lighting control, as well as Sec.~$7$ for comparison experiments with sota multiview reconstruction methods, generative methods and one-shot head avatar methods. In main paper, we unfold our method with key comparisons with sota, reenactments, mesh comparison and key ablations.

\begin{table}[htb]
\begin{center}
\caption{\small{\textbf{Quantitative Evaluation on TimeWalker-1.0.} 
We evaluate our method using two protocols. 
The left half of columns shows {\textit{\#Protocol-1}} (1 vs 1), 
and the right half shows {\textit{\#Protocol-2}} (1 vs N). 
\colorbox{colorbest}{Pink} is best, \colorbox{colorsecond}{orange} is second best.}}

\resizebox{0.5\textwidth}{!}{
\begin{tabular}{c|cccc|ccccc}
\toprule[1.5pt]
 & \multicolumn{4}{c|}{\textbf{1 vs 1 (Protocol-1)}} 
 & \multicolumn{5}{c}{\textbf{1 vs N (Protocol-2)}} \\
\cmidrule(lr){2-5} \cmidrule(lr){6-10}
\textbf{Metric} 
& \textit{INSTA}~\cite{zielonka2023instant} 
& \textit{INSTA++} 
& \textit{Flash Avatar}~\cite{xiang2024flashavatar}
& \textit{Ours} 
& \textit{GS}~\cite{Dai2024GaussianSurfels}
& \textit{GS++} 
& \textit{INSTA}~\cite{zielonka2023instant} 
& \textit{Flash Avatar}~\cite{xiang2024flashavatar} 
& \textit{Ours} \\ 
\midrule

\textbf{PSNR$\uparrow$} 
& 20.68 
& \colorbox{colorsecond}{26.39} 
& 22.14 
& \colorbox{colorbest}{27.28} 
& 26.98 
& \colorbox{colorbest}{27.61} 
& 25.47 
& 24.90 
& \colorbox{colorsecond}{27.28} 
\\

\textbf{SSIM$\uparrow$} 
& 0.697 
& \colorbox{colorsecond}{0.879} 
& 0.771 
& \colorbox{colorbest}{0.949} 
& \colorbox{colorbest}{0.950} 
& 0.948 
& 0.860 
& 0.848 
& \colorbox{colorsecond}{0.949} 
\\

\textbf{LPIPS$\downarrow$} 
& 0.299 
& \colorbox{colorsecond}{0.139} 
& 0.267 
& \colorbox{colorbest}{0.071} 
& 0.141 
& \colorbox{colorsecond}{0.134} 
& 0.170 
& 0.165 
& \colorbox{colorbest}{0.071} 
\\

\bottomrule[1.5pt]
\end{tabular}
}

\label{tab:exp_main}
\end{center}
\end{table}

\subsection{Datasets and Evaluation Protocol}

In our experiments, if not specified, we select $5$ representative identities from our TimeWalker-1.0 dataset, each comprising $8$-$13$ lifestages with a total of $8000$-$20000$ frames, with resized to $512$x$512$. It is important to note that the number of frames for each lifestage ranges from $500$ to $3000$, generally much less than the frames in other SOTA personalized head avatar methods, which increases the challenge of this task. Refer to Sec.~$6.3$ in Appendix for detailed settings and Sec.~$5$ for details of the dataset.

We address the uneven frame distribution among appearances by allocating the last 10\% of frames from each appearance as the test set and the remaining frames as the training set.
We conduct experiments with two protocals -- In $1)$ \textbf{\textit{\#Protocol-1 (1 vs 1 Comparison)}} We train one model for one identity with multiple lifestages.  In $2)$ \textbf{\textit{\#Protocol-2 (1 vs N Comparison)}} Our model maintains the same pipeline setup as \textit{\#Protocol-1}, but for the baseline models, we train one model for each lifestage, allowing multiple models for one identity. This allows us to compare the results of our pipeline using a single model against the baseline using multiple models.
We utilize three metrics, PSNR, SSIM, LPIPS, to assess the quality of individual generated frames.

\subsection{Comparisons with State-of-the-Arts} \label{comparison_with_sota}
\label{exp:compare_sota}
\noindent\textbf{Baseline. } 
We compare TimeWalker with {\textbf{8}} representative baselines: {\textbf{1) NeRF-based methods}} (INSTA~\cite{zielonka2023instant}); {\textbf{2) 3DGS-based methods}} (FlashAvatar~\cite{xiang2024flashavatar}, Gaussian-Surfels~\cite{Dai2024GaussianSurfels}, and its extension Gaussian-Surfels++); {\textbf{3) generative-based methods}} (GANAvatar), and large-scale one-shot head avatar methods (GAGAvatar~\cite{kabadayi2023ganavatar},GPHM~\cite{xu2024gphmv2}), and {\textbf{4) personalized avatar over time scale}} (a re-implementation of \textit{PAV}~\cite{caliskan2025pav}, denoted as \textbf{INSTA++}), \textit{\textbf{Refer to Appendix Sec.7 for the detailed description of these methods and comparisons on GANAvatar~\cite{kabadayi2023ganavatar}, GAGAvatar~\cite{chu2024generalizable} and GPHM~\cite{xu2024gphmv2}.}}

{
  \begin{figure}[tb]
  \vspace{-0.2cm}
  \centering
  \includegraphics[width=0.45\textwidth]{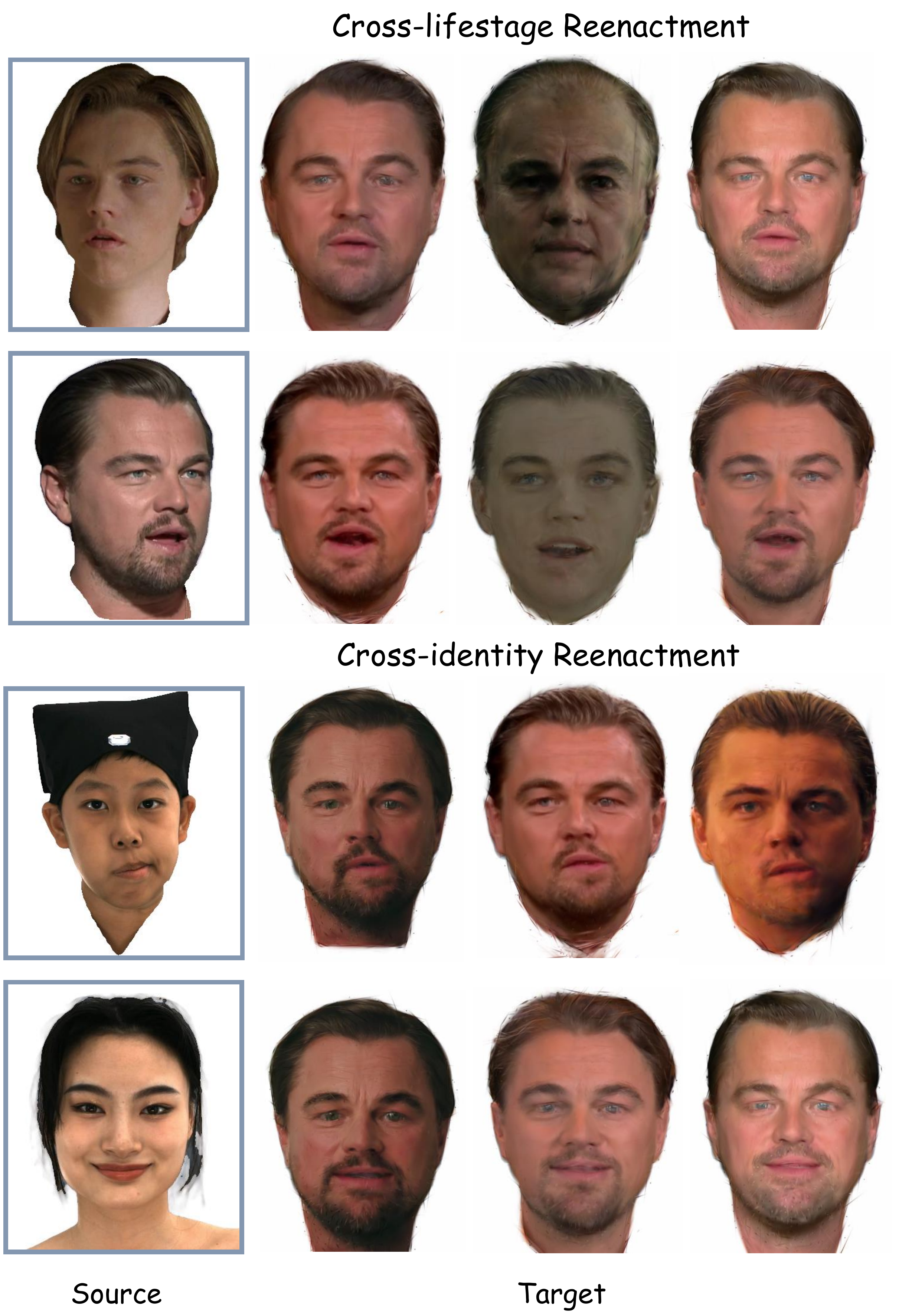}
 \vspace{-0.35cm}
  \captionsetup[figure]{hypcap=false}
  \caption{\small{\textbf{Reenactment.}  We include cross-lifestage reenactment (self-reenacment) with TimeWalker-1.0 and cross-identity reenactment with RenderMe-360~\cite{pan2024renderme}, in Leonardo personalized space. }}
  \label{fig:self_reenactment}
    \vspace{-0.6cm}
  \end{figure}
}

\noindent\textbf{Results. }Quantitatively, as shown in Tab.~\ref{tab:exp_main}, our method achieves the best results in the \textit{\#Protocol-1}, and best or second best results in the \textit{\#Protocol-2} on our TimeWalker-1.0 dataset. 
Our method significantly outperforms other baselines in both protocols when assessed using TimeWalker-1.0. Although INSTA++ shows considerable enhancements post multi-lifestage adaptation, there remains a noticeable gap compared to our approach, underscoring the necessity of meticulously designing our pipeline for creating high-fidelity lifestage replicas and highlighting the effectiveness of our design in handling multi-stage data, rather than simply relying on dataset memorization. In the 1 vs N Comparison, our results remain competitive, with lower LPIPS values.

\subsection{Reenactment}
\label{reenactment} 
Fig.~\ref{fig:self_reenactment} shows expression animation with two types of reenactments: $1)$ The {\textit{cross-lifestage reenactment} on the left showcases how head avatars from different lifestages can consistently perform the same expression, animated from a source avatar belonging to a different time period. $2)$ The {\textit{cross-identity reenactment}} on the right shows Leonardo in different lifestages are driven by unseen novel expression from RenderMe-360~\cite{pan2024renderme}. The rendering result shows that multiple head avatars generated by the same personal space from TimeWalker are able to extrapolate novel expressions.

{
  \begin{figure}
  \vspace{-0.1cm}
  \begin{center}
  \includegraphics[width=0.5\textwidth]{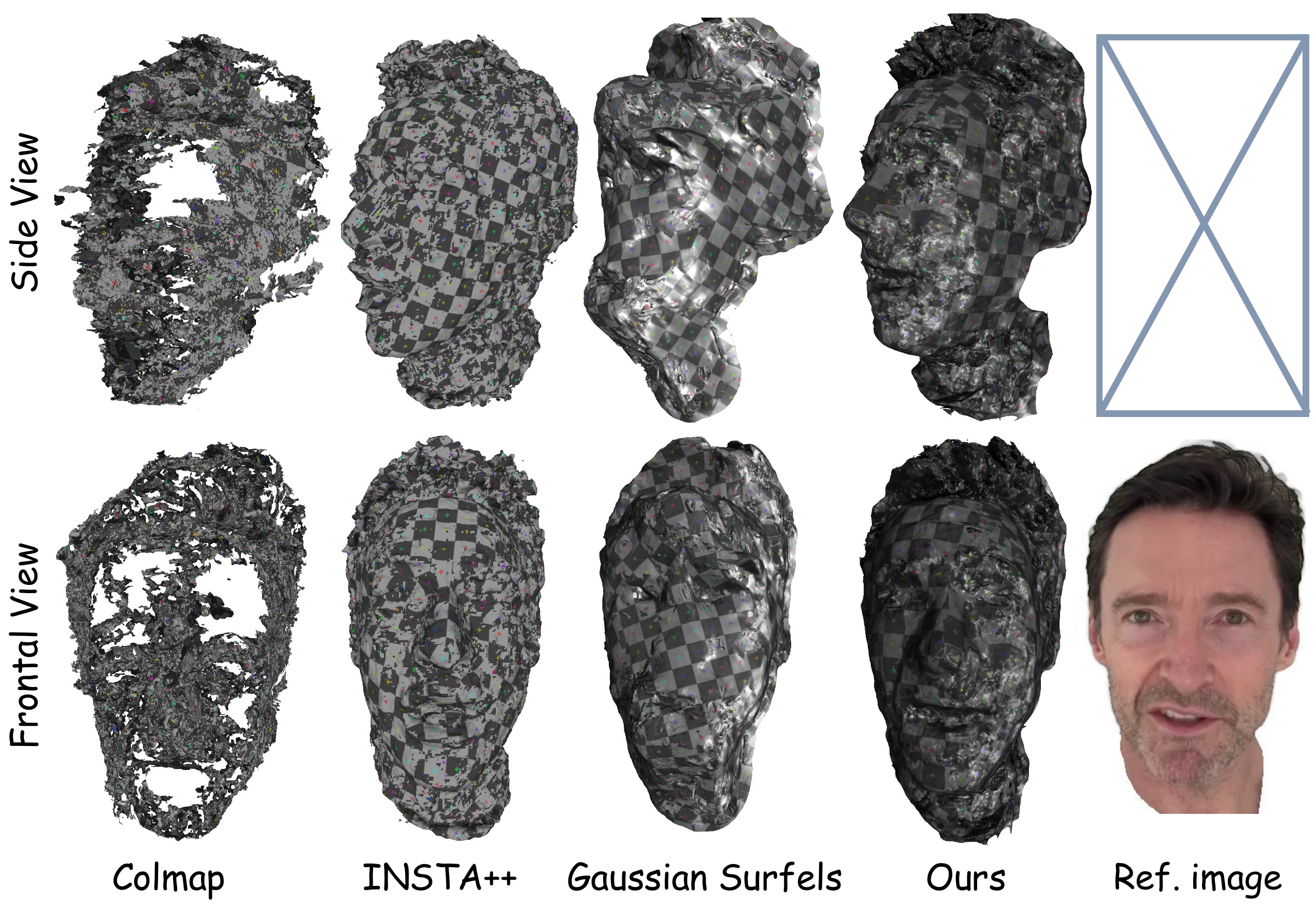}
  \vspace{-0.4cm}
  \captionsetup[figure]{hypcap=false}
  \captionof{figure}{\small{\textbf{Static Mesh Comparison.} We visualize and compare  static mesh reconstruction results from Gaussian Surfels~\cite{Dai2024GaussianSurfels} and Ours (adaptive version of GS with DNA-2DGS). We render the meshes in both frontal and side views using Blender software under identical rendering conditions. Despite these consistent settings, rendering results exhibit significant differences due to the diverse topologies of the meshes (Zoom in for details).}}
  
  \label{fig:ablation_mesh}
 \end{center}
   \vspace{-0.45cm}
  \end{figure}

}

\subsection{Mesh Comparison}
\label{mesh_comparison}







\begin{table}[]
\scriptsize
\centering
\renewcommand{\arraystretch}{0.9}
\caption{\small{\textbf{Ablation on Hashgrid.} 
\colorbox{colorbest}{Pink} is best, 
\colorbox{colorsecond}{orange} is second.}} 

\resizebox{0.47\textwidth}{!}{
\begin{tabular}{c|ccccc}
\toprule

& \textbf{w/o Dynamo} 
& \textbf{1 Hashgrid} 
& \textbf{All Hashgrid} 
& \textbf{20 Hashgrid} 
& \textbf{Ours} \\
\midrule

\textbf{PSNR$\uparrow$} 
& 21.69 
& 24.84 
& 26.86 
& 26.94 
& \colorbox{colorsecond}{27.20} 
\\

\textbf{SSIM$\uparrow$} 
& 0.767 
& 0.890 
& 0.938 
& 0.935 
& \colorbox{colorsecond}{0.941} 
\\

\textbf{LPIPS$\downarrow$} 
& 0.197 
& 0.119 
& \colorbox{colorsecond}{0.078} 
& \colorbox{colorbest}{0.077} 
& \colorbox{colorbest}{0.077} 
\\

\bottomrule
\end{tabular}
}
\label{tab:ablation}
\end{table}

To evaluate the effectiveness of our mesh reconstruction pipeline, we compare our geometric outcomes with meshes extracted from various pipelines: Colmap~\cite{schoenberger2016sfm}, INSTA++, and Gaussian Surfels~\cite{Dai2024GaussianSurfels}. 
Data from a single lifestage is provided to Colmap and Gaussian Surfels, tailored for static scene reconstruction. Conversely, our pipeline and INSTA++, designed for handling multiple lifestages, receive all data of an individual.

As demonstrated in Fig.~\ref{fig:ablation_mesh}, our pipeline successfully reconstructs the head mesh with precise shape, while the original Gaussian Surfel and Colmap fail to derive a meaningful mesh. This disparity is partly attributed to the restricted data from a single appearance and camera pose. Both pipelines struggle to recreate a plausible head mesh without prior knowledge of human head anatomy, given that monocular reconstruction is inherently an ill-posed problem. 
In contrast, our mesh reconstruction, utilizing shared representations learned data from multiple lifestages, achieves superior quality with more accurate surface details. Although INSTA++ can also reconstruct a plausible head mesh, it exhibits more noises. This comparison underscores the efficacy of our DNA-2DGS component in reconstructing head meshes from unstructured photo collections.

\subsection{Ablation Studies} 

To validate the effectiveness of TimeWalker's components, we conduct ablation experiments on $3$ individuals with respective $9$, $10$ and $13$ lifestages. We keep other settings unchanged except the ablation term. We show results about our Dynamo design here and \textbf{\textit{Refer to Appendix Sec.6 for more detailed ablations.}}

\noindent\textbf{Dynamo.}
We conduct ablation on Dynamo module in the Neural Head Basis, a key contribution that enables our pipeline to preserve the moment-specific attributes of individuals. Two protocols are used: $1)$ removing Dynamo and replacing it with the position $\mathbf{x}_{c}$ of each Gaussian surfels; $2)$ predefining and fixing hashgrid numbers (either 1, 20 or matching the number of life stages). The quantitative result in Tab.~\ref{tab:ablation} shows the necessity of Dynamo with multiple hashgrids, as reducing the hashgrid number or removing Dynamo leads to performance dropout. Interestingly, our pipeline with adaptive hashgrid number performs slightly better than the setting with all hashgrid and a large number of hashgrid(20) that exceeds the number of lifestages. This improvement demonstrates the effectiveness of our adaptation mechanism, which fosters the framework to learn compact representations of both shared and unique characteristics across different life stages, confirming that the dynamic setting achieves the best performance by balancing representational capacity and generalization.

\noindent\textbf{How Temporal Diversity Helps Single Stage.} To further evaluate the benefit of our multi-lifestage supervision, we conduct a controlled study comparing two training paradigms: Single- and Multi-Lifestage training.
As shown in Tab~\ref{multi-single}, our multi-lifestage training yields a substantial improvement in rendering quality. The core benefit of this paradigm relies on the learned prior knowledge afforded by our Dynamo and Residual Embedding modules.

\begin{table}[t]
\centering
\scriptsize
\caption{\textbf{Multi- vs Single-lifestage.}}
\vspace{-1ex}
\begin{tabular}{lccc}
\toprule
\textbf{Data} & \textbf{PSNR} $\uparrow$ & \textbf{SSIM} $\uparrow$ & \textbf{LPIPS} $\downarrow$ \\
\midrule
One stage data      
    & 23.69 & 0.959 & 0.072 \\
Full stages (Ours)
    & \colorbox{colorbest}{\strut 27.22}
    & \colorbox{colorbest}{\strut 0.955}
    & \colorbox{colorbest}{\strut 0.070} \\
\bottomrule
\end{tabular}\label{multi-single}
\vspace{-2ex}
\end{table}

\section{Conclusion}\label{5.diss} 

We propse to construct human head avatar in life-long scale. For this new task, we present TimeWalker, a baseline solution for constructing personalized spaces that maintain long-term identity consistency while enabling explicit, full-scale animation control. Rooted in the additive combination principles of classic 3DMM, our approach innovates through a neural feature-based design with two key components: a Dynamic Neural Basis-Blending model to represent the head variations in a compact manner and a Dynamic 2D GS module to construct a dynamic dense head mesh. Experiments show our method's effectiveness, and meanwhile also reveal the limitations, paving the opportunities for future research on human head avatars.


\clearpage

\appendix
\newpage

\makebox[\linewidth][c]{
  \begin{minipage}{\textwidth} 
    \centering
    \LARGE \bfseries Supplementary Material
  \end{minipage}
}
\vspace{2em}



\section{Additional Related Work}\label{related_appendix}

Since our approach constructs the personalized space from in-the-wild unstructured photos, we also discuss related developments in the research field of general 3D reconstruction from unstructured photos. Furthermore, we explore relevant research within age progression modeling as our pipeline endeavors to establish a coherent and interpretable lifelong personalized space encompassing various ages and stages of life.

\noindent\textbf{3D Reconstruction from Unstructured Photo.} 
Reconstruction from Internet photo collections has been a long-standing topic in computer vision and computer graphics. Thought-provoking research such as Photo Tourism~\cite{snavely2006photo}, Building Rome in a Day~\cite{agarwal2009building} and Skeletal Sets ~\cite{SSS-cvpr07} show great potential for applying structure from motion (SfM) algorithms on unstructured photo collections. Upon these pioneer works, multi-view stereo (MVS) algorithms~\cite{Curless,sch2016} and appearance modeling~\cite{kim16} are proposed to improve the reconstruction quality. More recently, several works~\cite{martinbrualla2020nerfw,sun2022neuconw} model the scene by grafting these ideas into the neural radiance fields~\cite{mildenhall2020nerf}. There are also a series of great works focused on reconstructing heads or heads from Internet photos. For example, in~\cite{iraface}, the author holds the premise of deriving a 3D head shape basis directly from a large amount of Internet collection, and proposes to reconstruct an arbitrary 3D head from a single view image based on the shape basis.  The subsequent work such as ~\cite{ira-steve} focuses on recovering the head from personal photo collections, and~\cite{liang2016head} aims to recover the personalized head. Still and all, not much attention has been paid to reconstructing human heads on a lifelong scale. The most related work to our project is PersonNeRF (~\cite{weng2023personnerf}), which is driven from NeRF-W~\cite{martinbrualla2020nerfw} to model personalize space for the human body across several years' data. However, this method assumes a person's body shape is roughly identical across years, which limits its scalability to lifelong settings. Beyond that, human head modeling is more challenging than body in terms of high-fidelity details and fine-grained capturing like subtle expression. 

\noindent\textbf{Age Progressing Modeling.} 
Research on simulating aging effects has been prominent in recent decades. Methods like RFA~\cite{wang2016recurrent} and IAAP~\cite{kemelmacher2014illumination} have led the way in creating average faces and transferring texture differences between age groups to model aging. GAN-based approaches like S2GAN, and Face Aging GAN,\cite{he2019s2gan, wang2018face} generate subtle texture variations across different ages. Acknowledging the importance of shape and texture in age modeling, various techniques~\cite{lanitis2002toward, suo2009compositional, suo2012concatenational, yang2016face} have emerged to address both simultaneously. Innovative diffusion-based text-to-video pipelines such as DreamMachine, Kling, and Gen3~\cite{DreamMachine,Kling,Gen3_runway} have showcased the ability to model age progression via hallucinating human-aging videos from textual cues, yielding impressive outcomes. However, these generative methods struggle to achieve explicit and comprehensive head animation (\textit{e.g.,} expressions and shape variations), and face challenges in maintaining robust 3D consistency, limiting their functionality in creating personalized spaces.

\section{Additional Preliminaries}
\label{preliminaries}

Extended from Gaussian Splatting~\cite{kerbl3Dgaussians}, Gaussian Surfels (2DGS) reduces one dimension and transforms Gaussian ellipsoids into Gaussian ellipses, while sharing the same attributes of the Gaussian kernels $\left\{ \mathbf{x}_i,\mathbf{r}_i,\mathbf{s}_i,\mathbf{\sigma}_i,\mathbf{\mathcal{C}}_i \right\}_{i\in \mathcal{P}}$. Practically, by blending all Gaussian kernels in the scene with depth-ordered rasterization, the color $\tilde{C}$, the normal value $\tilde{N}$, and the depth value $\tilde{D}$ of a pixel can be obtained by
\vspace{-0.2cm}
\begin{align} \label{eq:render}
\tilde{C}&=\sum_{i=0}^n{T_i\alpha _i\mathbf{c}_i} \\
\vspace{-0.2cm}
\tilde{N}&={\frac{1}{1-T_{n+1}}}\sum_{i=0}^n{T_i\alpha _i\mathbf{R}_i\left[ :,2 \right]} \\
\vspace{-0.2cm}
\tilde{D}&=\frac{1}{1-T_{n+1}}\sum_{i=0}^n{T_i\alpha _{i}d_i\left( \mathbf{u} \right)},
\end{align}

where $\alpha_i$ represents alpha-blending weight, and ${1}/({1-T_{n+1}})$ is a normalization scale for blending weight $T_i\alpha_i$, which is calculated with $T_i=\prod_{j=0}^{i-1}{(}1-\alpha _j)$. $\mathbf{R}_i$ is rotation matrix. $d_i\left( \mathbf{u} \right)$ represents the adjusted depth value of the center of the Gaussian kernel. Based on above formula, Gaussian Surfels are surface-conforming primitives that project directly onto the image plane, ensuring precise local depth and normal blending processes. In contrast, 3DGS uses volumetric Gaussians, which can blur depth boundaries and reduce precision, especially in areas with complex or discontinuous surfaces. Consequently, Gaussian Surfels exhibits remarkable performance even in sparse view settings, a crucial requirement for our setting.{\footnote{
More theoretical details about Gaussian Surfels and qualitative as well as quantitative comparison of mesh reconstruction between 3DGS and 2DGS can be referred to ~\cite{Dai2024GaussianSurfels}.}}


\section{Building a Life Long Personalized Space} \label{personal_space_appendix}
 
In the main paper, we introduce our approach for constructing a personalized space with full-scale animation in a disentangled manner. In this section, we first outline the core design principles that validate our lifelong personalized modeling framework, then elaborate on the realization of full-scale animation across multiple dimensions.

\subsection{Core Principles for Personalized Space}
Our framework’s capacity to sustain lifelong identity consistency is anchored in two key design choices that prioritize: {\textit{coherence over long-horizon assets}} and {\textit{robust generalization from limited and long-tail observations}}.

\noindent\textbf{Coherent Shared Latent Space with Personalized Neural Bases.}
Current reconstruction or generative methods struggle to adapt to lifelong modeling, often yielding disjointed representations or inconsistent identity preservation (See Sec~\ref{exp:compare_sota}) ,as they rely on isolated snapshot learning and assume a single canonical neutral shape rather than a personalized identity space. In contrast, our model learns a compact, shared latent space that unifies all life stages of an individual. This property is further evidenced from the number of adaptive neural bases (Tab~\ref{tab:num_lifestage}, Neural Basis), which is generally fewer than observed stages, confirming our model discovers reusable, cross-stage components rather than memorizing independent snapshots. This is critical, as such reusable components can help mitigate long-tail sparsity across life stages, enabling the model to generalize in both geometry and apperance aspects even when certain states (\textit{e.g.,} extreme expression, and unseen camera views) are underrepresented or only weakly observed, as demonstrated in Fig~\ref{fig:large_angle} here, and the mesh reconstruction experiment in the main paper.

\noindent\textbf{Temporal Generalization for Sparse Lifestage Data.}
Lifestage data is inherently sparse—there are no continuous aging recordings and typically no dense multi-view or dense expression captures. Yet our latent space acts as a {\textit{cross-stage prior}} that encodes long-term identity consistency. This enables two critical capabilities unavailable in snapshot-based models: interpolation across missing lifestage observations (generating plausible intermediate ages) and generalization to unseen poses. Both are validated by high-fidelity geometry reconstruction (Fig.~$5$ in main paper) and robust large-angle novel view synthesis (Fig.~\ref{fig:large_angle}).

\begin{table}[htbp]
    \centering
    \caption{\textbf{Number of the Lifestages and Neural Basis in selected ID.}}
    \label{tab:num_lifestage}
    \resizebox{0.4\textwidth}{!}{
    \begin{tabular}{lcc}
    \toprule
    \textbf{ID} & \textbf{Jackie Chan} & \textbf{Leonardo Wilhelm DiCaprio} \\
    \midrule
    Lifestages    & 8 & 13  \\
    Neural Basis  & 5 & 8  \\
    \bottomrule
    \end{tabular}
}
\end{table}

\subsection{Full-scale Animation}
\textbf{Lifestage.} During training, our pipeline learns different blending weights $\{\omega\}_{i=1}^N$ for data in different lifestages. After training, we can adjust these weights to drive the lifestage in a disentangled manner. Fig.~$4$ in the main paper illustrates the appearance diversity of individuals as they progress through different lifestages. This demonstrates the effectiveness of our pipeline in capturing a person's identity across different moments in their life. 

\textbf{Expression.} To achieve expression and shape changes of the character while maintaining a consistent appearance, we use a inverse warping operation inspired by INSTA~\cite{zielonka2023instant}. By manipulating expression parameters, we can update both the tracked mesh and the transformation matrix that maps from canonical space to deformation space. This enables us to achieve the desired expression-based warping. 

\textbf{Shape.} As the FLAME mesh can be driven by expression and shape parameters in a disentangled manner, our head avatar can also be animated by shape with the same approach as expression. 

\textbf{Novel view.} The Gaussian Splatting, as a type of 3D representation, can be rendered with arbitrary camera pose. 

The creation of a lifelong personalized space through our pipeline yields core benefits. 
The fine-grained designed personalized neural parametric model showcases its capacity to generate a comprehensive personalized space with multi-scale animation and steerable reenactment. 
Through a linear combination space in constructing the neural parametric model, incorporating mean representation and lifestage-specific features, we achieve remarkable identity preservation across various life stages. 
The efficient compensation mechanism across different life stages facilitates interpolation for previously unobserved side profiles, thereby augmenting the capacity for identity preservation across various camera views.
By extending the 3DGS to dynamic 2D Gaussian Splatting module, our approach drives the moment-specific representation to capture motion dynamics with precision, controllability, and realism in terms of both appearance and geometry.

\section{Implementation Details}\label{implementation_details}

We apply the end-to-end training manner with Gaussian Surfels as our basis representation. Instead of SfM~\cite{schoenberger2016sfm} based initializing in 3DGS~\cite{kerbl3Dgaussians} and Gaussian Surfels~\cite{Dai2024GaussianSurfels}, we leverage the FLAME template and initialize the Gaussian kernel on its surface, with one Gaussian kernel on the center of each triangle face. The initialization of other attributes follows the original 3DGS implementation. All the components, including Neural Head Basis, Residual embedding, and deformation networks, join to start the end-to-end formal training for $30000$ iterations. Please refer to Tab.~\ref{tab:hyperparameters} for detailed hyperparameters of TimeWalker setting during the training process.
We follow ~\cite{wu20234dgaussians} and apply the same initial learning rate and rate scheduler for all network components and Gaussian attributes. We use a single NVIDIA A100 GPU to train the model, and it costs $3\sim5$  hours on average for the whole training process.

\noindent\textbf{Training Losses.} As mentioned in Sec.~$3.4$ of the main paper, we include three levels of losses for training the framework:
$(1)$ Image Level Supervision. Similar to 3DGS~\cite{kerbl3Dgaussians}, This term includes photometric $L1$ loss $\mathcal{L}_{\mathrm{rgb}}$ and ssim loss $\mathcal{L}_{\mathrm{ssim}}$. An additional perceptual loss $\mathcal{L}_{\mathrm{lpips}}$ ~\cite{johnson2016perceptual} with AlexNet encoder~\cite{krizhevsky2012imagenet} is included to improve the rendering quality. 
$(2)$ Geometry Level Supervision. Inspired by INSTA~\cite{zielonka2023instant}, we include $\mathcal{L}_{\mathrm{depth}}$ to enforce a better Gaussian geometry based on FLAME tracked mesh. Specifically, we apply $L1$ loss between the predicted depth (see Eq.~$3$ in main paper) and GT depth rasterized from FLAME mesh, with respect to a specific face region segmented by a ready-to-use face parsing model~\cite{yu2021bisenet}. Following Guassian Surfels~\cite{Dai2024GaussianSurfels}, we apply both $\mathcal{L}_{\mathrm{normal}}$ and $\mathcal{L}_{\mathrm{consist.}}$. The former acts as a prior-based supervision to improve the training stability, and the latter enforces consistency between the rendered depth $\tilde{D}$ and rendered normal $\tilde{N}$:
\begin{align}
\label{eq:geo_loss}
\mathcal{L}_{\mathrm{normal}} &= 0.04 \cdot (1-\tilde{\mathbf{N}}\cdot \hat{\mathbf{N}})+ 0.005 \cdot L_1(\nabla{\tilde{\mathbf{N}}}, \mathbf{0}) \\
\mathcal{L}_{\mathrm{consist.}} &=1-\tilde{\mathbf{N}}\cdot N(V(\tilde{\mathbf{D}})),
\end{align}
where $\hat{\mathbf{N}}$ denotes the normal map from a pretrained monocular model from ~\cite{eftekhar2021omnidata} and $\nabla{\tilde{\mathbf{N}}}$ represents the gradient of the rendered normal. 
$(3)$ Regulation. To ensure that the Gaussian attributes do not deviate significantly from their mean representation, we employ a $L1$ regulation to the deformation of the Gaussian attributes [$\mathbf{\delta}_\mathbf{x}, \mathbf{\delta}_\mathbf{r}, \mathbf{\delta}_\mathbf{s}, \mathbf{\delta}_\mathbf{\sigma}, \mathbf{\delta}_\mathbf{C}$] and penalize large deformation. The total loss function can be constructed as:
\begin{align}
\label{eq:loss_total}
    \mathcal{L}_{\mathrm{image}} &= \lambda_{\mathrm{r}}\mathcal{L}_{\mathrm{rgb}} + \lambda_{\mathrm{s}}\mathcal{L}_{\mathrm{ssim}} + \lambda_{\mathrm{l}}\mathcal{L}_{\mathrm{lpips}} \\
    \mathcal{L}_{\mathrm{geometry}} &= \lambda_{\mathrm{d}}\mathcal{L}_{\mathrm{depth}} + \lambda_{\mathrm{n}}\mathcal{L}_{\mathrm{normal}} + \lambda_{\mathrm{c}}\mathcal{L}_{\mathrm{consist.}} \\
    \mathcal{L}_{\mathrm{regulation}} &= \lambda_{\mathrm{reg}}\mathcal{L}_{\mathrm{deform}} \\
    \mathcal{L}_{\mathrm{total}} &= \mathcal{L}_{\mathrm{image}} + \mathcal{L}_{\mathrm{geometry}} + \mathcal{L}_{\mathrm{regulation}},
\end{align}
where $\lambda_{\mathrm{r}}$ denotes dynamic weight based on a face parsing mask. 
Note that, for the warm-up phase, as we only optimize the attributes of Gaussian kernel, we do not include regulation terms. For the formal training phase, all loss terms are employed. {Refer to Tab.~\ref{tab:hyperparameters} for details of hyperparameters. }

\noindent{\textbf{Warm Up.} Prior to formal training, we incorporate a warm-up phase wherein we utilize a person's data across all his/her lifestages to individually optimize this set of Gaussian kernels, without including the Neural Head Basis module to learn moment-specific features. In this way, the Gaussian kernels in canonical space are solely optimized to accommodate multiple lifestages, approaching a mean representation. This approach offers the benefit of enabling the Gaussian kernels to promptly learn the mean head of a person in long-horizon time periods, thereby significantly expediting subsequent training convergence. Following the warm-up phase, we commence optimizing the neural head base while simultaneously fine-tuning the Gaussian Surfels to enhance the mean head representation. This optimization process ensures that both the neural head base and the Gaussian Surfels continually improve and refine the overall representation. The warmup phase lasts for $5000$ iterations before the formal training, with applying all loss items except deform regulation.

\begin{table}[h!]
    \centering
    \medskip
    \caption{\small{\textbf{Hyperparameters during training process.}}}
    \resizebox{0.45\textwidth}{!}{
    \begin{tabular}{ccc}
        \toprule
        Type & Parameter &     Value \\
        
        \midrule
        \multirow{5}*{Hashgrid~\cite{muller2022instant}} & Number of levels & 16  \\
        ~ & Hash table size & $2^{18}$  \\
        ~ & Number of features per entry & 8 \\
        ~ & Coarsest resolution & 16  \\
        ~ & Finest resolution & 2048 \\
        
        \midrule
        \multirow{3}*{Dynamo(Sec.~3.2.2)} & Perset threshold $\kappa$ & $0.0001$  \\
        ~ & Start iterations $Q$  & $10000$  \\
        ~ & Iteration interval $q$ & $10000$ \\

        \midrule
        \multirow{8}*{Weight of loss(Eq.~\ref{eq:loss_total})} & RGB loss(mouth\&eye region) $\lambda_{\mathrm{r}}$ & $40.0$  \\
        ~ & RGB loss(otherwise) $\lambda_{\mathrm{r}}$ & $1.0$ \\
        ~ & SSIM loss $\lambda_{\mathrm{s}}$ & $1.0$ \\
        ~ & LPIPS loss $\lambda_{\mathrm{l}}$ & $1.0$  \\
        ~ & Depth loss $\lambda_{\mathrm{d}}$ & $1.25$ \\
        ~ & Normal loss $\lambda_{\mathrm{n}}$ & $1.0$ \\
        ~ & Consistency loss $\lambda_{\mathrm{c}}$ & $1.0$ \\
        ~ & Regulation $\lambda_{\mathrm{reg}}$ & $0.01$ \\
        
        \bottomrule
    \end{tabular}
    \vspace{-0.25cm}
    \label{tab:hyperparameters}
    }
\end{table}

\section{TimeWalker-1.0}\label{dataset_details}
When modeling a lifelong head avatar, existing open-source datasets are limited by a lack of life-stage variations and insufficient data scale, please refer to Sec.~\ref{dataset_compare} for the details. To fill this data requirement, we construct a large-scale and high-resolution head dataset of the same individual at different lifestages. In the following section, we will first introduce the construction pipeline of our TimeWalker-1.0 in Sec.~\ref{dataset_construction}. Then we demonstrate the comprehensiveness of it in lifelong head modeling through \textbf{Statistics Analysis} (Sec.~\ref{dataset_statistics}) and \textbf{Comparison with Other Datasets} (Sec.~\ref{dataset_compare}).

\subsection{TimeWalker-1.0 Construction}
\label{dataset_construction}

The data collection process involves the following steps. Initially, we query high-quality videos from YouTube with resolutions greater than 1080P using predefined search prompts to gather a variety of person-specific videos that exhibit diversity in content and appearance. Additionally, for movie stars, we collect their movie sources as supplementary data. Subsequently, an automated video pre-processing pipeline is developed to extract headshots. This involves the detection and cropping of human faces from the raw videos, achieved by leveraging a pretrained face recognition model to isolate frames featuring the target individual. Following this step, aesthetic assessment models such as HPSv2~\cite{wu2023human} and LIQE~\cite{zhang2023blind} are employed to sift out low-quality head photos from the pool of selected images. Ultimately, the filtered output undergoes a manual review by human evaluators to ensure that the retained headshots meet the requisite quality standards.

\subsection{Statistics}
\label{dataset_statistics}
TimeWalker-1.0 consists of 40 celebrities' lifelong photo collections, with each celebrity containing diverse variations over different lifestages(\textit{e.g.,} shape, headpose, expression, and appearance). The data volume ranges from $15K$ to $260K$ for each celebrity.

We delve into an in-depth exploration of the data distribution within the dataset. \textbf{\textit{(1) Attributes: }}The statistical analysis of the overarching human-centric attributes is delineated in Fig.~\ref{fig:dataset_attrs}, showcasing a broad spectrum of attribute distribution following a long-tail pattern. \textbf{\textit{(2) Brightness: }}The brightness is calculated by averaging the pixels and then converting them to ``perceived brightness''~\cite{bezryadin2007brightness}. The lower variance of brightness indicates a more similar luminance within the video clip. We analyze the brightness variance from two perspectives: the video level and the celebrity level, where the video level calculates the inter-video brightness variance of the whole dataset while the celebrity level assesses the brightness variance of all the videos belonging to the same celebrity. As depicted in Fig.~\ref{fig:dataset_lights}, the collected dataset shows relatively flat changes over inter-video brightness variance, and possesses more diverse lighting changes in each life-long ID. While comparing the inter-video brightness variance with the conventional testbed INSTA dataset for single appearance reconstruction,  our dataset embraces more challenges with larger diversity and wider spectrum.  \textbf{\textit{(3) Age, Gender, and Headpose: }} As shown in Fig.~\ref{fig:dataset_age_headpose}, we demonstrate the age distribution of the TimeWalker-1.0, which indicates our dataset has a balanced age distribution without being biased towards certain age group. Moreover, the dataset includes celebrities of multiple ethnicities (Brown, Yellow, White, and Black). The headpose distribution graph in Fig.~\ref{fig:dataset_age_headpose} highlights the broad spectrum of captured head poses distributed in a nearly normal fashion.

\begin{table*}[]
    \scriptsize
    \centering
    \caption{\small{\textit{Dataset comparison.} We compare TimeWawlker-1.0 with face datasets from ``Lifestage'', ``Identity'', ``Age'', ``Expression'', ``Frame Count'', ``Ethnicity'', and ``Accessory''.}}
    \resizebox{0.9\textwidth}{!}{
    \begin{tabular}{ccccccccc}
        ~ & \multicolumn{2}{c}{\textbf{Realism}} & \multicolumn{6}{c}{\textbf{Diversity}} \\
        \hline
        \multicolumn{1}{c|}{\textbf{Dataset}} & Resolution &  \multicolumn{1}{c|}{Lifestage} & ID & Age & Expression & Frame Count & Ethnicity & Accessory \\
        \hline
        \textit{FFHQ}~\cite{ffhq} & 1024$\times$1024 & \textcolor[HTML]{D092A7}\xmark & - & - & \textcolor[HTML]{D092A7}\xmark & 70k & \textcolor[HTML]{A5B592}\cmark & \textcolor[HTML]{A5B592}\cmark \\
        \textit{CACD}~\cite{chen2014cross} & <512$\times$512 & \textcolor[HTML]{A5B592}\cmark & 2k & 16-62 & \textcolor[HTML]{A5B592}\cmark& 163k & - & - \\
        \textit{HRIP}~\cite{liang2016head} & 414$\times$464 & \textcolor[HTML]{A5B592}\cmark & 4 & - & \textcolor[HTML]{A5B592}\cmark & 4k & \textcolor[HTML]{D092A7}\xmark & \textcolor[HTML]{D092A7}\xmark \\
        \textit{INSTA}~\cite{zielonka2023instant} & 512$\times$512 & \textcolor[HTML]{D092A7}\xmark & 12 & - & \textcolor[HTML]{A5B592}\cmark & <54k & \textcolor[HTML]{D092A7}\xmark & \textcolor[HTML]{D092A7}\xmark \\
        \textit{RenderMe-360}~\cite{pan2024renderme} & 2448$\times$2048 & \textcolor[HTML]{D092A7}\xmark & 500 & 8-80 & \textcolor[HTML]{A5B592}\cmark & >243M & \textcolor[HTML]{A5B592}\cmark & \textcolor[HTML]{A5B592}\cmark \\
        \hline
        \textbf{\textit{TimeWalker-1.0}} & 1024$\times$1024 & \textcolor[HTML]{A5B592}\cmark & 40 & 20-80 & \textcolor[HTML]{A5B592}\cmark & ~2.7M & \textcolor[HTML]{A5B592}\cmark & \textcolor[HTML]{A5B592}\cmark\\
    \hline
    \end{tabular}
    \label{tab:comp_dataset}
    }
\end{table*}

\begin{figure*}
    \centering
      \vspace{-0.2cm}
    \includegraphics[width=1.0\linewidth]{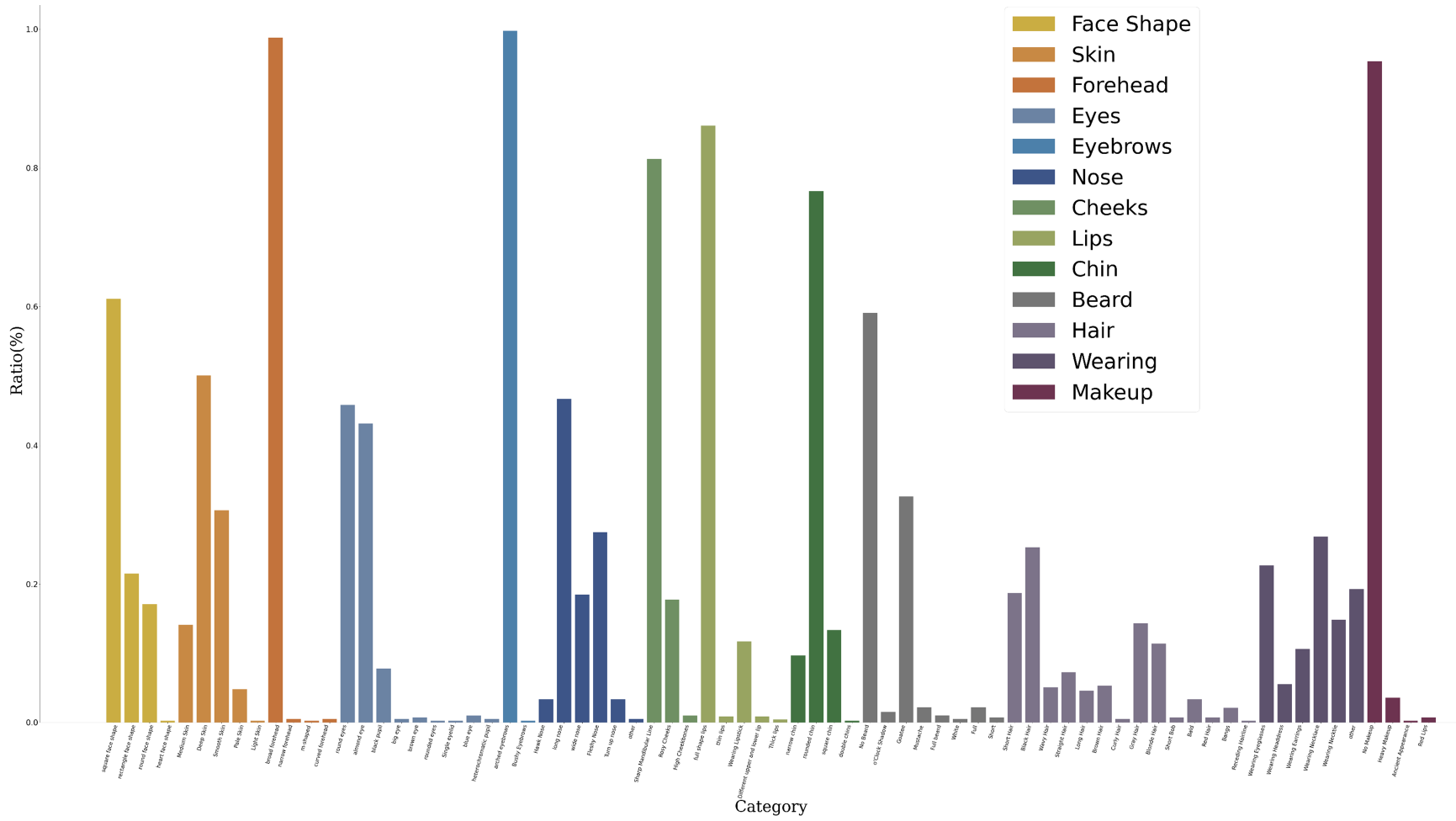}
      \captionsetup[figure]{hypcap=false}
        \vspace{-0.2cm}
      \captionof{figure}{\small{\textbf{Statistics: Attributes.} The statistics of TimeWalker-1.0 reveal a wide spectrum of attribute distribution characterized by a long-tail pattern.}}
    
    \label{fig:dataset_attrs}
\end{figure*}

\begin{figure*}
    \centering
    \includegraphics[width=0.93\linewidth]{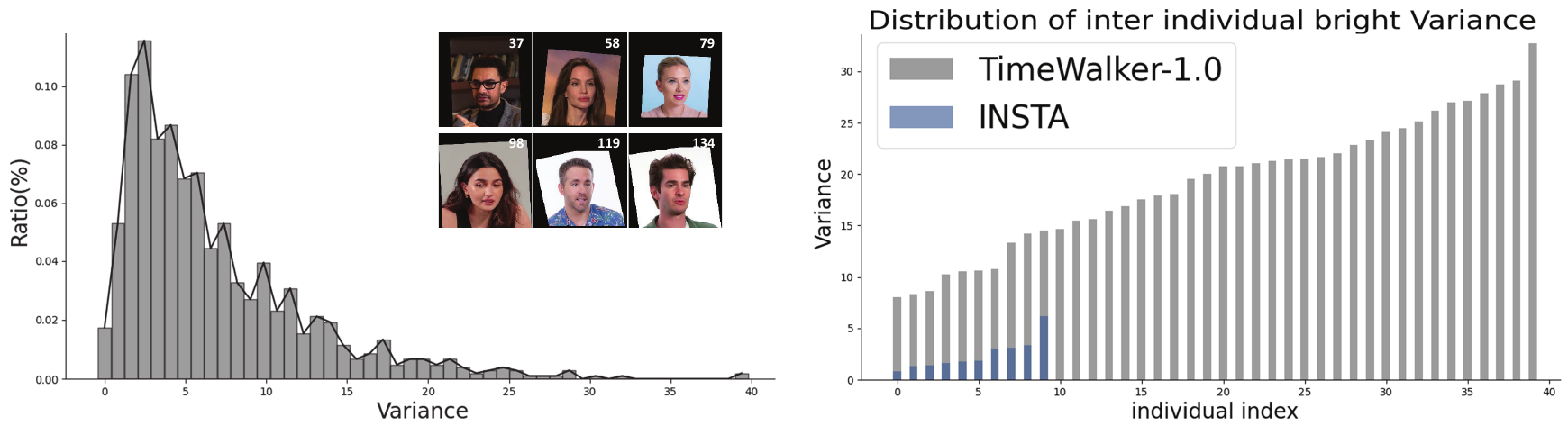}
      \captionsetup[figure]{hypcap=false}
      \captionof{figure}{\small{\textbf{Statistics: Brightness Variance.} The left part represents the inter-video brightness variance of the dataset, while the right part shows the luminance condition across the whole videos of the same celebrity. The right part illustrates that our dataset enjoys more diverse lighting changes than the INSTA~\cite{zielonka2023instant} dataset.}}
    \label{fig:dataset_lights}
\end{figure*}

\begin{figure*}
    \centering
    \includegraphics[width=0.95\linewidth]{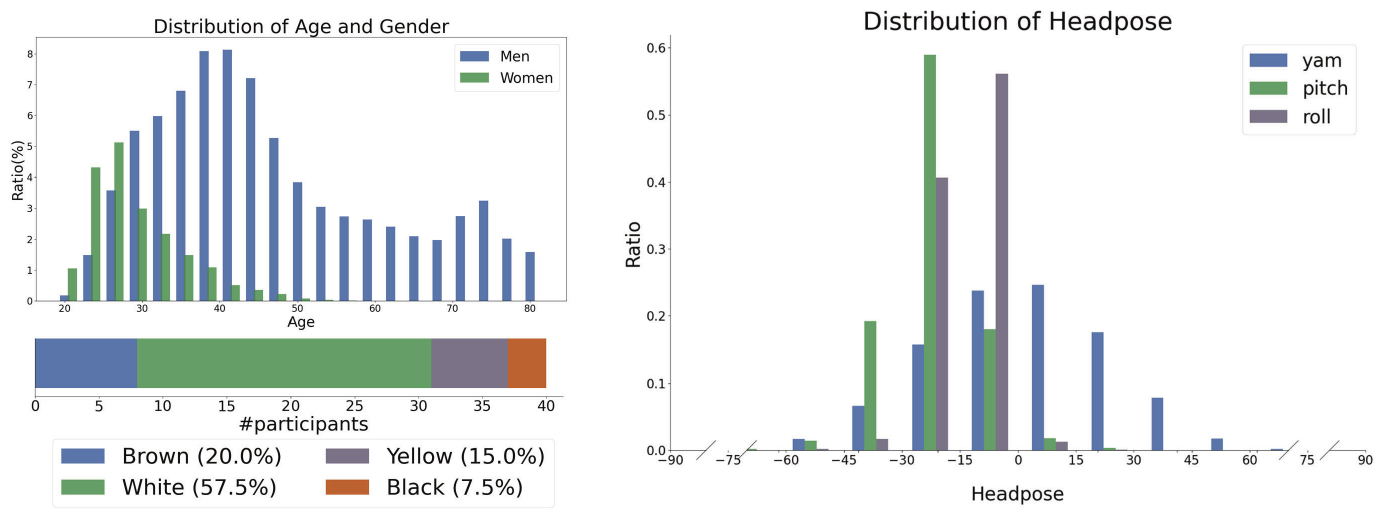}
      \captionsetup[figure]{hypcap=false}
      \captionof{figure}{\small{\textbf{Statistics: Age, Ethnicity, and Headpose.} The selected videos showcase a diverse range of ages, ethnicities, and headposes. We divide the headpose into $12$ clusters with each covering an angle range of $15^{\circ}$ and we calculate the ratio of each cluster to the total number of headposes as shown in the right figure.}}
    \label{fig:dataset_age_headpose}
\end{figure*}

\subsection{Comparison with Other Datasets}
\label{dataset_compare}
We also compare our TimeWalker-1.0 with other datasets to show its superiority. As shown in Tab.~\ref{tab:comp_dataset}. The previous dataset either concentrates on 2D head photo generation without 3D supervision, such as FFHQ~\cite{ffhq}, or records videos of a specific life-stage where the shape varies are not disputed, such as RenderMe-360~\cite{pan2024renderme} and INSTA~\cite{zielonka2023instant}. In addition, early datasets like CACD~\cite{chen2014cross} and HRIP~\cite{liang2016head} collect cross-age celebrity data for face recognition and rough head shape modeling. Their data scale and image resolution severely limit them from being implemented for life-long head avatars. Therefore, our TimeWalker-1.0 dataset, stands out in life-stage avatar modeling for its high image resolution, large-scale, wide age range, diverse ethnicity, and most importantly across life-stage data groups.

\section{Internal Experiments}\label{internal_experiments_appendix}

\subsection{Additional Ablation Studies}

\noindent\textbf{Dynamo.}
Whereas the main paper analyzes the quantitative ablation results of the Dynamo module, this section supplements the qualitative analysis. Fig.~\ref{fig:ablation} shows that, without Dynamo, the model tends to preserve moment-specific attributes in the learnable latent, leading to artifacts surrounding the head avatar ($(d)$ in Figure). Models trained with only one hashgrid cannot provide enough details on head surface and have obvious artifacts around the eyes ($(e)$), while those trained with adequate hashgrids perform similarly to our pipeline but require larger model sizes ($(f)$).

\noindent\textbf{Loss Term.}
As shown in Tab.~\ref{tab:ablation}, the performance significantly drops without $\mathcal{L}_{\text{lpips}}$ or $\mathcal{L}_{\text{deform}}$, and $\mathcal{L}_{\text{geometry}}$ does not contribute to the rendering result. This is consistent with the ablation experiments in ~\cite{Dai2024GaussianSurfels}, as the geometry-based loss mainly contributes to the high quality mesh reconstruction rather than the realism of avatar rendering. Fig.~\ref{fig:ablation_mesh_geo} visualizes the mesh without $\mathcal{L}_{\text{geometry}}$. As observed, the reconstructed mesh exhibits substantially inferior geometric quality and fails to accurately reconstruct the oral structures.
From the visualization of Fig.~\ref{fig:ablation}, key observations are: $1)$ including $\mathcal{L}_{\text{lpips}}$ helps reduce smoothness in high frequency parts like hair and beard. $2)$ $\mathcal{L}_{\text{geometry}}$ does not contribute to rendering result, but aids in mesh reconstruction. $3)$ Training without $\mathcal{L}_{\text{deform}}$ results in strip-shaped Gaussian ellipses or even generation failed.

{
  \begin{figure}
  \vspace{-0.1cm}
  \begin{center}
  \includegraphics[width=0.4\textwidth]{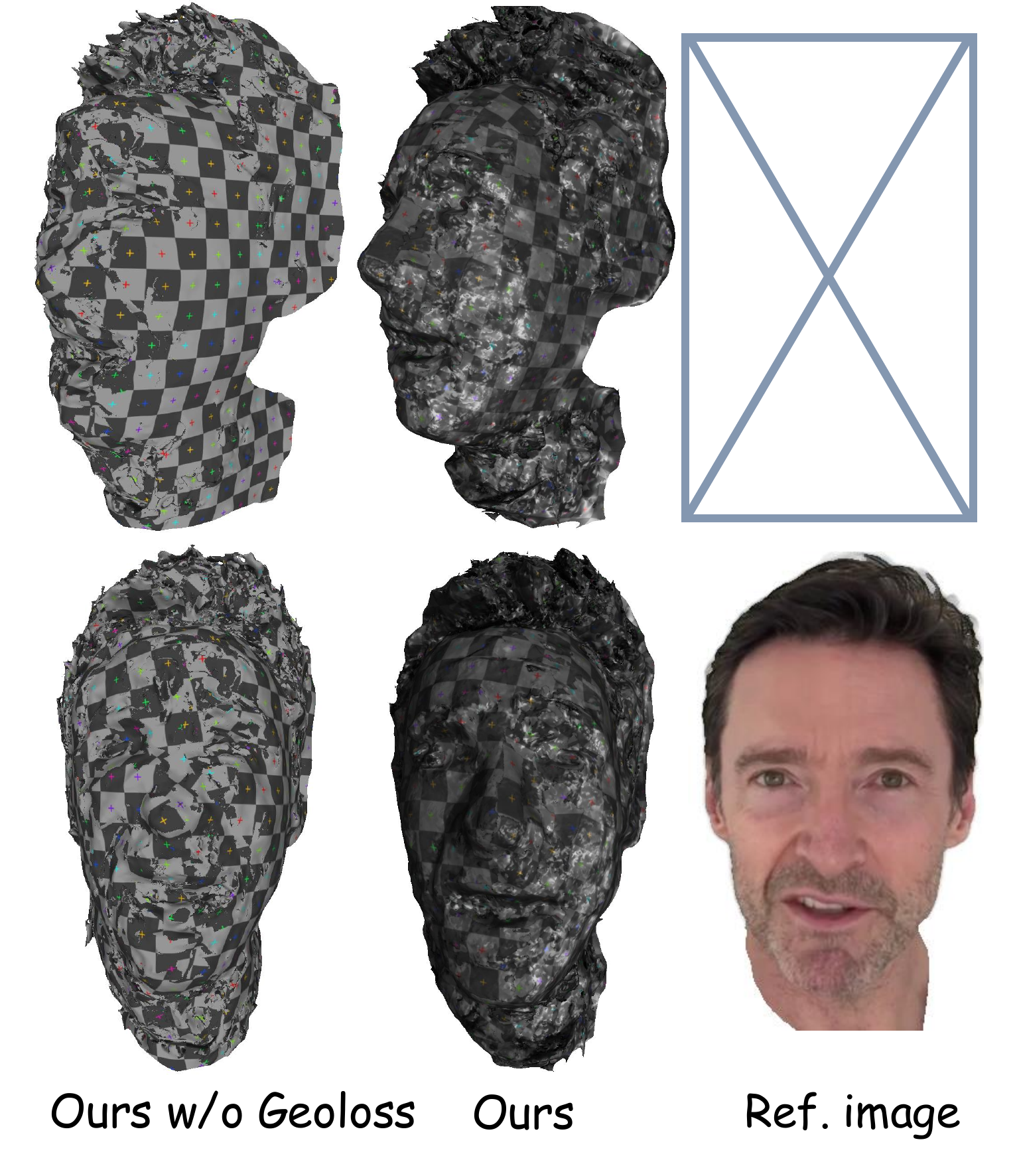}
  \vspace{-0.2cm}
  \captionsetup[figure]{hypcap=false}
  \captionof{figure}{\small{\textbf{Static Mesh Ablation.} We visualize and compare static mesh reconstruction ablation with and without geometry loss. We render the meshes in both frontal and side views using Blender software under identical rendering conditions. Despite these consistent settings, rendering results exhibit significant differences due to the diverse topologies of the meshes (Zoom in for details).}}
  
  \label{fig:ablation_mesh_geo}
 \end{center}
   \vspace{-0.45cm}
  \end{figure}

}








\begin{table}[]
\scriptsize
\centering
\renewcommand{\arraystretch}{0.9}
\caption{\small{\textbf{Ablation on Hashgrid.} 
\colorbox{colorbest}{Pink} is best, 
\colorbox{colorsecond}{orange} is second.}} 

\resizebox{0.47\textwidth}{!}{
\begin{tabular}{c|ccccc}
\toprule

& \textbf{w/o Dynamo} 
& \textbf{1 Hashgrid} 
& \textbf{All Hashgrid} 
& \textbf{20 Hashgrid} 
& \textbf{Ours} \\
\midrule

\textbf{PSNR$\uparrow$} 
& 21.69 
& 24.84 
& 26.86 
& 26.94 
& \colorbox{colorsecond}{27.20} 
\\

\textbf{SSIM$\uparrow$} 
& 0.767 
& 0.890 
& 0.938 
& 0.935 
& \colorbox{colorsecond}{0.941} 
\\

\textbf{LPIPS$\downarrow$} 
& 0.197 
& 0.119 
& \colorbox{colorsecond}{0.078} 
& \colorbox{colorbest}{0.077} 
& \colorbox{colorbest}{0.077} 
\\

\bottomrule
\end{tabular}
}
\label{tab:ablation}
\end{table}

\noindent\textbf{Residual Embedding.}
To validate the effectiveness of Residual Embedding, we conducted ablation experiments on a representative identity. We compared five settings: (1) No Residual Embedding(RE);(2) RE as input of color MLP; (3) RE as input of deformand color MLP, sharing the same input; (4) RE as input of deform and color MLP, seperately; (5) our proposed RE as the input of deform MLP. Quantitative results are shown in Table 4 below. These results confirm that the Residual Embedding plays a critical role in refining lifestage-specific details without compromising global identity, and our residual connection design optimally balances these two objectives.
\begin{table}[]
\scriptsize
\centering
\small
\renewcommand{\arraystretch}{0.8}
\caption{\small{\textbf{Ablation study of Residual Embedding.} \colorbox{colorbest}{Pink} indicates the best and \colorbox{colorsecond}{orange} indicates the second. }} 
\resizebox{0.48\textwidth}{!}{

\begin{tabular}{c|ccc}
\toprule
\textbf{Settings} & \textbf{PSNR↑} & \textbf{SSIM↑} & \textbf{LPIPS↓} \\
\midrule
No Residual Embedding & 26.74 & 0.961 & 0.064 \\
ColorMLP & 24.6 & 0.951 & 0.071 \\
DeformMLP \& ColorMLP (Share) & \colorbox{colorsecond}{26.97} & \colorbox{colorsecond}{0.965} & \colorbox{colorsecond}{0.057} \\
DeformMLP \& ColorMLP (Seperate) & 26.9 & 0.965 & 0.057 \\
\midrule
DeformMLP (ours) & \colorbox{colorbest}{27.1} & \colorbox{colorbest}{0.966} & \colorbox{colorbest}{0.055} \\
\bottomrule
\end{tabular}
}
\label{tab:ablation_residual_embedding}
\end{table}

{
  \begin{figure*}[tb]
    \vspace{-0.15cm}
    \centering
  \includegraphics[width=0.9\textwidth]{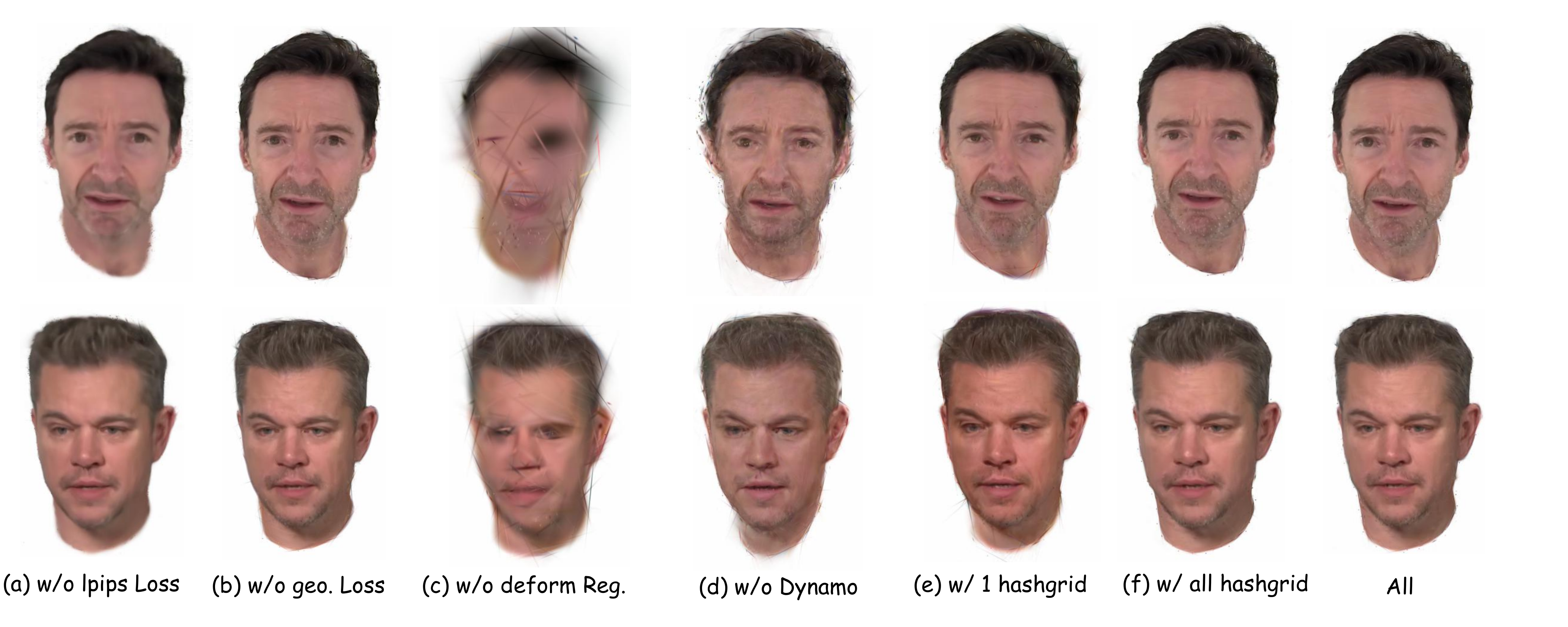}
    \vspace{-0.3cm}
  \captionsetup[figure]{hypcap=false}
  \caption{\small{\textbf{Ablation Study.}  Experiments with different loss setting are showed in $(a-c)$, while ablation with Dynamo and hashgrid are visualized in $(d-f)$. {\textit{All}} represents our final model. Better zoom in for details. }}
  \vspace{-0.25cm}
  \label{fig:ablation}
  \end{figure*}

}

\subsection{Personalized Space Visualization}
\label{personal_space}

Fig.~\ref{fig:lifestage} shows cases on several individuals. Part $(a)$ shows the model's ability to walk through long age period of a person without losing rendering realism, and to represent multiple appearances with diverse skin color. We own this to our powerful Neural Head Basis module which is capable of learning intrinsic features as well as appearance deformation. The disentangled design of neural deformation field and inverse warping operation enables changing lifestage but keeping other dimensions like shape, pose and expression unchanged. Part $(b)$ showcases model's capacity to delineate lifestages within a personalized space across different ethnicities and genders.

{
  \begin{figure*}[t]
  \vspace{-0.1cm}
  \includegraphics[width=0.9\textwidth]{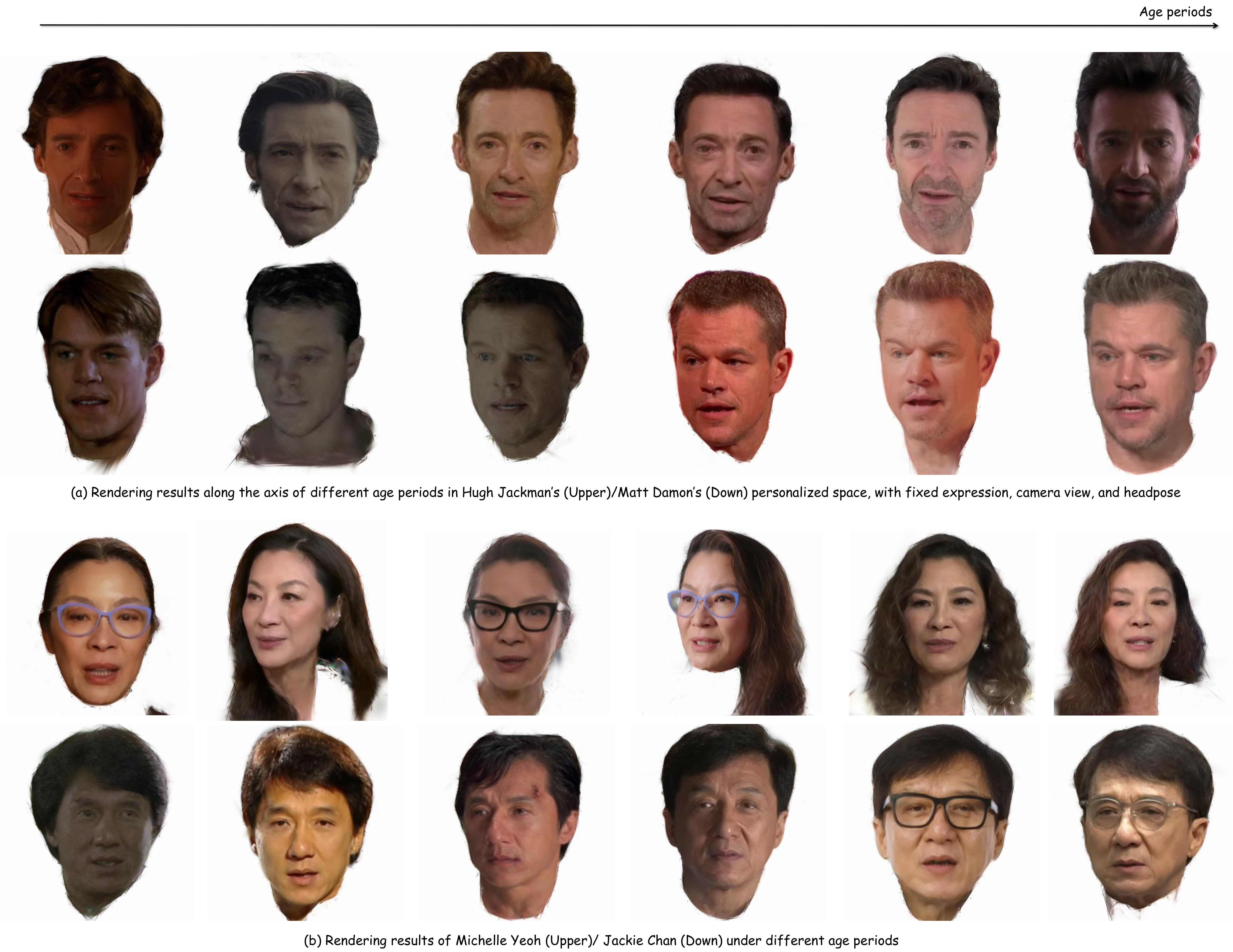}
   \vspace{-0.2cm}
  \captionsetup[figure]{hypcap=false}
  \captionof{figure}{\small{\textbf{Personalized Space: Lifestage.} We demonstrate multiple individuals and their replicas in different lifestages. (a) We adjust the value in Dynamo to animate the lifestages of individuals, but keep other animation values unchanged. (b) We show the lifestages of more individuals with different ethnicity. }}
  \vspace{-0.15cm}
  \label{fig:lifestage}
  \end{figure*}
}

\subsection{Large-Viewpoint Rendering}

To assess the robustness of our pipeline concerning varying viewpoints, we visualize the rendering outcomes at increasing yaw angles in Fig.~\ref{fig:large_angle}. Generally, the rendering quality diminishes as the rendering angle increases. TimeWalker produces reasonable and high-fidelity results within a 45-degree range, exhibiting minimal or negligible artifacts, which exhibits significant adaptability of our pipeline, allowing for effective compensation and interpolation for previously unseen side profiles.
However, when rendering from viewpoints exceeding 45 degrees, the method generates noticeable artifacts. This limitation can be attributed to the challenges posed by unstructured data. Our dataset, curated from in-the-wild sources, inherently exhibits a long-tail distribution. Although our method mitigates data imbalance to some extent, it remains unable to effectively handle extremely large yaw angles due to the lack of corresponding training data in those regions across all age periods. The decline in quality associated with large viewpoints has also been addressed in other head avatar pipelines~\cite{xu2023avatarmav,zhang2023learning,caliskan2025pav}. This issue can be mitigated by utilizing multiview lab data as a training set~\cite{xu2024gaussian,qian2023gaussianavatars} or by leveraging the capabilities of generative models~\cite{chan2022efficient,kirschstein2023diffusionavatars} for feature compensation.

{
    \begin{figure}
        \centering
        \includegraphics[width=1\linewidth]{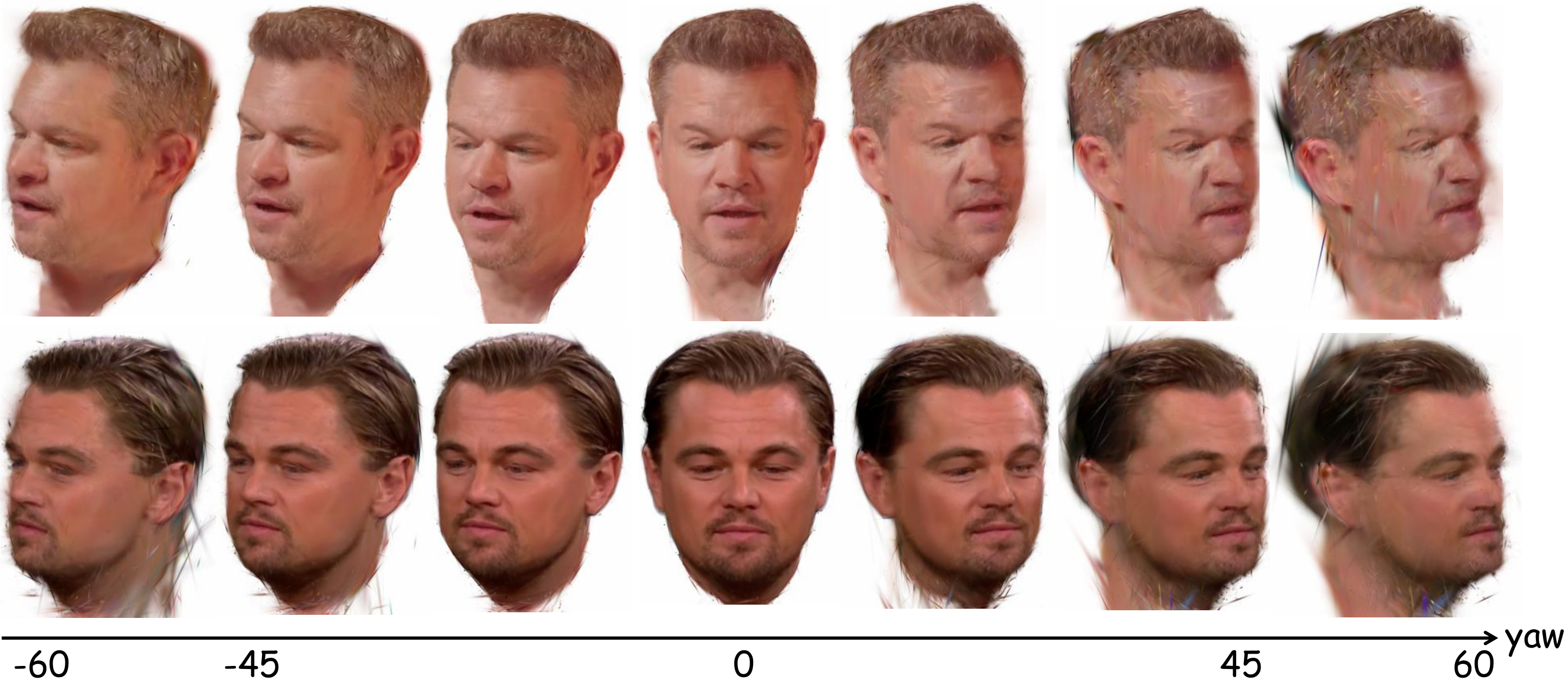}
        \caption{\small{\textbf{Large-Viewpoint Rendering.}} We present rendering results from a large viewpoint in the yaw direction. Our method yields satisfactory outcomes within a 45-degree range. However, the generated avatars exhibit noticeable artifacts at larger yaw angles, a common limitation in the research field, associated with training on in-the-wild data. }
        \label{fig:large_angle}
    \end{figure}
}

\subsection{Lighting Control}

The diverse lighting conditions encountered across different lifestages present significant challenges to fine-grained attribute disentanglement in the construction of head avatars. By integrating residual embeddings $\mathbf{l}_\mathbf{res}$, which act as global compensators for each lifestage, into our pipeline, we can partially separate lighting effects from lifestage attributes. This integration enables lighting adjustments while preserving the core characteristics of each life stage. In Fig.~\ref{fig:lighting}, we visualize the lighting control achieved by interpolating the residual embeddings while keeping all other conditions constant. In most cases, the appearance features of the identity remain consistent across varying lighting effects; however, in some instances, artifacts emerge or the life stage of the identity appears to change (third line of each lifestage in Fig.~\ref{fig:lighting}). This observation indicates that the disentanglement of lifestage and lighting by simply using the residual embedding is not yet perfect.
To enhance the disentanglement of lighting effects, several potential solutions can be considered: (1) incorporating albedo map modeling into the pipeline to isolate the intrinsic properties of the subject's appearance from external lighting influences; and (2) integrating lighting augmentations and equivalent representation learning during training to promote invariant representations that are robust to lighting variations.

{
    \begin{figure}
        \centering
        \includegraphics[width=1\linewidth]{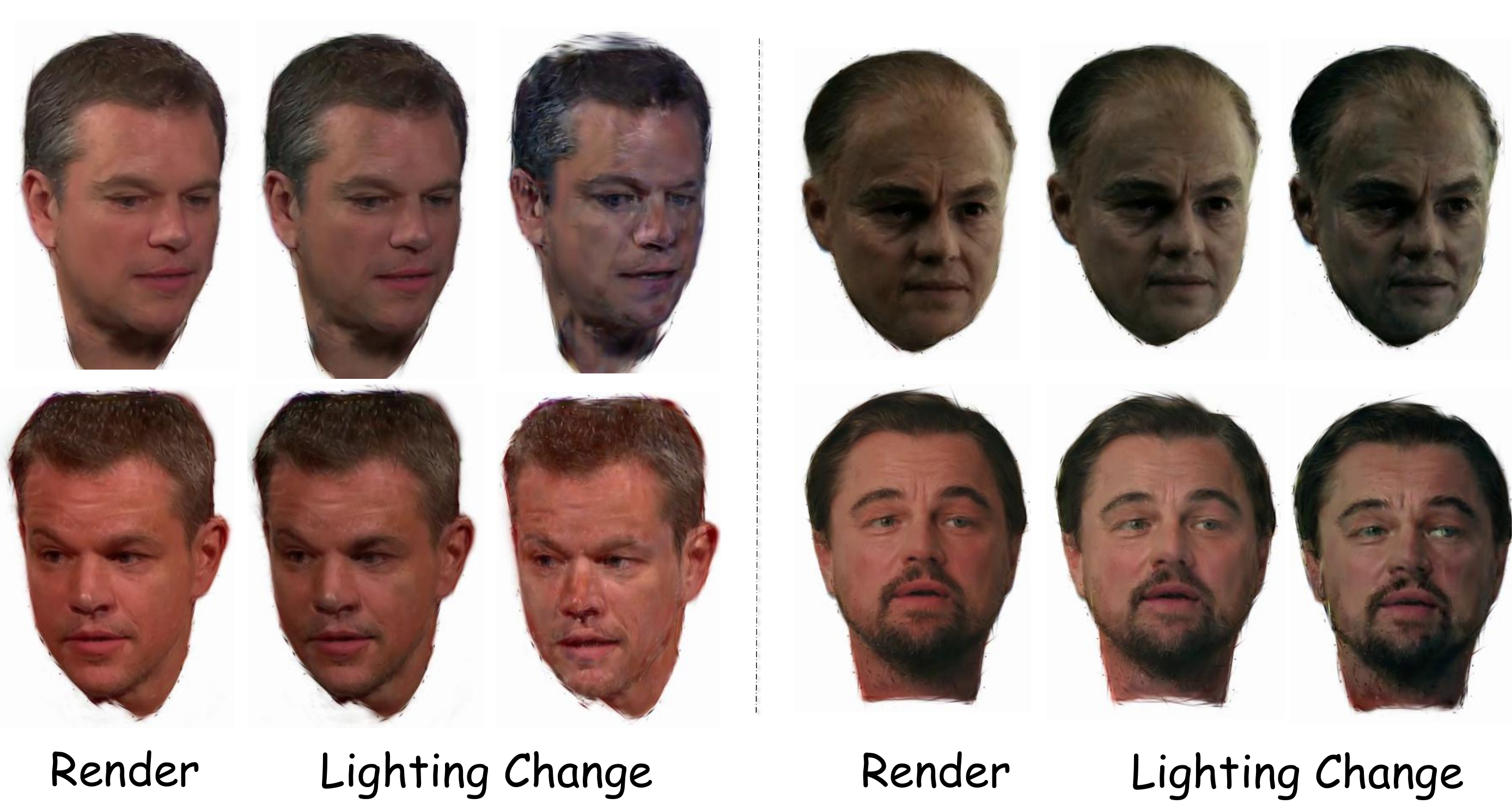}
        \caption{\small{\textbf{Lighting Control.}} We present rendering results of changing the lighting with the interpolation of residual embedding. Better zoom in for recognizing the light different.}
        \label{fig:lighting}
    \end{figure}
}

\subsection{More ID Results}

To show the generalization and effectiveness of our pipeline, we demonstrate more individual results from different genders and ethnicities in Fig.~\ref{fig:more_result}.

{
    \begin{figure}
        \centering
        \includegraphics[width=0.8\linewidth]{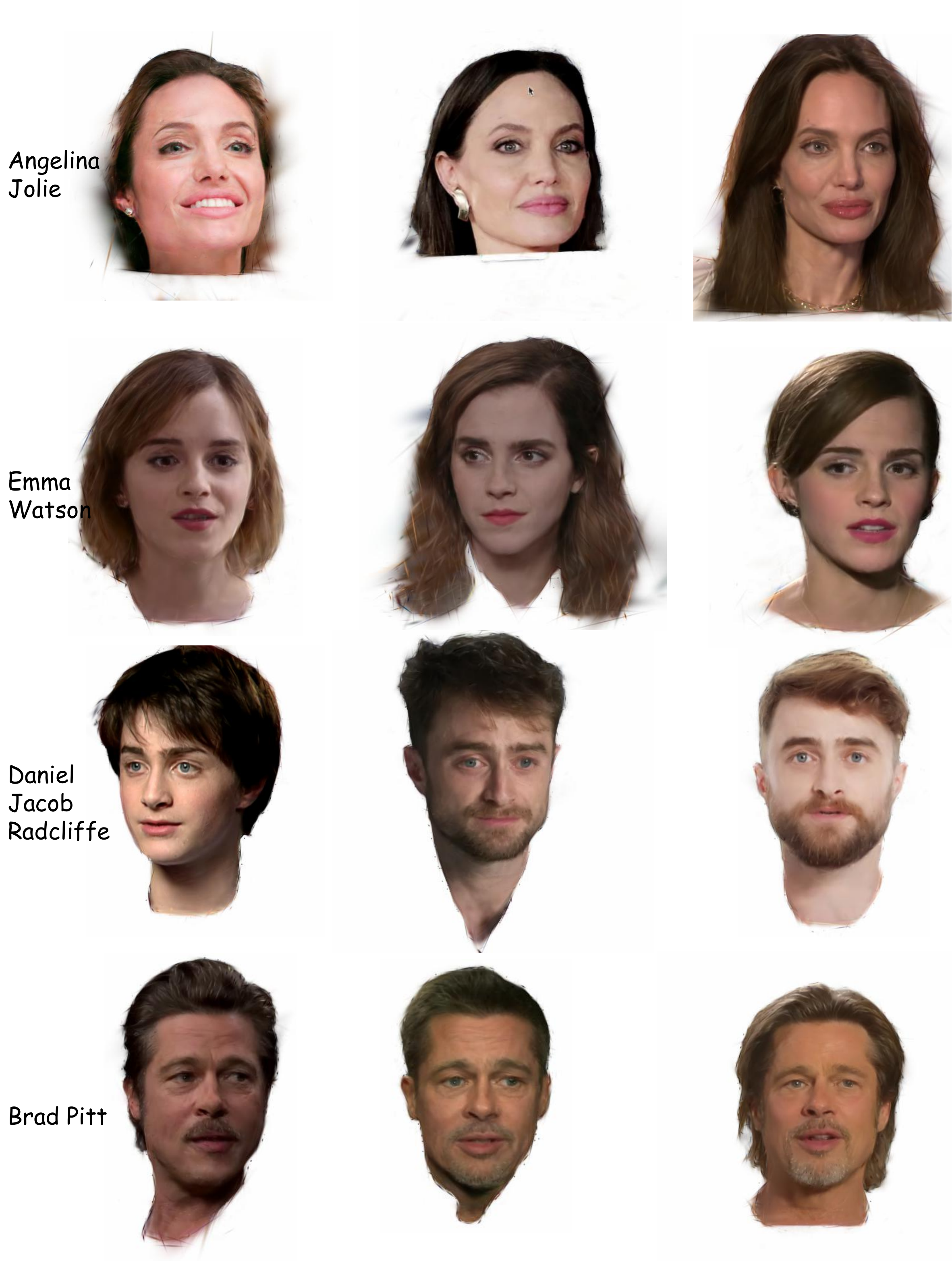}
        \caption{\small{\textbf{More sampled ID result.}} We visualize more individuals with different genders.}
        \label{fig:more_result}
    \end{figure}
}

\subsection{Time costing} 

\textbf{\textit{Training.}} To further validate the effectiveness of the Dynamo module, we present experimental statistics on training time costs in Table~\ref{tab:hashgrid_training_cost}. As observed, training time generally increases with the number of hashgrids. However, model performance does not exhibit a corresponding continuous improvement(Refer to Table.~$3$ in main paper). This finding further confirms the effectiveness of the Dynamo module, which demonstrates that it does not rely on an increased number of hashgrids to achieve superior results.

\begin{table}[htbp]
  \centering
  \caption{\small{\textbf{Ablation of the training cost with different Hashgrid.}}}
  \label{tab:hashgrid_training_cost}
  \resizebox{0.5\textwidth}{!}{
  \begin{tabular}{ccc}
  \toprule
    \textbf{w/ 5 Hashgrid} & \textbf{Dyn. Hashgrid(ours)} & \textbf{w/ 20 Hashgrid} \\
    \hline
    240min & 270min & 390min \\
  \bottomrule
  \end{tabular}
  }
\end{table}

\textbf{\textit{Meshing Process.}} In the context of single mesh generation, our approach necessitates approximately 10 minutes, a timeframe comparable to or longer than that required by Gaussian Surfels and INSTA++. Notably, our methodology exhibits a notable advantage in the creation of mesh sequences, as we leverage direct animation of the static mesh to generate a new mesh. In contrast, Gaussian Surfels and INSTA++ are compelled to iteratively execute the entire mesh generation process, leading to a linear growth in time consumption. Colmap's inability to generate mesh sequences stems from its requirement of the complete sequence to reconstruct a single mesh.

\section{Comparisons with State-of-the-Arts}\label{exp:compare_sota}
\subsection{Comparisons with Multiview Reconstruction Methods}\label{exp:compare_multiview}
\label{exp:compare_sota}
\noindent\textbf{Baseline. } 
We compare our method with state-of-the-art multiview reconstruction methods, including animatable head avatar and also neural rendering methods, extending from Tab.~$1$ of the main paper.
\textbf{INSTA}~\cite{zielonka2023instant} stands out for its ability to generate high-quality head avatars rapidly, leveraging a multi-resolution hashgrid defined in canonical space to store learned features. For ray points in deformed space, INSTA identifies the nearest triangles of the FLAME tracked mesh and computes the transformation matrix between the tracked mesh and the template mesh. Subsequently, the points are warped back to canonical space, enabling the animation of the avatar using FLAME expression parameters by updating the tracked mesh. 
To the best of our knowledge, existing state-of-the-art methods for 3D animatable head avatars do not focus on animating the head across diverse time periods, with the exception of \textbf{PAV}~\cite{caliskan2025pav}. PAV shares similar settings to our approach, as it constructs personalized head avatars from unstructured data. Building upon INSTA, PAV introduces learnable feature embeddings to adapt to various appearances and proposes an appearance-conditioned density formulation to address shape variations. However, PAV does not demonstrate the capabilities of mesh reconstruction and head shape animation in its work, both of which can be effectively accomplished using our pipeline.
Since this newly proposed method is not open-sourced, we have chosen to extend the INSTA method to \textbf{INSTA++}, enabling it to capture multiple lifestages within a single model. Concretely, we introduce a per-lifestage learnable latent code as a supplementary condition to the density MLP, enabling the storage of lifestage-specific latent information.
\textbf{FlashAvatar}~\cite{xiang2024flashavatar} integrates the Gaussian representation with a 3D parametric model by initializing Gaussian points on a 2D UV texture map attached to the mesh surface. It employs a small MLP network conditioned on the Gaussian canonical position and expression code to learn the Gaussian offset, thereby enabling motion driven by facial expressions. 
\textbf{Gaussian-Surfels}~\cite{Dai2024GaussianSurfels} aims to solve the inherent normal and depth ambiguity of 3DGS by cutting one dimension of gaussian ellipsoid, and can reconstruct the dense mesh without losing the realism of static scene rendering. After training, a Poisson mesh with post-process is applied to extract the dense mesh. Notably, although Gaussian Surfels was not specifically designed for animatable head avatars, we include it in our comparison since TimeWalker has adapted the Gaussian Surfels representation for mesh reconstruction, making it a relevant baseline for our evaluation. 
We further extend the Gaussian-surfels to \textbf{Gaussian-Surfels++}, a dynamic version to handle motion driven by expression changes. Specifically, we heuristically introduce an MLP-based warping field with expression parameters as conditions, to learn the non-rigid motion in terms of Gaussian attribute shifts. 

{
  \begin{figure*}
  \includegraphics[width=1\textwidth]{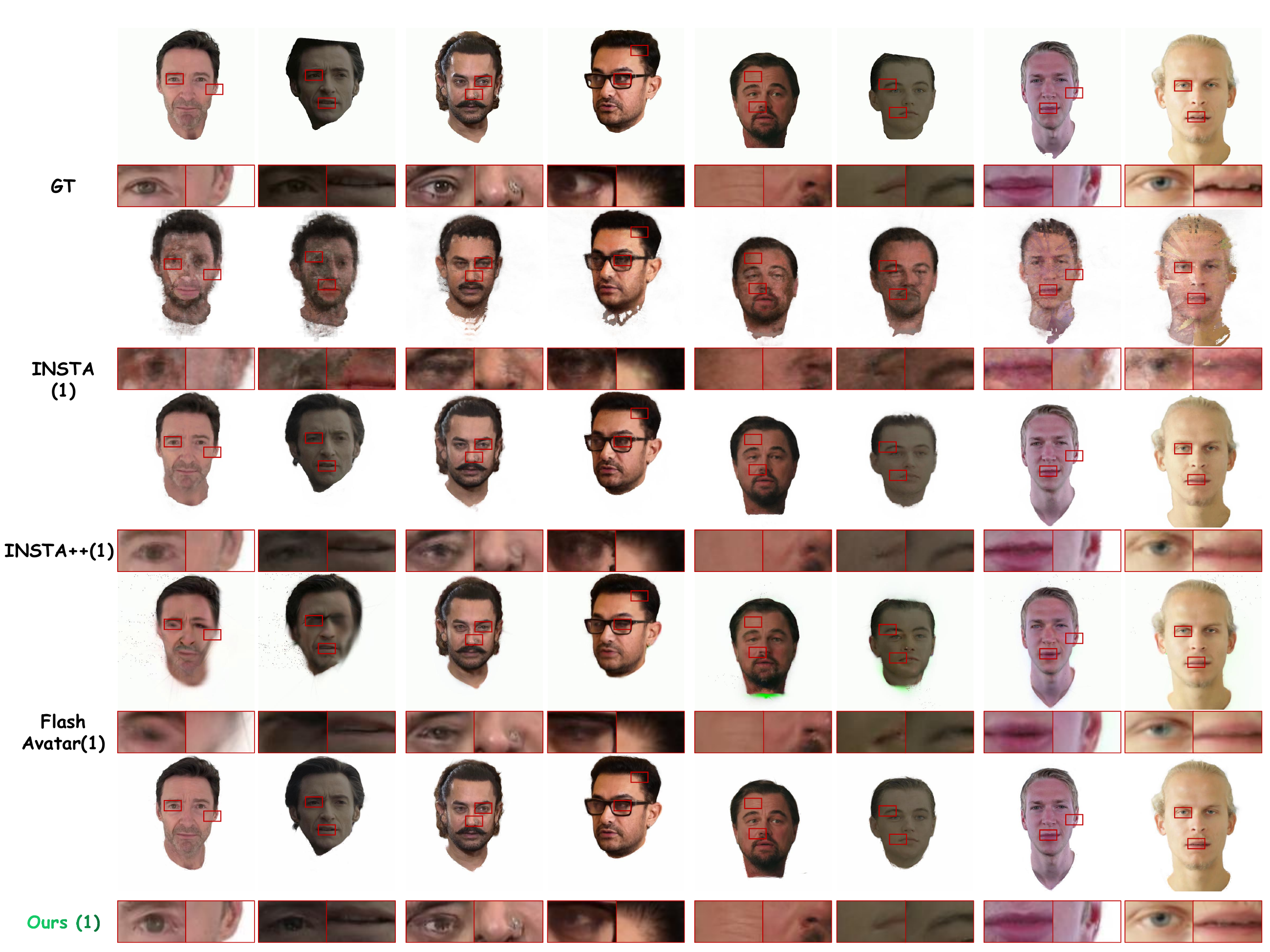}
  
  \captionsetup[figure]{hypcap=false}
  \captionof{figure}{\small{\textbf{Qualitative comparison with SOTA with \textit{\#Protocol-1}}(1 vs 1). All methods, including both ours and the baselines, involve training a single model for each individual, encompassing various lifestages.}}
  
  \label{fig:comp_1v1}
  \end{figure*}
}

{
  \begin{figure*}[t]
  \includegraphics[width=1\textwidth]{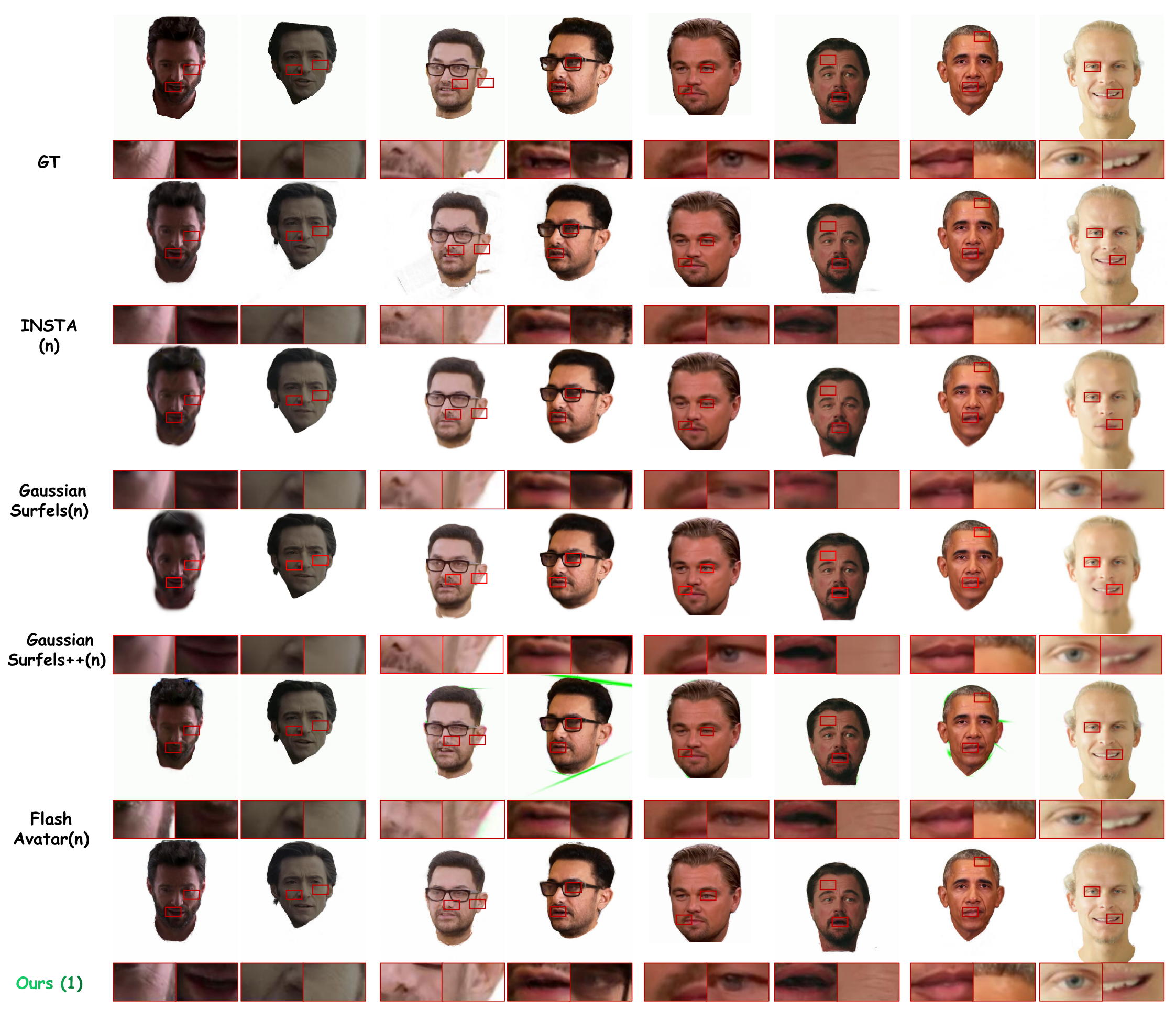}
  \captionsetup[figure]{hypcap=false}
  \captionof{figure}{\small{\textbf{Qualitative comparison with SOTA with \textit{\#Protocol-2}} (1 vs N). In this setting, our method trains one model for one individual across multiple lifestages, while the baselines train multiple models for one individual, \textit{i.e.,} one model for each lifestage.}}
  \label{fig:comp_1vn}
  \end{figure*}
}

\noindent\textbf{Results.} In main paper we mainly show the quantitative result in our TimeWalker-1.0 dataset. To further validate the robustness of our method, we conduct an additional experiment on the open-source INSTA dataset, which features 9 subjects with diverse appearances. In our setup, we treat different individuals as distinct appearances of the same person, thereby covering all 9 subjects with a single model. Quantitatively, as shown in Tab.~\ref{tab:exp_sub}, our method achieves the best results in the \textit{\#Protocol-1}, and best or second best results in the \textit{\#Protocol-2} on opensource INSTA dataset. 
Compared to the Tab.~$2$ in main paper, similar trends are observed in Tab.~\ref{tab:exp_sub}, where our method surpasses all baselines in \#Protocol-1 and delivers competitive outcomes in \#Protocol-2.
Notably, most pipelines exhibit superior performance in the publicly available INSTA dataset, but they exhibit lower performance in our TimeWalker-1.0 dataset. This disparity can be attributed to the distinct characteristics of the two datasets. The INSTA dataset is curated in a controlled lab environment with consistent lighting conditions, pre-defined head motions, and expressions. In contrast, the data in TimeWalker-1.0 is gathered from real-world scenarios, introducing greater variability in lighting, appearance, accessories, etc. This increased diversity poses significant challenges for constructing personalized spaces.

We also visualized the rendering result of protocol. In Fig. \ref{fig:comp_1v1}, it can be observed that methods like INSTA~\cite{zielonka2023instant} and FlashAvatar~\cite{xiang2024flashavatar} struggle to handle appearance variations, resulting in disruptions and blurry outcomes. Even the multi-lifestage extended INSTA++ produces unsatisfactory results with artifacts noticeable in high-frequency areas such as the eyes. Conversely, our method produces rendering outcomes with reduced blurriness and artifacts, showcasing the effectiveness of the pipeline. 
In \#Protocol-2, as indicated Fig.~\ref{fig:comp_1vn}, our method showcases the ability to adapt to varying appearances despite being trained with unstructured data. This adaptability enables our method to effectively manage appearance changes of different shapes, yielding competitive outcomes comparable to models trained on singular appearances.

\subsection{Comparison with Generative Method }\label{exp:compare_generative}
\label{exp:compare_generative}
We further compare our method with \textbf{GANAvatar}~\cite{kabadayi2023ganavatar}, a generative head avatar model, {\textit{to reveal whether the current generative model paradigm is capable of modeling personalized space in the lifelong scale.}} GANAvatar~\cite{kabadayi2023ganavatar} utilizes a two-stage training scheme for 3D head avatar, where it firstly leverages a 3D-aware generative model for personalized appearance reconstruction by training on corresponding 2D images. Then a mapping module driven by 3DMM facial expression parameters is employed for achieving facial expression control on the personalized generative model. As shown in Tab.~\ref{tab:exp_gan}, our method achieves superior quantitative results compared to GANAvatar~\cite{kabadayi2023ganavatar} on both \#Protocol-1 and \#Protocol-2, while demonstrating a similar performance trend on the INSTA and TimeWalker-1.0 datasets as the aforementioned baselines. The visual comparisons presented in Fig.~\ref{fig:comp_gan} indicate that, in the 1 vs N Comparison, GANAvatar struggles with illumination and texture details (e.g., the eyebrows). In the 1 vs 1 Comparison, GANAvatar performs even worse, as it is a personalized generative model that can not effectively manage life-stage appearance variations. As illustrated in block (a) of Fig.~\ref{fig:gan_limit}, the model trained on celebrity life-stage data fails to generate a consistent identity as the 1 vs N Comparison. Additionally, we find that the expression mapping module tends to overfit to specific appearances, as depicted in Fig.~\ref{fig:gan_limit} block (b). Therefore, single appearance-based methods, like GANAvatar, are not suitable for life-stage head avatar modeling.

\begin{table}[htb]
\begin{center}
\caption{\small{\textbf{Quantitative Evaluation on TimeWalker-1.0.} We evaluate our method with two different protocols. The \textbf{Upper} table demonstrates {\textit{\#Protocol-1}} (1 vs 1 Comparison), while the \textbf{Lower} table shows {\textit{\#Protocol-2}} (1 vs N Comparison). \colorbox{colorbest}{Pink} indicates the best and \colorbox{colorsecond}{orange} indicates the second.}}

\resizebox{0.45\textwidth}{!}{
\begin{tabular}{c|ccc}
\toprule[1.5pt]
\textbf{Method} & PSNR$\uparrow$ & SSIM$\uparrow$ & LPIPS$\downarrow$  \\

\midrule
\multicolumn{4}{c}{\textbf{1 vs 1 Comparison}} \\
\midrule

\textit{INSTA}~\cite{zielonka2023instant}  & 20.68 & 0.697 & 0.299  \\
\textit{INSTA++} & \colorbox{colorsecond}{26.39} & \colorbox{colorsecond}{0.879} & \colorbox{colorsecond}{0.139}\\
\textit{Flash Avatar}~\cite{xiang2024flashavatar} & 22.14 & 0.771 & 0.267   \\
\textit{Ours} & \colorbox{colorbest}{27.28} & \colorbox{colorbest}{0.949} & \colorbox{colorbest}{0.071}   \\

\midrule
\multicolumn{4}{c}{\textbf{1 vs N Comparison}} \\
\midrule

\textit{Gaussian Surfels}~\cite{Dai2024GaussianSurfels}  & 26.98 & \colorbox{colorbest}{0.950} & 0.141  \\
\textit{Gaussian Surfels++}    & \colorbox{colorbest}{27.61} & 0.948 &  \colorbox{colorsecond}{0.134}   \\
\textit{INSTA}~\cite{zielonka2023instant} & 25.47 & 0.86 & 0.170  \\
\textit{Flash Avatar}~\cite{xiang2024flashavatar} & 24.9 & 0.848 & 0.165    \\
\textit{Ours} & \colorbox{colorsecond}{27.28} & \colorbox{colorsecond}{0.949} & \colorbox{colorbest}{0.071}  \\

\bottomrule[1.5pt]
\end{tabular}
}

\label{tab:exp_sub0}
\end{center}
\end{table}

\begin{table}[htb]
\begin{center}
\caption{\small{\textbf{Quantitative Evaluation on INSTA Data.} We evaluate our method with two different protocols. The \textbf{Upper} table demonstrates {\textit{\#Protocol-1}} (1 vs 1 Comparison), while the \textbf{Lower} table shows {\textit{\#Protocol-2}} (1 vs N Comparison). \colorbox{colorbest}{Pink} indicates the best and \colorbox{colorsecond}{orange} indicates the second.}}

\resizebox{0.45\textwidth}{!}{
\begin{tabular}{c|ccc}
\toprule[1.5pt]
\textbf{Method} & PSNR$\uparrow$ & SSIM$\uparrow$ & LPIPS$\downarrow$  \\

\midrule
\multicolumn{4}{c}{\textbf{1 vs 1 Comparison}} \\
\midrule

\textit{INSTA}~\cite{zielonka2023instant}  & 19.52 & 0.682 & 0.308  \\
\textit{INSTA++} & \colorbox{colorsecond}{27.74} & 0.915 & \colorbox{colorsecond}{0.09}\\
\textit{Flash Avatar}~\cite{xiang2024flashavatar} & 26.98 & \colorbox{colorsecond}{0.925} & 0.097   \\
\textit{Ours} & \colorbox{colorbest}{28.57} & \colorbox{colorbest}{0.966} & \colorbox{colorbest}{0.056}   \\

\midrule
\multicolumn{4}{c}{\textbf{1 vs N Comparison}} \\
\midrule

\textit{Gaussian Surfels}~\cite{Dai2024GaussianSurfels}  & 27.32 & \colorbox{colorbest}{0.969} & 0.121  \\
\textit{Gaussian Surfels++}    & 27.96 & 0.965 &  0.123   \\
\textit{INSTA}~\cite{zielonka2023instant} & 26.98 & 0.935 & \colorbox{colorsecond}{0.077}  \\
\textit{Flash Avatar}~\cite{xiang2024flashavatar} & \colorbox{colorsecond}{28.15} & 0.925 & 0.096    \\
\textit{Ours} & \colorbox{colorbest}{28.57} & \colorbox{colorsecond}{0.966} & \colorbox{colorbest}{0.056}  \\

\bottomrule[1.5pt]
\end{tabular}
}

\label{tab:exp_sub}
\end{center}
\end{table}

\begin{table}[htb]
\begin{center}
\caption{\small{\textbf{Quantitative Evaluation with Generative Method.} We evaluate our method with a generative method, \textbf{GANAvatar}~\cite{kabadayi2023ganavatar}, also with two different protocols: {\textit{\#Protocol-1}} (1 vs 1 Comparison) and {\textit{2}} (1 vs N Comparison). \colorbox{colorbest}{Pink} indicates the best.}}

\resizebox{0.45\textwidth}{!}{
\begin{tabular}{c|c|c|c|c}
\toprule[1.5pt]
\textbf{Dataset} & \textbf{Method} & \textbf{1 vs 1} & \textbf{1 vs N} & \textbf{Ours} \\
\midrule
\multirow{3}{*}{\textbf{TimeWalker-1.0}} & PSNR$\uparrow$ & 15.40 & 18.44 & \colorbox{colorbest}{27.28} \\
& SSIM$\uparrow$ & 0.809 & 0.852 & \colorbox{colorbest}{0.949} \\
& LPIPS$\downarrow$ & 0.227 & 0.151 & \colorbox{colorbest}{0.071} \\
\hline
\multirow{3}{*}{\textbf{INSTA Data}} & PSNR$\uparrow$ & 18.15 & 24.76 & \colorbox{colorbest}{28.57} \\
& SSIM$\uparrow$ & 0.832 & 0.900 & \colorbox{colorbest}{0.966} \\
& LPIPS$\downarrow$ & 0.172 & 0.0637 & \colorbox{colorbest}{0.056} \\
\bottomrule[1.5pt]
\end{tabular}
}

\label{tab:exp_gan}
\end{center}
\end{table}

{
    \begin{figure*}
        \centering
        \includegraphics[width=1.0\linewidth]{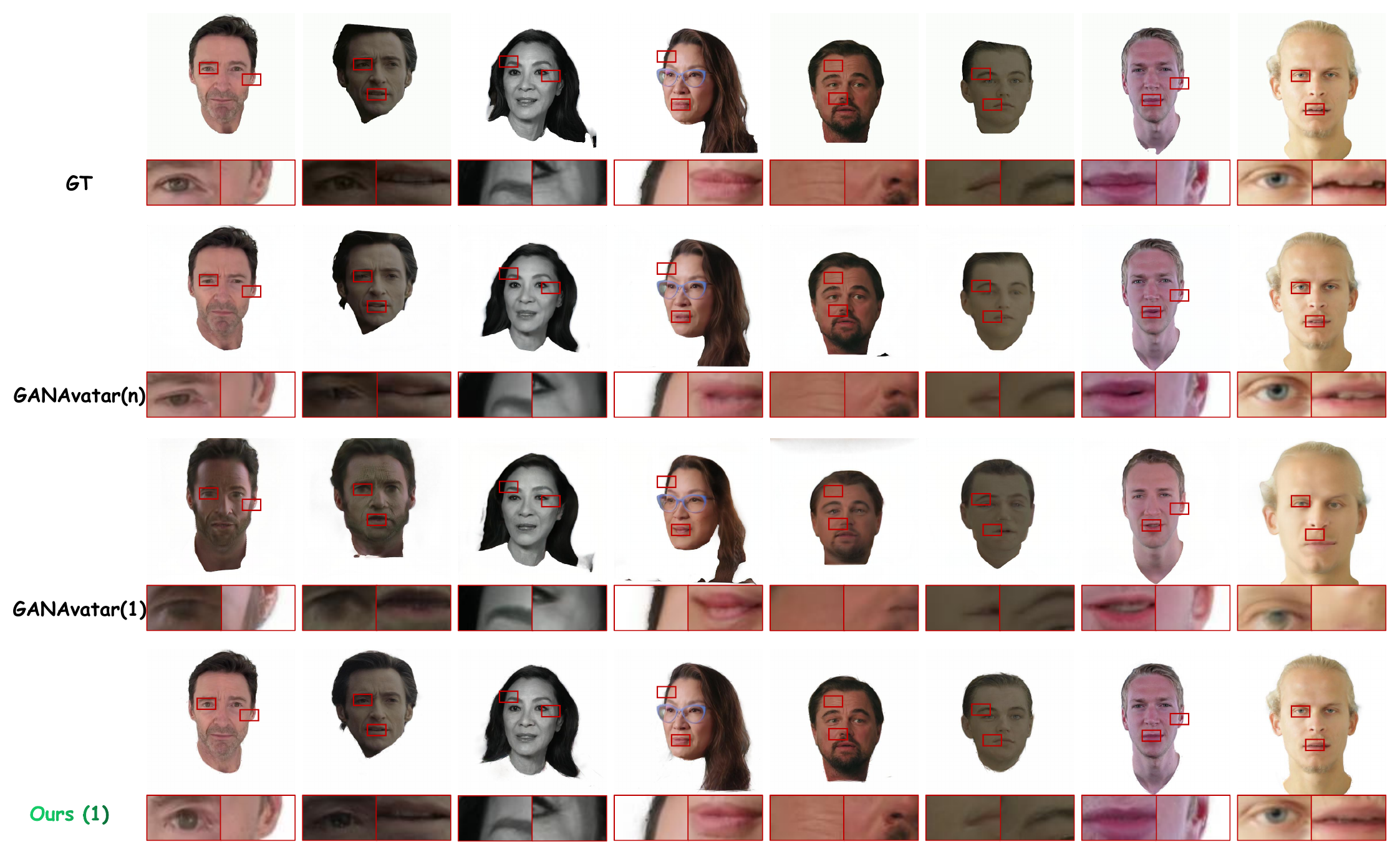}
        \caption{\small{\textbf{Qualitative Comparison with Generative Method.} Symbol $(1)$ in figure means train one model for each individual, symbol $(n)$ means train $N$ models for each individual, \textit{i.e.,} one model for each lifestage.}}
        \label{fig:comp_gan}
        \vspace{-0.5cm}
    \end{figure*}

    \begin{figure}
        \centering
        \includegraphics[width=\linewidth]{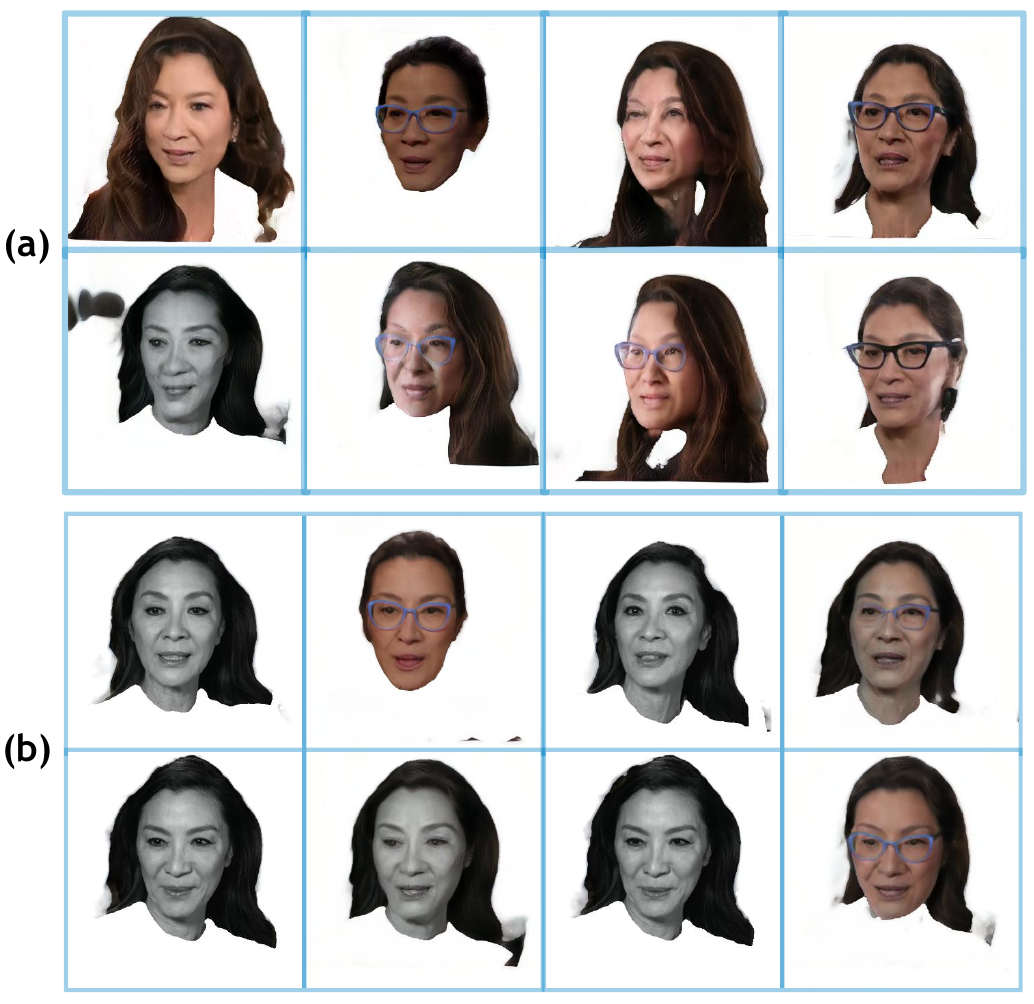}
  \captionsetup[figure]{hypcap=false}
  \captionof{figure}{\small{\textbf{Sampled Results of GANAvatar~\cite{kabadayi2023ganavatar}.} In the $\#$Protocal-1, (a) GANAvatar struggles to generate consistent identity, which is rooted in (b) the expression mapping network overfits on different lifestages. Even close expressions could generate significantly different head images.}}
        \label{fig:gan_limit}
    \end{figure}
}

\subsection{Comparison with One-Shot Head Avatar} 
\label{exp:compare_oneshot}
Recently, some approaches~\cite{ye2024real3d,li2024generalizable,chu2024gpavatar,chu2024generalizable,deng2025portrait4d,xu2024gphmv2} have effectively achieved head animation using a generalizable one-shot methodology. However, these methods primarily focus on momentary representations and do not incorporate personalized spaces over a lifelong scale, which distinguishes them from our framework. Nonetheless, to ensure a comprehensive experimental comparison and discussion, we have included two representative one-shot methods in our study, \textbf{GAGAvatar}~\cite{chu2024generalizable} and \textbf{GPHM}~\cite{xu2024gphmv2}.

\subsubsection{Compare with GAGAvatar}

We selected \textbf{GAGAvatar}~\cite{chu2024generalizable} as it also utilizes Gaussian Splatting as its underlying representation. The GAGAvatar model can efficiently generate Gaussian parameters through a straightforward forward pass. It firstly predicts global Gaussian features by simultaneously enhancing the local feature plane extracted from DINOv2~\cite{oquab2023dinov2}, then further incorporates an expression branch to precisely convey expression-related information. In this experiment, we choose three distinct identities to perform two types of comparisons. Note that identities selected for this analysis differ from those discussed in Sec.~\ref{exp:compare_sota}, leading to variations in the evaluation metrics. Initially, we evaluate the pretrained model from GAGAvatar directly, given its generalizability. The initial image of each lifestage within an identity serves as the source image, while the testset acts as driving images. Subsequently, we finetune the pretrained model using the data specific to each identity, employing identical training and testing sets as per our pipeline. 

\noindent\textbf{\textit{Rendering Quality.}}
As shown in Tab.~\ref{tab:exp_oneshot}, our method outperforms both the pretrained and finetuned models of GAGAvatar in terms of SSIM and LPIPS. While GAGAvatar exhibits strong generalizability during one-shot inference, attributed to the robustness of its vision fundamental model, its performance declines when finetuned for a specific identity. As Fig.~\ref{fig:comp_oneshot} demonstrated, GAGAvatar typically produces high-fidelity results; however, it occasionally struggles to maintain identity consistency, leading to identity switching during animation. This issue is exacerbated in the finetuned model, highlighting the inadequacy of merely supplying lifelong data to achieve a more effective personalized space and robust identity consistency.

\noindent\textbf{\textit{Identity Consistency.}}
To quantitatively evaluate the identity consistency on both TimeWalker and GAGAvatar, we apply the face recognition method from ArcFace~\cite{deng2019arcface}. Specifically, we encode the images of ground truth and prediction with pretrained Arcface model, and then calculate the distance of the L2 norm encoded feature, which is finally transferred to the score. As Tab.~\ref{tab:exp_arcface} shown, our method achieves constantly better ID consistency. The second row in Fig.~\ref{fig:comp_oneshot} also shows a similar phenomenon. These results highlight that $(1)$ simply feeding lifelong data is insufficient to achieve robust identity consistency; and $(2)$ the critical importance of learning a lifelong personalized representation space that effectively balances shared characteristics while preserving individual uniqueness -- a capability notably absent in current one-shot talking head methods.

\subsubsection{Compare with GPHM}
GPHM introduces a 3D Gaussian Parametric Head Model that combines the concept of 3DMM with 3D Gaussian representation, and use large-scale pretrainning for prior learning, and separate encoders with correspondent learnable latent codes for explicitly reconstructing and storing facial expression and non-face motion of key frames as the expression basis (among 50 basis per id, and id is moment-level). In contrast, our method realizes translation among different lifestage by introducing the Dynamo module, which mainly contains multiple learnable 3D hashgrid latents for learning the local feature, and residual embedding for learning global residual of each lifestage. Although both works leverage the concept of 3DMM, it is different in how to extend the 3DMM, and how to realize full scale animation. Since GPHM~\cite{xu2024gphmv2} is not open-sourced, we are unable to conduct direct, fair comparisons under identical experimental settings. To the best of our knowledge, the only feasible comparison is through metric alignment with a common baseline (INSTA), which is shown in Tab.~\ref{tab:gphm_ours_compare}. Both methods achieve comparable quantitative performance, with our method showing a higher SSIM. 
\begin{table}[h]
\centering
\caption{\small{\textbf{Comparison of 3D Gaussian Parametric Head Model(GPHM) and Ours. \colorbox{colorbest}{Pink} indicates the best and \colorbox{colorsecond}{orange} indicates the second. Note that, our training data volume is far less than GPHM.}}}
\label{tab:gphm_ours_compare}
\begin{tabular}{c|ccc}
\toprule[1.5pt]
\textbf{Method} & PSNR$\uparrow$ & SSIM$\uparrow$ & LPIPS$\downarrow$ \\
\midrule
\textit{GPHM wo FT motion} & 28.1  & 0.93  & \colorbox{colorsecond}{0.046} \\
\textit{GPHM}              & \colorbox{colorbest}{28.9}  & \colorbox{colorsecond}{0.94}  & \colorbox{colorbest}{0.041} \\
\textit{Ours}              & \colorbox{colorsecond}{28.57} & \colorbox{colorbest}{0.966} & 0.056 \\
\bottomrule[1.5pt]
\end{tabular}
\end{table}

\begin{table}[htb]
\begin{center}
\caption{\small{\textbf{Quantitative Evaluation with One-Shot Method.} We evaluate our method with a generalizable one-shot method, \textbf{GAGAvatar}~\cite{chu2024generalizable} \colorbox{colorbest}{Pink} indicates the best.}}

\resizebox{0.45\textwidth}{!}{
\begin{tabular}{c|c|c|c|c}
\toprule[1.5pt]
\textbf{Dataset} & \textbf{Method} & \textbf{Pretrained} & \textbf{Finetuned} & \textbf{Ours} \\
\midrule
\multirow{3}{*}{\textbf{TimeWalker-1.0}} & PSNR$\uparrow$ & \colorbox{colorbest}{28.69} & 26.78 & 28.33 \\
& SSIM$\uparrow$ & 0.908 & 0.898 & \colorbox{colorbest}{0.96} \\
& LPIPS$\downarrow$ & 0.08 & 0.09 & \colorbox{colorbest}{0.06} \\
\bottomrule[1.5pt]
\end{tabular}
}

\label{tab:exp_oneshot}
\end{center}
\end{table}

\begin{table}[htb]
\begin{center}
\caption{\small{\textbf{Comparison of Identity Consistency Score. } We compare our method with \textbf{GAGAvatar}~\cite{chu2024generalizable}, with the metric of identity consistency score computed from Arcface. A higher scroe indicates superior performance. \colorbox{colorbest}{Pink} indicates the best.}}

\resizebox{0.45\textwidth}{!}{
\begin{tabular}{c|c|c|c|c}
\toprule[1.5pt]
\textbf{Method} & \textbf{ID1} & \textbf{ID2} & \textbf{ID3} & \textbf{Mean}\\
\midrule
$GAGAvatar_{Pretrained}$ & 84 & 77.7 & 87.6 & 83.7 \\
$GAGAvatar_{Finetuned}$ & 77.6 & 76.7 & 86.0 & 80.7 \\
Ours & \colorbox{colorbest}{88.8} & \colorbox{colorbest}{82.7} & \colorbox{colorbest}{87.8} & \colorbox{colorbest}{86.8} \\
\bottomrule[1.5pt]
\end{tabular}
}

\label{tab:exp_arcface}
\end{center}
\end{table}

{
    \begin{figure}
        \centering
        \includegraphics[width=0.8\linewidth]{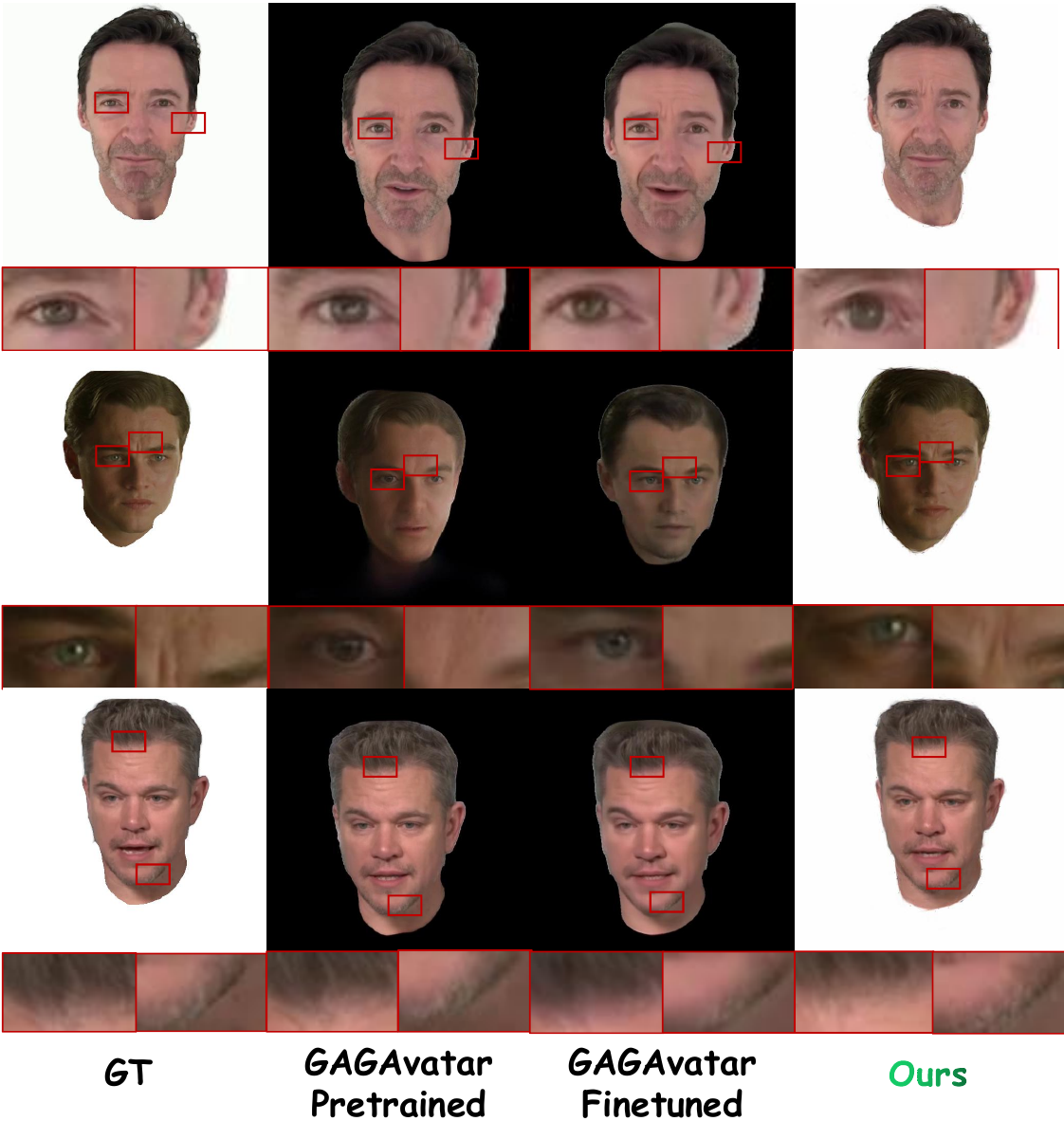}
        \captionof{figure}{\small{\textbf{Qualitative comparison with One-Shot Method.} We compare our method with the results obtained from both pretrained and finetuned versions of GAGAvatar. It is important to note that the training settings are consistent between our approach and the finetuned models.}}
        \label{fig:comp_oneshot}
    \end{figure}
}

\section{Application}\label{application}
{
    \begin{figure}
        \centering
        \vspace{-0.25cm}
        \includegraphics[width=1\linewidth]{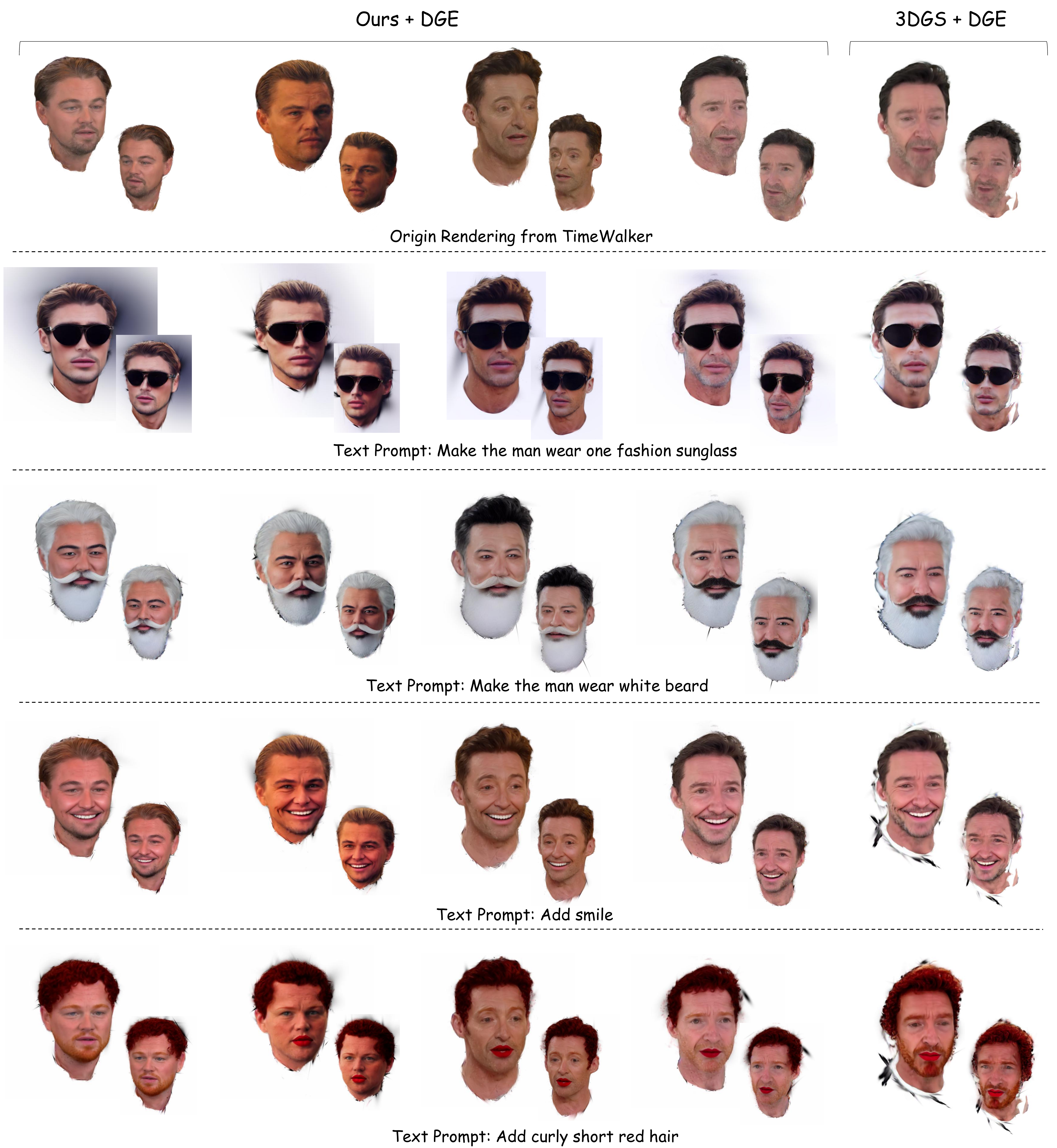}
        \vspace{-0.35cm}
        \caption{\small{\textbf{3D Editing as a Downstream Task: Visual Results with Our Model and DGE.}} We present the editing outcomes elicited by a variety of text prompts. The upper two examples illustrate the capability of DGE~\cite{chen2024dge} method to introduce new elements to the human head model, while the lower two examples illustrate how it functions when altering the attributes of head components. We compare the editing results between our models with DGE, and the original 3DGS with DGE.}
        \label{fig:dge}
        \vspace{-0.3cm}
    \end{figure}
}
\noindent\textbf{3D Editting.} To confirm the validity of our neural parametric model that produces 3D consistent and personalized outcomes, we conduct additional experiments in the 3D head editing application. We evaluate the results of our Gaussian Surfels-based approach by referring to DGE~\cite{chen2024dge}, which performs direct 3DGS editing with a pre-trained diffusion model while maintaining multi-view consistency. During the optimization, we switch the representation to Gaussian Surfels and keep the other settings unchanged from the original implementation. Fig.~\ref{fig:dge} presents a visual comparison of the re-rendered results, based on various editing prompts, highlighting the 3D consistency achieved by our method. The direct editing method fails to disentangle animated expressions (as can be observed from the visual result of the text prompt: "Add smile"), and it also undergoes other unforeseen changes, such as alterations in skin color. In contrast, our method executes disentangled and fine-grained multi-dimensional animation. 
Focusing on the final column of the illustration, the editing performance of the original implementation (DGE with 3DGS) within our TimeWalker-1.0 dataset exhibits shortcomings, primarily attributed to the following factors. Primarily, noticeable blurring is evident in both the original rendering and the edited output, particularly pronounced in the facial features such as the eyes and mouth. Given that 3DGS is designed to handle static objects rather than dynamic data sequences, areas that undergo frequent modifications lead to inconsistencies and produce blurred outcomes.
Furthermore, the inadequate diversity in camera pose distribution within the training data at a single lifestage results in depth ambiguity, notably observable in the neck region. This issue persists after 3D editing, manifesting as discontinuities and dark artifacts encircling the neck. In contrast, our model effectively addresses this challenge through the automated interpolation of data from various lifestages during training, resulting in a more coherent geometric representation.

\section{Discussion}

\subsection{Large Scale Training}\label{large_scale_training}
Large models trained on large-scale datasets have gained significant attention from the community. As an {\textit{orthogonal}} direction to our personalized, lifelong modeling framework under limited data, we further investigate large-scale models trained on massive datasets. In the context of human avatars, such models learn rich priors about human geometry, appearance, and motion from either controlled laboratory captures or large-scale in-the-wild collections. Existing approaches broadly fall into two categories: 1) models trained on specialized laboratory datasets or curated web-scale human data for dedicated head-avatar reconstruction, and 2) general-purpose models trained on diverse, large-scale datasets without avatar-specific supervision.
Three natural questions arise: {\textit{can these large-scale models effectively support lifelong modeling? What are their current limitations? And how to combine the strengths of our method and large-scale models?}}

To answer this, we examine both categories through targeted analysis and preliminary experiments, revealing their capabilities and shortcomings in the lifelong setting.

{
  \begin{figure}
  \centering
  \includegraphics[width=0.5\textwidth]{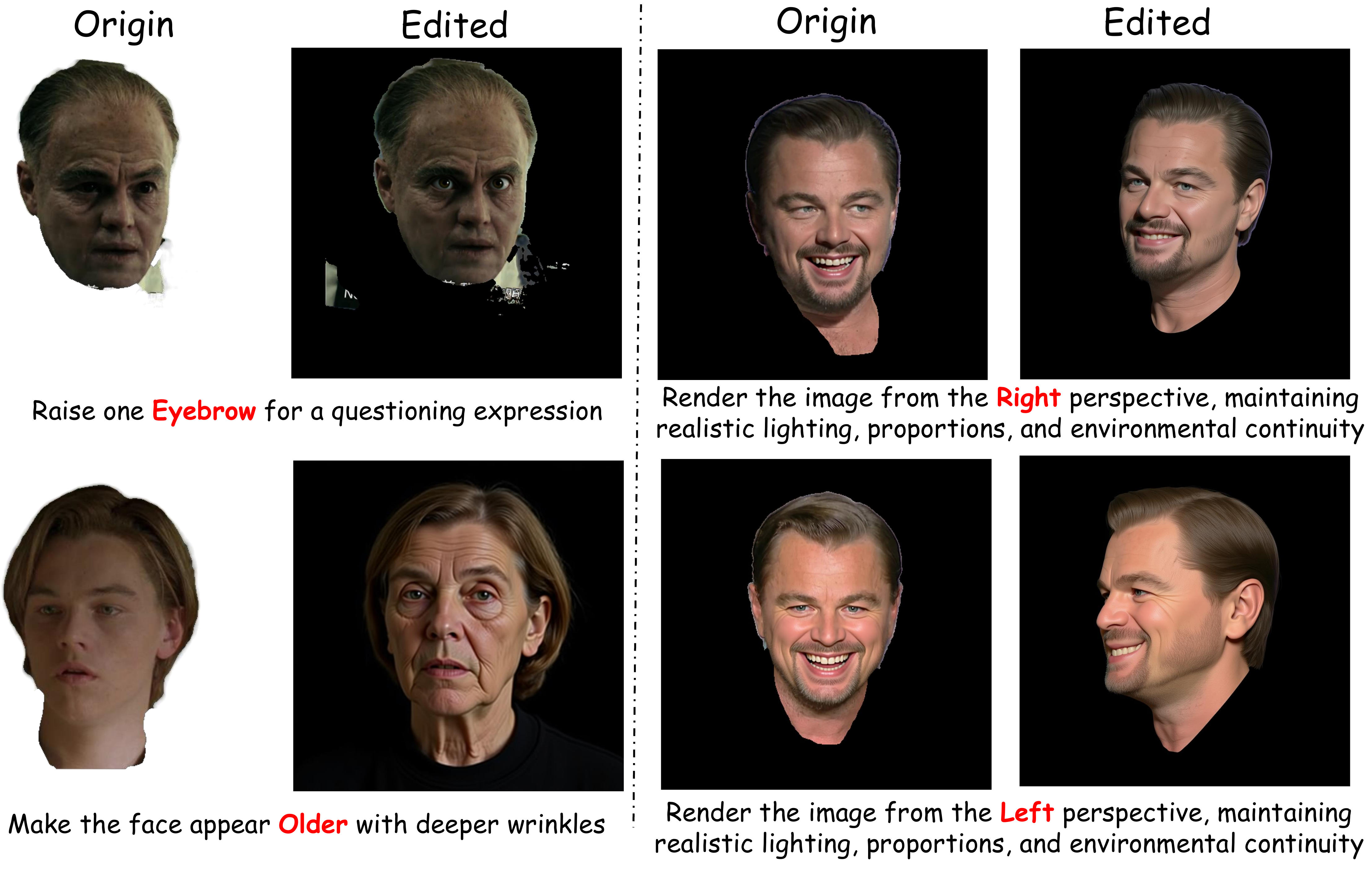}
  \captionsetup[figure]{hypcap=false}
  \captionof{figure}{\small{\textbf{Failure Case of Flux Editting.}  We visualize the failure cases of directly using the Flux~\cite{labs2025flux1kontextflowmatching} pretrained model for image editing, including ID inconsistency, poor prompt adherence, and gender inconsistency. }}
  \label{fig:flux_edit}
  \end{figure}
}

{
  \begin{figure}
  \centering
  \includegraphics[width=0.35\textwidth]{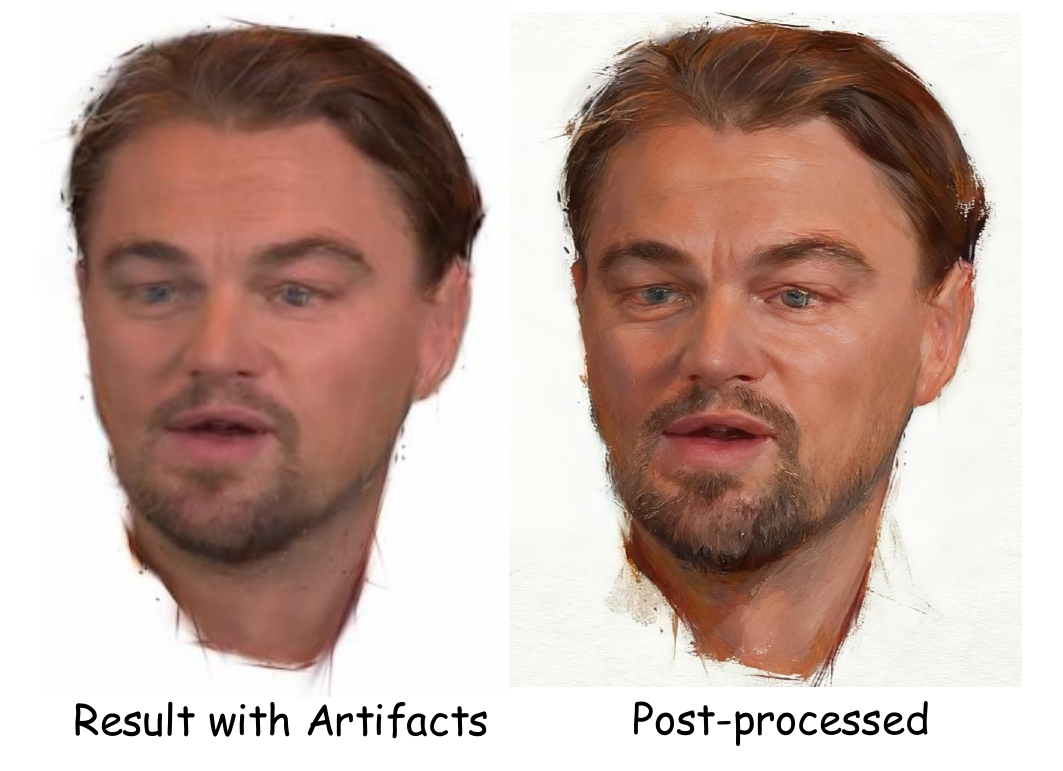}
  \captionsetup[figure]{hypcap=false}
  \captionof{figure}{\small{\textbf{Result of Post Processing.}  We leveraged the DouBao API tools for refinement in the form of image editing. }}
  \label{fig:post_process}
  \end{figure}
}

{
  \begin{figure*}[!h]
  \centering
  \includegraphics[width=\textwidth]{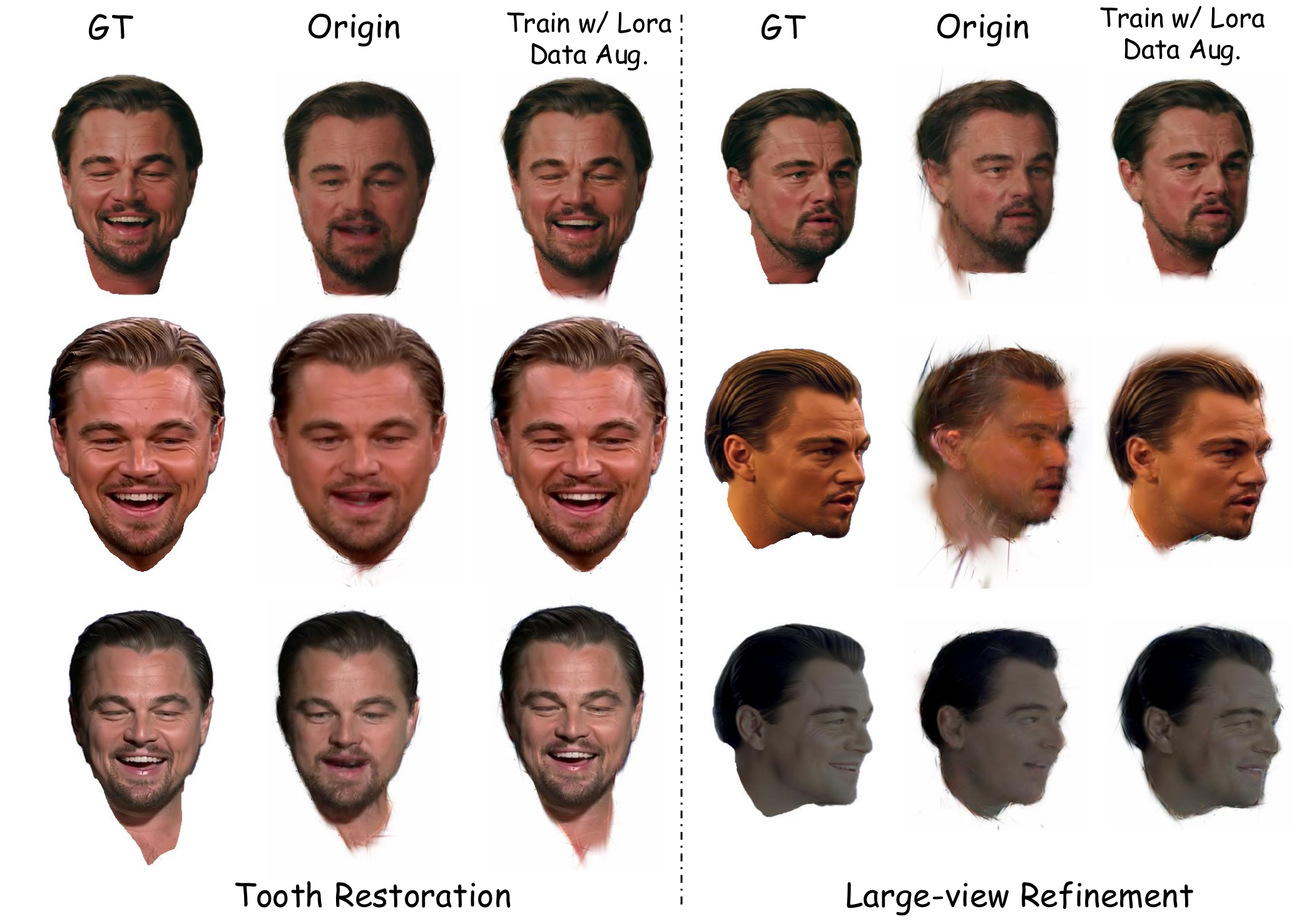}
  \captionsetup[figure]{hypcap=false}
  \captionof{figure}{\small{\textbf{Preliminary results on combining our method with large-scale pretraining.}  We present a qualitative comparison between the original renderings and those generated by models trained with LoRA data augmentation. We utilized a large-scale pretrained model to augment the  dataset then feed it to train our model. The rendering results demonstrate that, by leveraging the human prior of the pretrained model, our approach achieves improved rendering quality for tooth renderings and extreme-profile faces. }}
  \label{fig:qwen_img_lora}
  \end{figure*}
}
\subsubsection{Large-scale Head Avatar Generative Models}

Large-scale head avatar generative models are trained on massive head-specific datasets, enabling them to learn head-related prior knowledge for 3D head avatar one-shot generation~\cite{chu2024generalizable} or 3D Parametric Head Model capabilities~\cite{xu2024gphmv2}. However, they still exhibit certain limitations compared to our approach. As demonstrated in the Sec.~\ref{exp:compare_oneshot}, GAGAvatar, as a one-shot generation pipeline, exhibit issues in maintaining identity consistency, and merely feeding large-scale lifelong data is insufficient to achieve robust identity consistency. In terms of generation quality, the two methods we compared in experiments—GAGAvatar~\cite{chu2024generalizable} and GPHM~\cite{xu2024gphmv2}—did not significantly outperform our method in metrics despite leveraging large-scale pretraining; instead, their performance was generally comparable. We attribute this to the fact that, in addition to increasing data volume, appropriate and targeted model designs are equally essential. Our proposed framework demonstrates that a well-designed, non-pre-trained architecture, which harnesses the potential of a compact personalized space, can achieve competitive, and in some cases superior performance, underscoring the effectiveness and robustness of our approach.

\subsubsection{Large-scale General Pretraining Models}

Integrating the capabilities of pretrained models to further enhance personalized avatar construction is identified as our future research direction. Herein, we present relatively simple preliminary attempts, divided into two parts: 1) Leveraging pretrained models to generate additional data for pre-training data augmentation; 2) Performing post-processing optimization on the rendered results of trained personalized avatars to improve quality. Both attempts were conducted on a single identity (Leonardo DiCaprio).

{
  \begin{figure*}[!h]
  \centering
  \includegraphics[width=1\textwidth]{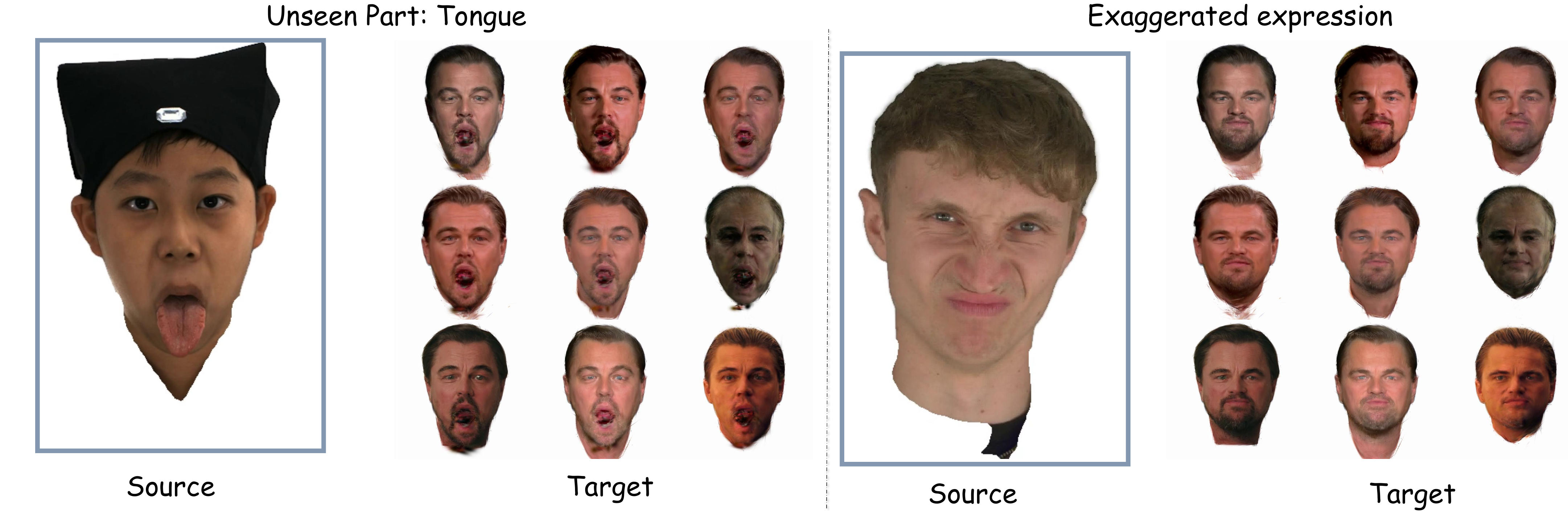}
  \captionsetup[figure]{hypcap=false}
  \captionof{figure}{\small{\textbf{Limitations.}  Our method fails to model unseen parts like tongue during animation, and performs moderate expressions when driven by exaggerated expressions. (Source from Renderme-360~\cite{pan2024renderme} and NerSemble~\cite{kirschstein2023nersemble})}}
  \label{fig:limitation}
  \end{figure*}
}

\noindent\textbf{Large-scale Pretraining Models for Post-Processing.} When the data for a specific appearance is scarce, the generated results of our trained model sometimes exhibit artifacts in high-frequency regions, such as the eyes and hair region. We attempted to use a large model to directly refine the generated results, eliminating artifacts and improving quality. As shown in Fig.~\ref{fig:post_process}, the current large model did not eliminate the artifacts; instead, it enhanced the overall clarity of the results while retaining the artifacts, similar to a super-resolution effect. This result also indicates that leveraging pretrained models to enhance personalized space construction requires appropriate design rather than a simple combination. 

\noindent\textbf{Large-scale Pretraining Models for Data Augmentation.} The in-the-wild data of Leonardo DiCaprio exhibits data distribution imbalance, particularly lacking data of extreme-profile faces or smiles that reveal the internal appearance of the mouth. To address this issue, we fine-tuned the Qwen-Image~\cite{wu2025qwenimage} model using LoRA~\cite{hu2022lora} on all 15,000+ training images of Leonardo, with an image resolution of 512×512. For caption annotation, we converted camera parameters into the subject’s head pose and estimated the age corresponding to each of Leonardo appearances using a Vision-Language Model~\cite{bai2025qwen25vl}. All generated captions thus include both age and head pose information. After LoRA fine-tuning, we attempted to generate data of Leonardo with various head poses and smile expressions (laughing expressions) , and additionally, we tried generating appearance data of new ages beyond the age range in the training data. Interestingly, our generation results show that the LoRA-fine-tuned model struggled to generate new data beyond the head pose distribution or age range of the training data. In contrast, the untrained model~\cite{labs2025flux1kontextflowmatching} could generate extreme-profile face data and age-varied data, but exhibited substantial degradation in identity consistency(Fig.~\ref{fig:flux_edit}). Through the aforementioned training, we obtained approximately 3,000 new images of DiCaprio, which were incorporated into the training process. The visualization comparisons of the model trained with augmented data are presented in Figure~\ref{fig:qwen_img_lora}. Based on the rendering results, models trained with data augmentation achieve an improvement in rendering quality for long-tailed scenarios in in-the-wild data, such as tooth exposure and extreme-profile faces. These preliminary findings highlight a promising future direction for the community: leveraging large-scale pretrained generative models not as direct lifelong models, but as targeted data augmentors to enrich personalized avatar space.

\subsection{Limitations}\label{limitation}
There are three main limitations of our pipeline: $1)$ due to the FLAME base, our personalized avatar struggles with exaggerated expressions and unseen parts like the tongue, causing artifacts around the mouth (Fig.~\ref{fig:limitation}). $2)$ Large-scale in-the-wild datasets introduce noise and imbalance, such as complex backgrounds leading to imprecise matting and additional noise around the head edges. Consequently, FLAME parameter accuracy is lower compared to standardized inputs. These challenges underscore the need for methods that create high-fidelity head avatars with reduced dependence on preprocessing. $3)$ Due to data constraints, our study provides only a preliminary exploration of lifelong perspectives rather than fully spanning infancy to old age. We hope this work establishes a baseline, inspiring more effective approaches toward comprehensive lifelong personalized spaces, potentially leveraging pre-trained generative models' prior knowledge of head structures.

\subsection{Broader Impact $\&$ Ethics Statement}\label{broader_impact}

In this study, our objective is to accurately generate images of an individual, focusing solely on rendering head avatars and seamlessly altering their appearance within a predefined set of lifestages. It is noteworthy that our work does not endeavor to create fictitious appearance; instead, we strive to learn a compact and steerable latent space for an individual from limited, sparse, unstructured data, faithfully represent the subject's appearance and geometry, and enable steerable animation.

{\subsubsection{Broader Impact.}} $1)$ {\textit{A new testbed, baseline, and direction for the field.} Currently, the field mainly focuses on momentary-level 3D face/head reconstruction.  We hope TimeWalker, its benchmarks, dataset, and our open-source code base will become a test bed for future work on lifelong avatar modeling, which brings challenges as well as new opportunities for head attribute disentanglement, appearance modeling, and geometry reconstruction from restricted data.  $2)$ {\textit{3D prior for downstream tasks.}} As our learned space facilitates ID consistency across views and ages, it could be utilized to boost downstream tasks' geometric coherency, as demonstrated in the 3D Editing application section. $3)$ {\textit{Social Impacts.}} In scenarios like virtual communication, our method could power more expressive and consistent digital avatars across age spans, enhancing telepresence. In historical archiving and cultural preservation, it could be used to reconstruct and visualize the appearance of individuals across time using limited photographic evidence. 

{\subsubsection{Ethics Statement.}} Due to ID sensitivity, we strictly adhere to the non-commercial license for the dataset and will implement additional safeguards before its release. While the real and rich human head assets generated by our method could be misused for deceptive applications, such as fake talking head video generation, they also have the potential to be leveraged for positive uses, such as training robust deepfake detection models. To minimize risks, we will only make the dataset available to individuals or institutions who have reviewed and agreed to our proposed data usage agreement, ensuring ethical and responsible use.


\clearpage

{
    \small
    \bibliographystyle{ieeenat_fullname}
    \bibliography{main}

@String(IJCV = {Int. J. Comput. Vis.})

@String(CVPR= {IEEE Conf. Comput. Vis. Pattern Recog.})

@String(ICCV= {Int. Conf. Comput. Vis.})

@String(ECCV= {Eur. Conf. Comput. Vis.})

@String(TOG= {ACM Trans. Graph.})

@String(IJCV  = {IJCV})

@String(CVPR  = {CVPR})

@String(ICCV  = {ICCV})

@String(ECCV  = {ECCV})

@String(TOG   = {ACM TOG})

@inproceedings{mildenhall2020nerf,
  year={2020},
  title={Nerf: Representing scenes as neural radiance fields for view synthesis},
  author={Mildenhall, Ben and Srinivasan, Pratul P and Tancik, Matthew and Barron, Jonathan T and Ramamoorthi, Ravi and Ng, Ren},
  booktitle={ECCV}
}

@inproceedings{yu2020humbi,
  year={2020},
  title={Humbi: A large multiview dataset of human body expressions},
  author={Yu, Zhixuan and Yoon, Jae Shin and Lee, In Kyu and Venkatesh, Prashanth and Park, Jaesik and Yu, Jihun and Park, Hyun Soo},
  booktitle={CVPR}
}

@article{yu2021bisenet,
  year={2021},
  title={Bisenet v2: Bilateral network with guided aggregation for real-time semantic segmentation},
  author={Yu, Changqian and Gao, Changxin and Wang, Jingbo and Yu, Gang and Shen, Chunhua and Sang, Nong},
  journal={IJCV}
}

@article{FLAME:SiggraphAsia2017, 
  year={2017},
  title = {Learning a model of facial shape and expression from {4D} scans}, 
  author = {Li, Tianye and Bolkart, Timo and Black, Michael. J. and Li, Hao and Romero, Javier}, 
  journal = {TOG}
}

@inproceedings{3dmm,
  year={1999},
  title={A morphable model for the synthesis of 3D faces},
  author={Blanz, Volker and Vetter, Thomas},
  booktitle={CGIT}
}

@article{mildenhall2021nerf,
  year={2021},
  title={Nerf: Representing scenes as neural radiance fields for view synthesis},
  author={Mildenhall, Ben and Srinivasan, Pratul P and Tancik, Matthew and Barron, Jonathan T and Ramamoorthi, Ravi and Ng, Ren},
  journal={Communications}
}

@article{muller2022instant,
  year={2022},
  title={Instant neural graphics primitives with a multiresolution hash encoding},
  author={M{\"u}ller, Thomas and Evans, Alex and Schied, Christoph and Keller, Alexander},
  journal={TOG}
}

@inproceedings{yu2021plenoctrees,
  year={2021},
  title={Plenoctrees for real-time rendering of neural radiance fields},
  author={Yu, Alex and Li, Ruilong and Tancik, Matthew and Li, Hao and Ng, Ren and Kanazawa, Angjoo},
  booktitle={ICCV}
}

@inproceedings{hong2022headnerf,
  year={2022},
  title={Headnerf: A real-time nerf-based parametric head model},
  author={Hong, Yang and Peng, Bo and Xiao, Haiyao and Liu, Ligang and Zhang, Juyong},
  booktitle={CVPR}
}

@inproceedings{zheng2022avatar,
  year={2022},
  title={Im avatar: Implicit morphable head avatars from videos},
  author={Zheng, Yufeng and Abrevaya, Victoria Fern{\'a}ndez and B{\"u}hler, Marcel C and Chen, Xu and Black, Michael J and Hilliges, Otmar},
  booktitle={CVPR}
}

@article{zheng2022pointavatar,
  year={2022},
  title={PointAvatar: Deformable Point-based Head Avatars from Videos},
  author={Zheng, Yufeng and Yifan, Wang and Wetzstein, Gordon and Black, Michael J and Hilliges, Otmar},
  journal={arXiv}
}

@article{livingstone2018ryerson,
  year={2018},
  title={The Ryerson Audio-Visual Database of Emotional Speech and Song (RAVDESS): A dynamic, multimodal set of facial and vocal expressions in North American English},
  author={Livingstone, Steven R and Russo, Frank A},
  journal={PO}
}

@inproceedings{deng2019arcface,
  title={Arcface: Additive angular margin loss for deep face recognition},
  author={Deng, Jiankang and Guo, Jia and Xue, Niannan and Zafeiriou, Stefanos},
  booktitle={CVPR},
  year={2019}
}

@inproceedings{yang2020facescape,
  year={2020},
  title={Facescape: a large-scale high quality 3d face dataset and detailed riggable 3d face prediction},
  author={Yang, Haotian and Zhu, Hao and Wang, Yanru and Huang, Mingkai and Shen, Qiu and Yang, Ruigang and Cao, Xun},
  booktitle={CVPR}
}

@inproceedings{cosker2011facs,
  year={2011},
  title={A FACS valid 3D dynamic action unit database with applications to 3D dynamic morphable facial modeling},
  author={Cosker, Darren and Krumhuber, Eva and Hilton, Adrian},
  booktitle={ICCV}
}

@InProceedings{Gafni_2021_CVPR,
  year={2021},
    author    = {Gafni, Guy and Thies, Justus and Zollh{\"o}fer, Michael and Nie{\ss}ner, Matthias},
    title     = {Dynamic Neural Radiance Fields for Monocular 4D Facial Avatar Reconstruction},
    booktitle = {CVPR}
}

@INPROCEEDINGS{9577855,
  year={2021},
  author={Yenamandra, Tarun and Tewari, Ayush and Bernard, Florian and Seidel, Hans-Peter and Elgharib, Mohamed and Cremers, Daniel and Theobalt, Christian},
  booktitle={CVPR}, 
  title={i3DMM: Deep Implicit 3D Morphable Model of Human Heads}, 
  
  }

@inproceedings{ffhq,
year={2019},
  title={A style-based generator architecture for generative adversarial networks},
  author={Karras, Tero and Laine, Samuli and Aila, Timo},
  booktitle={CVPR}
}

@inproceedings{schoenberger2016sfm,
  year={2016},
    author={Sch\"{o}nberger, Johannes Lutz and Frahm, Jan-Michael},
    title={Structure-from-Motion Revisited},
    booktitle={CVPR},
}

@inproceedings{chan2022efficient,
  year={2022},
  title={Efficient geometry-aware 3D generative adversarial networks},
  author={Chan, Eric R and Lin, Connor Z and Chan, Matthew A and Nagano, Koki and Pan, Boxiao and De Mello, Shalini and Gallo, Orazio and Guibas, Leonidas J and Tremblay, Jonathan and Khamis, Sameh and others},
  booktitle={CVPR}
}

@inproceedings{Dai2024GaussianSurfels,
  author    = {Dai, Pinxuan and Xu, Jiamin and Xie, Wenxiang and Liu, Xinguo and Wang, Huamin and Xu, Weiwei},
  title     = {High-quality Surface Reconstruction using Gaussian Surfels},
  publisher = {Association for Computing Machinery},
  booktitle = {SIGGRAPH 2024 Conference Papers},
  year      = {2024},
  doi       = {10.1145/3641519.3657441}
}

@Article{kerbl3Dgaussians,
      author       = {Kerbl, Bernhard and Kopanas, Georgios and Leimk{\"u}hler, Thomas and Drettakis, George},
      title        = {3D Gaussian Splatting for Real-Time Radiance Field Rendering},
      journal      = {ACM Transactions on Graphics},
      month        = {July},
      year         = {2023},
      url          = {https://repo-sam.inria.fr/fungraph/3d-gaussian-splatting/}
}

@article{kazhdan2013screened,
  title={Screened poisson surface reconstruction},
  author={Kazhdan, Michael and Hoppe, Hugues},
  journal={ACM Transactions on Graphics (ToG)},
  year={2013},
  publisher={ACM New York, NY, USA}
}

@inproceedings{johnson2016perceptual,
  title={Perceptual losses for real-time style transfer and super-resolution},
  author={Johnson, Justin and Alahi, Alexandre and Fei-Fei, Li},
  booktitle={ECCV},
  year={2016},
  organization={Springer}
}

@article{krizhevsky2012imagenet,
  title={Imagenet classification with deep convolutional neural networks},
  author={Krizhevsky, Alex and Sutskever, Ilya and Hinton, Geoffrey E},
  journal={Advances in neural information processing systems},
  year={2012}
}

@inproceedings{zielonka2023instant,
  title={Instant volumetric head avatars},
  author={Zielonka, Wojciech and Bolkart, Timo and Thies, Justus},
  booktitle={CVPR},
  year={2023}
}

@inproceedings{xiang2024flashavatar,
    author    = {Jun Xiang and Xuan Gao and Yudong Guo and Juyong Zhang},
    title     = {FlashAvatar: High-fidelity Head Avatar with Efficient Gaussian Embedding},
    booktitle = {CVPR},
    year      = {2024},
}

@incollection{snavely2006photo,
  title={Photo tourism: exploring photo collections in 3D},
  author={Snavely, Noah and Seitz, Steven M and Szeliski, Richard},
  booktitle={ACM siggraph 2006 papers},
  year={2006}
}

@inproceedings{liang2016head,
  title={Head reconstruction from internet photos},
  author={Liang, Shu and Shapiro, Linda G and Kemelmacher-Shlizerman, Ira},
  booktitle={ECCV},
  year={2016},
  organization={Springer}
}

@inproceedings{weng2023personnerf,
  title={Personnerf: Personalized reconstruction from photo collections},
  author={Weng, Chung-Yi and Srinivasan, Pratul P and Curless, Brian and Kemelmacher-Shlizerman, Ira},
  booktitle={CVPR},
  year={2023}
}

@inproceedings{eftekhar2021omnidata,
  title={Omnidata: A Scalable Pipeline for Making Multi-Task Mid-Level Vision Datasets From 3D Scans},
  author={Eftekhar, Ainaz and Sax, Alexander and Malik, Jitendra and Zamir, Amir},
  booktitle={ICCV},
  year={2021}
}

@misc{kabadayi2023ganavatar,
      title={GAN-Avatar: Controllable Personalized GAN-based Human Head Avatar}, 
      author={Berna Kabadayi and Wojciech Zielonka and Bharat Lal Bhatnagar and Gerard Pons-Moll and Justus Thies},
      year={2023},
      eprint={2311.13655},
      archivePrefix={arXiv},
      primaryClass={cs.CV}
}

@article{wu20234dgaussians,
  title={4D Gaussian Splatting for Real-Time Dynamic Scene Rendering},
  author={Wu, Guanjun and Yi, Taoran and Fang, Jiemin and Xie, Lingxi and Zhang, Xiaopeng and Wei Wei and Liu, Wenyu and Tian, Qi and Wang Xinggang},
  journal={arXiv preprint arXiv:2310.08528},
  year={2023}
}

@article{pan2024renderme,
  title={RenderMe-360: A Large Digital Asset Library and Benchmarks Towards High-fidelity Head Avatars},
  author={Pan, Dongwei and Zhuo, Long and Piao, Jingtan and Luo, Huiwen and Cheng, Wei and Wang, Yuxin and Fan, Siming and Liu, Shengqi and Yang, Lei and Dai, Bo and others},
  journal={Advances in Neural Information Processing Systems},
  year={2024}
}

@inproceedings{paysan20093d,
  title={A 3D face model for pose and illumination invariant face recognition},
  author={Paysan, Pascal and Knothe, Reinhard and Amberg, Brian and Romdhani, Sami and Vetter, Thomas},
  booktitle={2009 sixth IEEE international conference on advanced video and signal based surveillance},
  year={2009},
  organization={Ieee}
}

@inproceedings{giebenhain2023nphm,
 author={Simon Giebenhain and Tobias Kirschstein and Markos Georgopoulos and  Martin R{\"{u}}nz and Lourdes Agapito and Matthias Nie{\ss}ner},
 title={Learning Neural Parametric Head Models},
 booktitle = {CVPR},
 year = {2023}}

@article{qian2023gaussianavatars,
  title={Gaussianavatars: Photorealistic head avatars with rigged 3d gaussians},
  author={Qian, Shenhan and Kirschstein, Tobias and Schoneveld, Liam and Davoli, Davide and Giebenhain, Simon and Nie{\ss}ner, Matthias},
  journal={arXiv preprint arXiv:2312.02069},
  year={2023}
}

@article{kirschstein2023nersemble,
  title={Nersemble: Multi-view radiance field reconstruction of human heads},
  author={Kirschstein, Tobias and Qian, Shenhan and Giebenhain, Simon and Walter, Tim and Nie{\ss}ner, Matthias},
  journal={ACM Transactions on Graphics (TOG)},
  year={2023},
  publisher={ACM New York, NY, USA}
}

@inproceedings{zhang2023blind,
  title={Blind image quality assessment via vision-language correspondence: A multitask learning perspective},
  author={Zhang, Weixia and Zhai, Guangtao and Wei, Ying and Yang, Xiaokang and Ma, Kede},
  booktitle={CVPR},
  year={2023}
}

@article{wu2023human,
  title={Human preference score v2: A solid benchmark for evaluating human preferences of text-to-image synthesis},
  author={Wu, Xiaoshi and Hao, Yiming and Sun, Keqiang and Chen, Yixiong and Zhu, Feng and Zhao, Rui and Li, Hongsheng},
  journal={arXiv preprint arXiv:2306.09341},
  year={2023}
}

@inproceedings{barron2022mip,
  title={Mip-nerf 360: Unbounded anti-aliased neural radiance fields},
  author={Barron, Jonathan T and Mildenhall, Ben and Verbin, Dor and Srinivasan, Pratul P and Hedman, Peter},
  booktitle={CVPR},
  year={2022}
}

@inproceedings{chen2023mobilenerf,
  title={Mobilenerf: Exploiting the polygon rasterization pipeline for efficient neural field rendering on mobile architectures},
  author={Chen, Zhiqin and Funkhouser, Thomas and Hedman, Peter and Tagliasacchi, Andrea},
  booktitle={CVPR},
  year={2023}
}

@inproceedings{SSS-cvpr07,
  title =      "Skeletal graphs for efficient structure from motion",
  author =     "Noah Snavely and Steven M. Seitz and Richard Szeliski",
  year =       "2008",
  booktitle =  "CVPR",
}

@inproceedings{agarwal2009building,
author = {Agarwal, Sameer and Snavely, Noah and Simon, Ian and Sietz, Steven M. and Szeliski, Rick},
title = {Building Rome in a Day},
booktitle = {ICCV},
year = {2009} 
}

@INPROCEEDINGS {Curless,
author = {B. Curless and S. M. Seitz and R. Szeliski and Y. Furukawa},
booktitle = {CVPR},
title = {Towards Internet-scale multi-view stereo},
year = {2010}
}

@InProceedings{sch2016,
author="Sch{\"o}nberger, Johannes L.
and Zheng, Enliang
and Frahm, Jan-Michael
and Pollefeys, Marc",
editor="Leibe, Bastian
and Matas, Jiri
and Sebe, Nicu
and Welling, Max",
title="Pixelwise View Selection for Unstructured Multi-View Stereo",
booktitle="ECCV",
year="2016"
}

@InProceedings{kim16,
author="Kim, Kichang
and Torii, Akihiko
and Okutomi, Masatoshi",
editor="Leibe, Bastian
and Matas, Jiri
and Sebe, Nicu
and Welling, Max",
title="Multi-view Inverse Rendering Under Arbitrary Illumination and Albedo",
booktitle="ECCV ",
year="2016"
}

@inproceedings{martinbrualla2020nerfw,
    author = {Martin-Brualla, Ricardo and Radwan, Noha
    and Sajjadi, Mehdi S. M.
    and Barron, Jonathan T.
    and Dosovitskiy, Alexey
    and Duckworth, Daniel},
    title = {{NeRF in the Wild: Neural Radiance Fields for Unconstrained Photo Collections}},
    booktitle = {CVPR},
    year={2021}
}

@inproceedings{sun2022neuconw,
  title={Neural {3D} Reconstruction in the Wild},
  author={Sun, Jiaming and Chen, Xi and Wang, Qianqian and Li, Zhengqi and Averbuch-Elor, Hadar and Zhou, Xiaowei and Snavely, Noah},
  booktitle={SIGGRAPH Conference Proceedings},
  year={2022}
}

@INPROCEEDINGS{iraface,
  author={Kemelmacher-Shlizerman, Ira},
  title={Internet Based Morphable Model}, 
  booktitle={ICCV}, 
  year={2013}}

@INPROCEEDINGS{ira-steve,
  author={Kemelmacher-Shlizerman, Ira and Seitz, Steven M.},
  booktitle={ICCV}, 
  title={Face reconstruction in the wild}, 
  year={2011}}

@article{kirschstein2023diffusionavatars,
  title={DiffusionAvatars: Deferred Diffusion for High-fidelity 3D Head Avatars},
  author={Kirschstein, Tobias and Giebenhain, Simon and Nie{\ss}ner, Matthias},
  journal={arXiv preprint arXiv:2311.18635},
  year={2023}
}

@inproceedings{zhang2023adding,
  title={Adding conditional control to text-to-image diffusion models},
  author={Zhang, Lvmin and Rao, Anyi and Agrawala, Maneesh},
  booktitle={ICCV},
  year={2023}
}

@INPROCEEDINGS{PersonalizedFaceModeling,
author={Bindita Chaudhuri and Noranart Vesdapunt and Linda Shapiro and Baoyuan Wang},
title={Personalized Face Modeling for Improved Face Reconstruction and Motion Retargeting},
booktitle={ECCV},
year={2020}
}

@article{DBLP:journals/pami/ZhuYHLWL23,
  author       = {Xiangyu Zhu and
                  Chang Yu and
                  Di Huang and
                  Zhen Lei and
                  Hao Wang and
                  Stan Z. Li},
  title        = {Beyond 3DMM: Learning to Capture High-Fidelity 3D Face Shape},
  journal      = {{IEEE} Trans. Pattern Anal. Mach. Intell.},
  year         = {2023}
}

@inproceedings{DBLP:conf/wacv/RaiGPVTAKPT24,
  author       = {Aashish Rai and
                  Hiresh Gupta and
                  Ayush Pandey and
                  Francisco Vicente{-}Carrasco and
                  Shingo Jason Takagi and
                  Amaury Aubel and
                  Daeil Kim and
                  Aayush Prakash and
                  Fernando De la Torre},
  title        = {Towards Realistic Generative 3D Face Models},
  booktitle    = {{IEEE/CVF} Winter Conference on Applications of Computer Vision, {WACV}
                  2024, Waikoloa, HI, USA, January 3-8, 2024},
  publisher    = {{IEEE}},
  year         = {2024}
}

@inproceedings{kemelmacher2014illumination,
  title={Illumination-aware age progression},
  author={Kemelmacher-Shlizerman, Ira and Suwajanakorn, Supasorn and Seitz, Steven M},
  booktitle={CVPR},
  year={2014}
}

@article{suo2009compositional,
  title={A compositional and dynamic model for face aging},
  author={Suo, Jinli and Zhu, Song-Chun and Shan, Shiguang and Chen, Xilin},
  journal={IEEE Transactions on Pattern Analysis and Machine Intelligence},
  year={2009},
  publisher={IEEE}
}

@article{suo2012concatenational,
  title={A concatenational graph evolution aging model},
  author={Suo, Jinli and Chen, Xilin and Shan, Shiguang and Gao, Wen and Dai, Qionghai},
  journal={IEEE transactions on pattern analysis and machine intelligence},
  year={2012},
  publisher={IEEE}
}

@article{lanitis2002toward,
  title={Toward automatic simulation of aging effects on face images},
  author={Lanitis, Andreas and Taylor, Christopher J. and Cootes, Timothy F},
  journal={IEEE Transactions on pattern Analysis and machine Intelligence},
  year={2002},
  publisher={IEEE}
}

@inproceedings{wang2016recurrent,
  title={Recurrent face aging},
  author={Wang, Wei and Cui, Zhen and Yan, Yan and Feng, Jiashi and Yan, Shuicheng and Shu, Xiangbo and Sebe, Nicu},
  booktitle={CVPR},
  year={2016}
}

@inproceedings{he2019s2gan,
  title={S2gan: Share aging factors across ages and share aging trends among individuals},
  author={He, Zhenliang and Kan, Meina and Shan, Shiguang and Chen, Xilin},
  booktitle={ICCV},
  year={2019}
}

@inproceedings{wang2018face,
  title={Face aging with identity-preserved conditional generative adversarial networks},
  author={Wang, Zongwei and Tang, Xu and Luo, Weixin and Gao, Shenghua},
  booktitle={CVPR},
  year={2018}
}

@article{yang2016face,
  title={Face aging effect simulation using hidden factor analysis joint sparse representation},
  author={Yang, Hongyu and Huang, Di and Wang, Yunhong and Wang, Heng and Tang, Yuanyan},
  journal={IEEE Transactions on Image Processing},
  year={2016},
  publisher={IEEE}
}

@inproceedings{ruiz2023dreambooth,
  title={Dreambooth: Fine tuning text-to-image diffusion models for subject-driven generation},
  author={Ruiz, Nataniel and Li, Yuanzhen and Jampani, Varun and Pritch, Yael and Rubinstein, Michael and Aberman, Kfir},
  booktitle={CVPR},
  year={2023}
}

@article{hu2021lora,
  title={Lora: Low-rank adaptation of large language models},
  author={Hu, Edward J and Shen, Yelong and Wallis, Phillip and Allen-Zhu, Zeyuan and Li, Yuanzhi and Wang, Shean and Wang, Lu and Chen, Weizhu},
  journal={arXiv preprint arXiv:2106.09685},
  year={2021}
}

@misc{DreamMachine,
  author = {AI, Luma},
  title = {Luma Dream Machine},
  url = {https://lumalabs.ai/dream-machine},
  organization = {Luma Dream Machine}
}

@misc{Kling,
  url = {https://kling.kuaishou.com/en},
  urldate = {2024-09-06},
  year = {2024},
  organization = {Kuaishou.com}
}

@misc{Gen3_runway,
  title = {Runway Research | Introducing Gen-3 Alpha: A New Frontier for Video Generation},
  url = {https://runwayml.com/research/introducing-gen-3-alpha?utm_source=xinquji},
  year = {2024},
  organization = {Runwayml.com}
}

@article{chen2024dge,
      title={DGE: Direct Gaussian 3D Editing by Consistent Multi-view Editing},
      author={Minghao Chen and Iro Laina and Andrea Vedaldi},
      journal={arXiv preprint arXiv:2404.18929},
      year={2024}
}

@article{Identity_development,
  author = {Waterman, A. S.},
  title = {Identity development from adolescence to adulthood: an extension of theory and a review of research.},
  journal = {Developmental Psychology},
  year = {1982},
  doi = {10.1037/0012-1649.18.3.341}
}

@book{kro,
author = {Kroger, Jane},
year = {2007},
title = {Identity Development: Adolescence Through Adulthood}
}

@inproceedings{bezryadin2007brightness,
  title={Brightness calculation in digital image processing},
  author={Bezryadin, Sergey and Bourov, Pavel and Ilinih, Dmitry},
  booktitle={International symposium on technologies for digital photo fulfillment},
  year={2007},
  organization={Society for Imaging Science and Technology}
}

@inproceedings{chen2014cross,
  title={Cross-age reference coding for age-invariant face recognition and retrieval},
  author={Chen, Bor-Chun and Chen, Chu-Song and Hsu, Winston H},
  booktitle={ECCV},
  year={2014},
  organization={Springer}
}

@inproceedings{caliskan2025pav,
  title={PAV: Personalized Head Avatar from Unstructured Video Collection},
  author={Caliskan, Akin and Kicanaoglu, Berkay and Kim, Hyeongwoo},
  booktitle={European Conference on Computer Vision},
  pages={109--125},
  year={2025},
  organization={Springer}
}

@article{chu2024generalizable,
  title={Generalizable and Animatable Gaussian Head Avatar},
  author={Chu, Xuangeng and Harada, Tatsuya},
  journal={arXiv preprint arXiv:2410.07971},
  year={2024}
}

@inproceedings{xu2024gaussian,
  title={Gaussian head avatar: Ultra high-fidelity head avatar via dynamic gaussians},
  author={Xu, Yuelang and Chen, Benwang and Li, Zhe and Zhang, Hongwen and Wang, Lizhen and Zheng, Zerong and Liu, Yebin},
  booktitle={Proceedings of the IEEE/CVF Conference on Computer Vision and Pattern Recognition},
  pages={1931--1941},
  year={2024}
}

@inproceedings{shao2024splattingavatar,
  title={Splattingavatar: Realistic real-time human avatars with mesh-embedded gaussian splatting},
  author={Shao, Zhijing and Wang, Zhaolong and Li, Zhuang and Wang, Duotun and Lin, Xiangru and Zhang, Yu and Fan, Mingming and Wang, Zeyu},
  booktitle={Proceedings of the IEEE/CVF Conference on Computer Vision and Pattern Recognition},
  pages={1606--1616},
  year={2024}
}

@article{li2024generalizable,
  title={Generalizable one-shot 3D neural head avatar},
  author={Li, Xueting and De Mello, Shalini and Liu, Sifei and Nagano, Koki and Iqbal, Umar and Kautz, Jan},
  journal={Advances in Neural Information Processing Systems},
  volume={36},
  year={2024}
}

@article{chu2024gpavatar,
  title={GPAvatar: Generalizable and precise head avatar from image (s)},
  author={Chu, Xuangeng and Li, Yu and Zeng, Ailing and Yang, Tianyu and Lin, Lijian and Liu, Yunfei and Harada, Tatsuya},
  journal={arXiv preprint arXiv:2401.10215},
  year={2024}
}

@inproceedings{deng2025portrait4d,
  title={Portrait4d-v2: Pseudo multi-view data creates better 4d head synthesizer},
  author={Deng, Yu and Wang, Duomin and Wang, Baoyuan},
  booktitle={European Conference on Computer Vision},
  pages={316--333},
  year={2025},
  organization={Springer}
}

@article{ye2024real3d,
  title={Real3d-portrait: One-shot realistic 3d talking portrait synthesis},
  author={Ye, Zhenhui and Zhong, Tianyun and Ren, Yi and Yang, Jiaqi and Li, Weichuang and Huang, Jiawei and Jiang, Ziyue and He, Jinzheng and Huang, Rongjie and Liu, Jinglin and others},
  journal={arXiv preprint arXiv:2401.08503},
  year={2024}
}

@inproceedings{xu2023avatarmav,
  title={Avatarmav: Fast 3d head avatar reconstruction using motion-aware neural voxels},
  author={Xu, Yuelang and Wang, Lizhen and Zhao, Xiaochen and Zhang, Hongwen and Liu, Yebin},
  booktitle={ACM SIGGRAPH 2023 Conference Proceedings},
  pages={1--10},
  year={2023}
}

@inproceedings{zhang2023learning,
  title={Learning Neural Proto-Face Field for Disentangled 3D Face Modeling in the Wild},
  author={Zhang, Zhenyu and Chen, Renwang and Cao, Weijian and Tai, Ying and Wang, Chengjie},
  booktitle={Proceedings of the IEEE/CVF Conference on Computer Vision and Pattern Recognition},
  pages={382--393},
  year={2023}
}

@article{oquab2023dinov2,
  title={Dinov2: Learning robust visual features without supervision},
  author={Oquab, Maxime and Darcet, Timoth{\'e}e and Moutakanni, Th{\'e}o and Vo, Huy and Szafraniec, Marc and Khalidov, Vasil and Fernandez, Pierre and Haziza, Daniel and Massa, Francisco and El-Nouby, Alaaeldin and others},
  journal={arXiv preprint arXiv:2304.07193},
  year={2023}
}

@article{xu2024gphmv2,
    title={GPHM: Gaussian Parametric Head Model for Monocular Head Avatar Reconstruction},
    author={Xu, Yuelang and Su, Zhaoqi and Wu, Qingyao and Liu, Yebin},
    booktitle={ArXiv},
    year={2024}
}

@misc{bai2025qwen25vl,
    title={Qwen2.5-VL Technical Report},
    author={Shuai Bai and Keqin Chen and Xuejing Liu and Jialin Wang and Wenbin Ge and Sibo Song and Kai Dang and Peng Wang and Shijie Wang and Jun Tang and Humen Zhong and Yuanzhi Zhu and Mingkun Yang and Zhaohai Li and Jianqiang Wan and Pengfei Wang and Wei Ding and Zheren Fu and Yiheng Xu and Jiabo Ye and Xi Zhang and Tianbao Xie and Zesen Cheng and Hang Zhang and Zhibo Yang and Haiyang Xu and Junyang Lin},
    year={2025},
    eprint={2502.13923},
    archivePrefix={arXiv},
    primaryClass={cs.CV}
}

@inproceedings{
hu2022lora,
title={Lo{RA}: Low-Rank Adaptation of Large Language Models},
author={Edward J Hu and Yelong Shen and Phillip Wallis and Zeyuan Allen-Zhu and Yuanzhi Li and Shean Wang and Lu Wang and Weizhu Chen},
booktitle={International Conference on Learning Representations},
year={2022},
url={https://openreview.net/forum?id=nZeVKeeFYf9}
}

@misc{labs2025flux1kontextflowmatching,
      title={FLUX.1 Kontext: Flow Matching for In-Context Image Generation and Editing in Latent Space},
      author={Black Forest Labs and Stephen Batifol and Andreas Blattmann and Frederic Boesel and Saksham Consul and Cyril Diagne and Tim Dockhorn and Jack English and Zion English and Patrick Esser and Sumith Kulal and Kyle Lacey and Yam Levi and Cheng Li and Dominik Lorenz and Jonas Müller and Dustin Podell and Robin Rombach and Harry Saini and Axel Sauer and Luke Smith},
      year={2025},
      eprint={2506.15742},
      archivePrefix={arXiv},
      primaryClass={cs.GR},
      url={https://arxiv.org/abs/2506.15742},
}

@misc{wu2025qwenimage,
    title={Qwen-Image Technical Report},
    author={Chenfei Wu and Jiahao Li and Jingren Zhou and Junyang Lin and Kaiyuan Gao and Kun Yan and Sheng-ming Yin and Shuai Bai and Xiao Xu and Yilei Chen and Yuxiang Chen and Zecheng Tang and Zekai Zhang and Zhengyi Wang and An Yang and Bowen Yu and Chen Cheng and Dayiheng Liu and Deqing Li and Hang Zhang and Hao Meng and Hu Wei and Jingyuan Ni and Kai Chen and Kuan Cao and Liang Peng and Lin Qu and Minggang Wu and Peng Wang and Shuting Yu and Tingkun Wen and Wensen Feng and Xiaoxiao Xu and Yi Wang and Yichang Zhang and Yongqiang Zhu and Yujia Wu and Yuxuan Cai and Zenan Liu},
    year={2025},
    eprint={2508.02324},
    archivePrefix={arXiv},
    primaryClass={cs.CV}
}
}


\end{document}